\documentclass{article}
\PassOptionsToPackage{table}{xcolor}
\usepackage{iclr2027_conference,times}

\usepackage{microtype}
\usepackage{graphicx}
\usepackage{subcaption}
\usepackage{booktabs}
\usepackage{float}

\usepackage[table]{xcolor}
\usepackage{hyperref}
\usepackage{xspace}
\usepackage{wrapfig}
\usepackage{url}

\usepackage{amsmath}
\usepackage{amssymb}
\usepackage{mathtools}
\usepackage{amsthm}
\usepackage{enumitem}
\usepackage{algorithm}
\usepackage{algorithmic}

\usepackage{longtable}
\usepackage{tabularx}

\newcommand{\dataname}{\textsc{Chimera}\xspace}



\usepackage[capitalize,noabbrev]{cleveref}

\theoremstyle{plain}

\theoremstyle{definition}

\theoremstyle{remark}

\usepackage[textsize=tiny]{todonotes}
\usepackage{setspace}

\newcommand{\eg}{{\sl e.g.}}
\newcommand{\linkicon}[1]{\raisebox{-1pt}{\includegraphics[height=9pt]{figure/logos/#1}}}

\title{%
  \vspace{-0.45cm}
  \raisebox{-0.3cm}{\includegraphics[height=1.2cm]{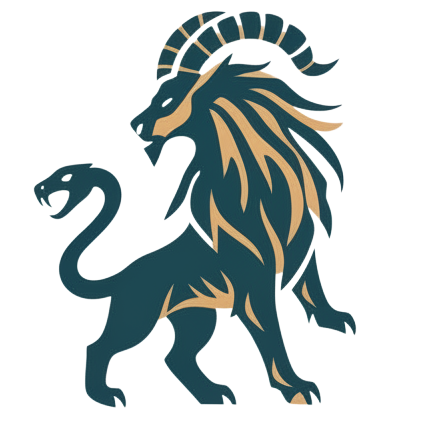}}%
  \hspace{-0.2cm} \dataname: Compact Synthetic Data for\\ Generalizable LLM Reasoning%
  \vspace{0.15cm}
}

\author{%
  Xinyu Zhu$^{1}$\thanks{Correspondences:
    \texttt{\{xinyuzhu, yumeng5\}@virginia.edu},
    \texttt{gabrielzhuyun@gmail.com}} \quad
  Yihao Feng$^{2}$ \quad
  Yanchao Sun$^{2}$ \quad
  Xianzhi Du$^{2}$ \\[0.3em]
  \textbf{Pingzhi Li}$^{3}$ \quad
  \textbf{Olli Saarikivi}$^{2}$ \quad
  \textbf{Yun Zhu}$^{2}$\footnotemark[1] \quad
  \textbf{Yu Meng}$^{1}$\footnotemark[1] \\[0.5em]
  $^{1}$University of Virginia \quad $^{2}$Apple \quad
  $^{3}$University of North Carolina at Chapel Hill\\[4pt]
  \hphantom{$^{1}$\,}\linkicon{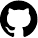}~\href{https://github.com/TianHongZXY/CHIMERA}{Code}\quad
\linkicon{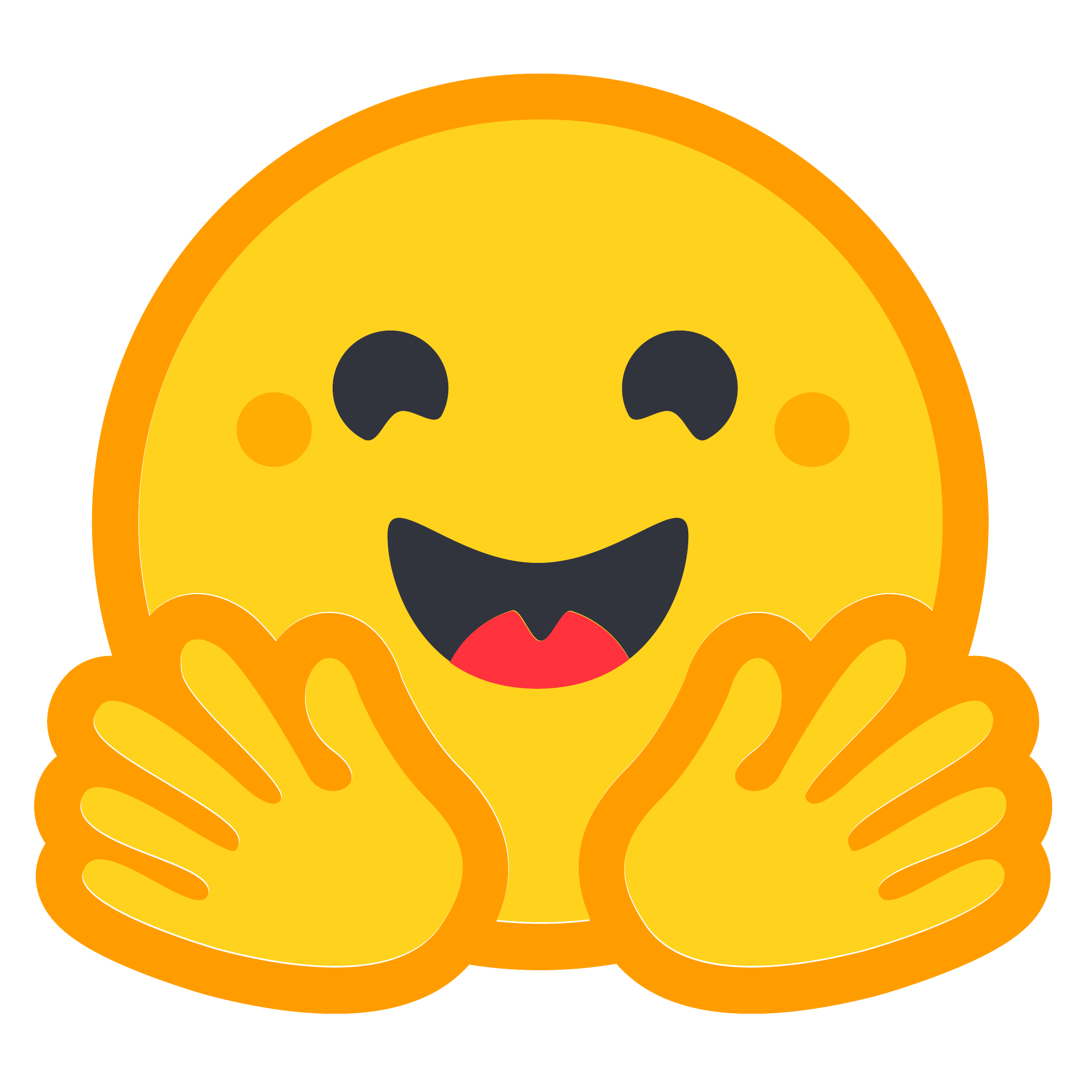}~\href{https://huggingface.co/datasets/TianHongZXY/CHIMERA}{Dataset}\quad
\linkicon{huggingface}~\href{https://huggingface.co/collections/TianHongZXY/chimera}{Models}}

\iclrfinalcopy
\begin{document}

\maketitle

\begin{abstract}
Large Language Models (LLMs) have recently exhibited remarkable reasoning capabilities, largely enabled by supervised fine-tuning (SFT)- and reinforcement learning (RL)-based post-training on high-quality reasoning data. However, reproducing and extending these capabilities in open and scalable settings is hindered by three fundamental data-centric challenges:
(1) the \textit{cold-start problem}, arising from the lack of seed datasets with detailed, long Chain-of-Thought (CoT) trajectories needed to initialize reasoning policies;
(2) \textit{limited domain coverage}, as most existing open-source reasoning datasets are concentrated in mathematics, with limited coverage of broader scientific disciplines; and
(3) the \textit{annotation bottleneck}, where the difficulty of frontier-level reasoning tasks makes reliable human annotation prohibitively expensive or infeasible.
To address these challenges, we introduce \dataname, a compact synthetic reasoning dataset comprising 9K samples designed to support generalizable reasoning across domains.
\dataname is constructed with three key properties:
(1) it provides rich, long CoT reasoning trajectories synthesized by state-of-the-art reasoning models;
(2) it has broad and structured coverage, spanning 8 major scientific disciplines and over 1K fine-grained topics organized via a model-generated hierarchical taxonomy; and
(3) it employs a fully automated, scalable evaluation pipeline that uses strong reasoning models to cross-validate both problem validity and answer correctness, removing the reliance on human annotations.
We use \dataname to post-train a 4B Qwen3 model using a combination of SFT and RL. Despite the dataset's modest size, the resulting model achieves strong performance on a suite of challenging reasoning benchmarks, including GPQA-Diamond, AIME 24/25/26, HMMT 25 and Humanity's Last Exam, approaching or matching the reasoning performance of substantially larger models such as DeepSeek-R1 and Qwen3-235B.%
\end{abstract}

\section{Introduction}

Large language models (LLMs) have recently demonstrated substantial advances in complex, multi-step reasoning across mathematics, science, and general problem-solving tasks~\citep{guo2025deepseek, jaech2024openai,lambert2024t}. These capabilities are largely enabled by post-training procedures, most notably reinforcement learning~\citep{shao2024deepseekmath, minimax-m1, team2025kimi}, that encourage LLMs to generate explicit intermediate reasoning steps, often in the form of long CoT trajectories.
Models trained under such regimes exhibit improved planning, abstraction, and self-correction behaviors~\citep{gandhi2025cognitive, zeng2025simplerl, muennighoff2025s1}.
Despite these successes, replicating and extending frontier-level reasoning capabilities in open and resource-constrained settings remains challenging.
In particular, progress is increasingly limited not by training techniques, but by the availability and quality of reasoning data.
As reasoning-oriented post-training becomes more central to LLM development, the construction of scalable, high-quality reasoning datasets has emerged as a key bottleneck.

\begin{figure*}[!t]
    \centering
    \includegraphics[width=1.0\linewidth]{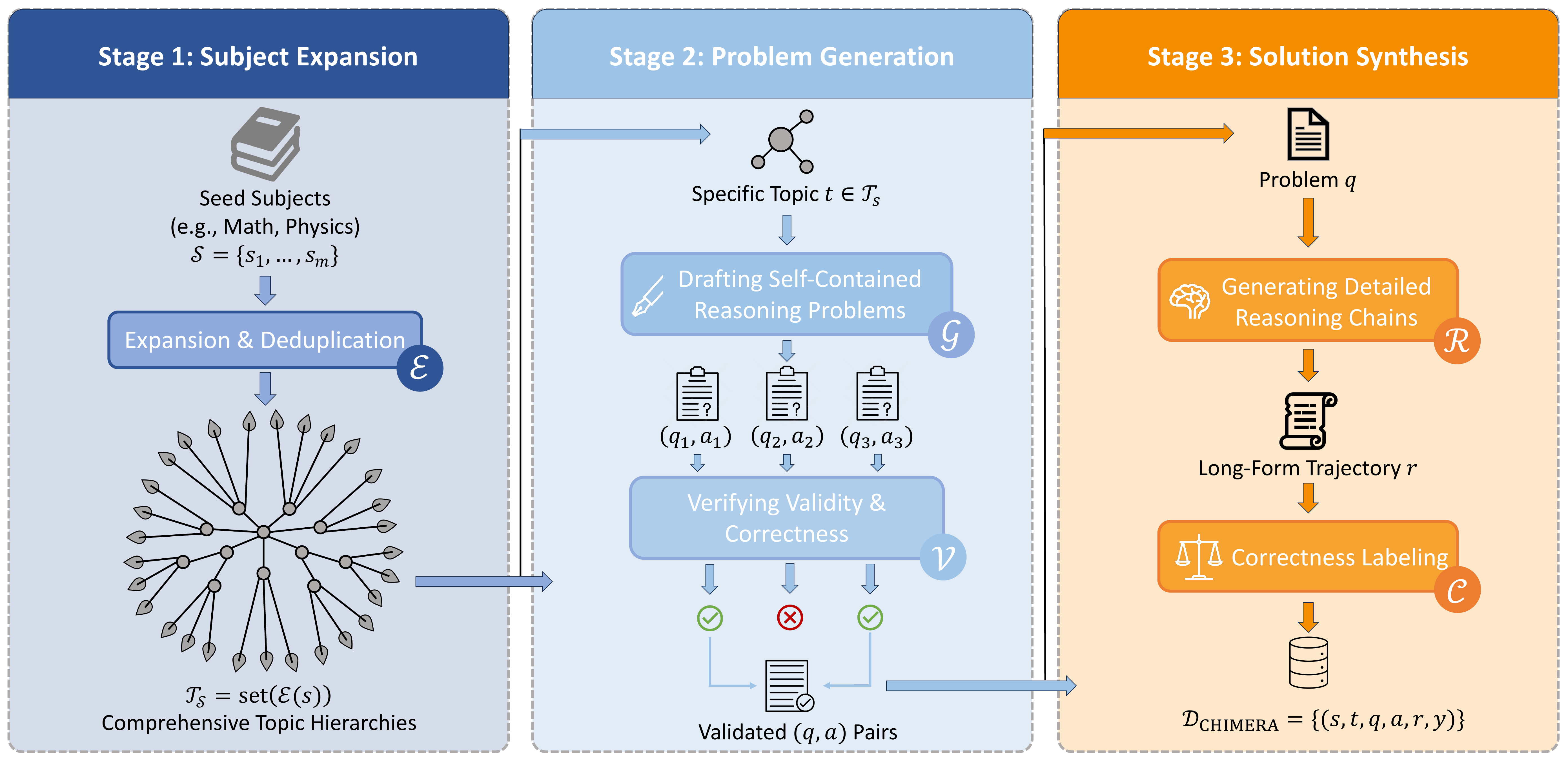}
    \caption{Data synthesis pipeline overview: stage 1 expands a small set of seed subjects into thousands of fine-grained topics; stage 2 creates well-defined problems with concise, verifiable answers based on these topics; stage 3 generates detailed reasoning trajectories and labels their correctness.}
    \label{fig:overview}
    \vspace{-6mm}

\end{figure*}

Notably, three fundamental data-centric challenges hinder the development of open general-purpose reasoning models:
(1) \textbf{Cold-start data scarcity}. Effective reasoning-oriented post-training typically requires an initial corpus of examples containing detailed, long CoT trajectories to bootstrap policy learning~\citep{wang_octothinker_2025, yeo2025demystifying}.
However, existing datasets often provide either only ground-truth answers~\citep{yu2025dapo} or brief explanations~\citep{cobbe2021gsm8k,hendrycks2021math,numina_math_datasets, yumetamath}, which are insufficient for initializing models that must learn to perform long-horizon complex reasoning.
This cold-start problem is particularly critical for smaller or mid-sized models, which have more reliance on the quality and structure of supervision.
(2) \textbf{Limited domain coverage}.
Most publicly available reasoning datasets focus narrowly on mathematics and coding tasks~\citep{yumetamath, deepmath103k2024, yu2025dapo, huggingface2025openr1, openthoughts2024}.
While these domains are valuable, they represent only a small fraction of the reasoning demands encountered in real-world problem solving.
As a result, models trained on these datasets often struggle to generalize their reasoning strategies to other scientific disciplines or interdisciplinary problems.
(3) \textbf{The annotation bottleneck}.
As reasoning benchmarks approach or exceed human expert difficulty, reliable manual annotation becomes increasingly impractical~\citep{hle2025, wang2023selfinstruct,guha2025openthoughts,manem2025sandmathusingllmsgenerate}.
Producing correct answers and especially high-quality CoT explanations for frontier-level questions requires deep domain expertise and substantial time investment.
This makes large-scale human annotation costly, slow, and in many cases unreliable.

Recent advances in synthetic data generation suggest a promising direction~\citep{lee2024rlaif, cui2023ultrafeedback, Xu2024MagpieADA}: leveraging strong models themselves to synthesize high-quality training data.
When carefully designed, model-generated CoT trajectories can exhibit rich intermediate structure, cover diverse reasoning patterns, and scale to domains where human annotation is infeasible.
However, naive synthetic data generation risks narrow coverage, error accumulation, and uncontrolled quality, limiting its effectiveness for post-training~\citep{yu2023data_generator, chen2024on_the_diversity}.
In this work, we explore whether a compact yet carefully constructed synthetic dataset can meaningfully support reasoning-oriented post-training.
Rather than maximizing dataset size, we focus on three design principles: (1) rich, long CoT reasoning trajectories, (2) broad and structured domain coverage, and (3) scalable quality control without human supervision.

Guided by these principles, we introduce \dataname, a synthetic reasoning dataset consisting of 9K high-quality samples.
Each sample includes a long CoT reasoning trajectory generated by state-of-the-art reasoning models, providing rich supervision for multi-step reasoning behaviors.
For broad domain coverage, \dataname spans 8 major scientific disciplines and over 1K fine-grained topics.
These topics are derived from a structured, model-generated hierarchical taxonomy, enabling systematic coverage of both core concepts and specialized subfields.
To ensure quality and scalability, \dataname employs a fully automated evaluation protocol.
Instead of relying on human annotators, we use strong reasoning models to cross-validate problem validity and answer correctness and filter low-quality or inconsistent samples.

We evaluate the effectiveness of \dataname by post-training a Qwen3-4B model~\citep{yang2025qwen3} using a combination of SFT and RL.
Despite the compact size of \dataname, the resulting model demonstrates strong reasoning performance across a diverse set of challenging benchmarks, including GPQA-Diamond~\citep{rein2024gpqa}, AIME24, AIME25~\citep{aime2025problems}, AIME26, HMMT25~\citep{balunovic_srimatharena_2025}, and Humanity's Last Exam (HLE)~\citep{hle2025}. Notably, the fine-tuned 4B model achieves performance competitive with substantially larger models (\eg, DeepSeek-R1 and Qwen3-235B).

In summary, our contributions are as follows:
\begin{enumerate}[leftmargin=2em]
\item We identify and formalize key data-centric challenges that limit scalable reasoning post-training for LLMs.
\item We introduce \dataname, a compact, fully synthetic dataset featuring long CoT trajectories, broad subject coverage, and automated quality control.
\item We show that post-training on \dataname enables a 4B-parameter model to match or approach the performance of substantially larger models (\eg, DeepSeek-R1 and Qwen3-235B) across a range of challenging reasoning benchmarks.
\end{enumerate}

\section{Method}
\label{sec:method}

\subsection{Overview}
\label{sec:overview}

\begin{wrapfigure}{r}{0.45\textwidth}
\vspace{-1em}
\centering
\small
\hrule height 1pt
\vspace{0.4em}
\captionsetup{type=algorithm, labelfont=bf}
\captionof{algorithm}{\textsc{Chimera} Data Synthesis Pipeline}
\label{alg:pipeline}
\vspace{-0.6em}
\hrule
\vspace{0.4em}

\begin{minipage}{\linewidth}
\begin{algorithmic}[1]
\REQUIRE Seed subjects $\mathcal{S}=\{s_1,\dots,s_m\}$; number of samples per topic $n$;
topic expander $\mathcal{E}$;
problem/answer generator $\mathcal{G}$;
problem validator $\mathcal{V}$;
correctness verifier $\mathcal{C}$;
reasoning trajectory generator $\mathcal{R}$
\ENSURE \dataname dataset $\mathcal{D}_{\text{\dataname}}$

\STATE Initialize $\mathcal{D}_{\text{\dataname}} \leftarrow \emptyset$

\STATE \textbf{Stage 1: Subject expansion} \hfill \textcolor{red}{\S2.2}
\FOR{each subject $s \in \mathcal{S}$}
    \STATE $\mathcal{T}_s \leftarrow \mathrm{set}\!\left(\mathcal{E}(s)\right)$
\ENDFOR

\STATE \textbf{Stage 2: Problem generation} \hfill \textcolor{red}{\S2.3}
\FOR{each subject $s \in \mathcal{S}$}
    \FOR{each topic $t \in \mathcal{T}_{s}$}
        \FOR{$j=1$ to $n$}
            \STATE $(q, a) \leftarrow \mathcal{G}(t)$
            \IF{\textbf{not} $\mathcal{V}(q,a)$}
                \STATE \textbf{continue}
            \ENDIF

            \STATE \textbf{Stage 3: Solution synthesis} \hfill \textcolor{red}{\S2.4}
            \STATE $r \leftarrow \mathcal{R}(q)$
            \STATE $y \leftarrow \mathcal{C}(q, a, r)$
            \STATE $\mathcal{D}_{\dataname} \leftarrow
            \mathcal{D}_{\dataname} \cup \{(s,t,q,a,r,y)\}$
        \ENDFOR
    \ENDFOR
\ENDFOR
\STATE \textbf{return} $\mathcal{D}_{\text{\dataname}}$
\end{algorithmic}
\end{minipage}
\vspace{0.4em}
\hrule
\vspace{-1em}
\end{wrapfigure}

Our goal is to automatically construct a reasoning dataset that covers as broad a range of topics as possible without human annotation. To this end, we propose a modular, LLM-driven data synthesis pipeline that consists of three decoupled stages, illustrated in Figure~\ref{fig:overview}. The complete procedure is formalized in Algorithm~\ref{alg:pipeline}. Specifically, the pipeline comprises:
(1) \textbf{subject expansion}, given a small set of high-level subjects (\eg, math, physics, etc.), we leverage \texttt{gpt-5-2025-08-07} (hereafter referred to as \texttt{gpt-5}) to list as many topics under the subject as possible, which leads to a comprehensive topic hierarchy for each subject;
(2) \textbf{problem generation}, given a specific topic, we further use \texttt{gpt-5} to propose a clear, self-contained and easy-to-verify problem along with the corresponding answer, this process can be repeated several times to synthesize multiple examples for one topic;
(3) \textbf{solution synthesis}, for each problem, we generate a detailed reasoning trajectory with \texttt{Qwen3-235B-A22B-Thinking-2507}, a state-of-the-art open reasoning language model.\footnote{Many proprietary LLMs such as \texttt{gpt-5} do not provide full access to its intermediate thinking trajectory and thus cannot be used to synthesize the detailed solution.}

All the stages are separate and intermediate artifacts are saved, enabling subsequent dataset filtering and curation.
The pipeline is designed to be simple, scalable, and extensible, allowing new subjects to be added or adjusting topic distributions with little additional effort. All the prompts (e.g., subject expansion, problem generation, solution synthesis, problem validator and correctness verifier) can be found in Appendix~\ref{appendix:prompts}.

\subsection{Subject Expansion}
\label{sec:subject-expansion}
We first collect a small set of high-level subjects $\mathcal{S}$ that are broad and abstract, such as mathematics, physics, and computer science. These subjects are intentionally coarse-grained to minimize human design choices and to encourage wide domain coverage.
For each subject $s$, we prompt \texttt{gpt-5} to generate a comprehensive list of fine-grained topics $\mathcal{T}_{s}$ that span the conceptual space of the subject. To cover the most foundational concepts of mathematics, which has too many subfields, we sample multiple times for it. After the expansion, we conduct deduplication to ensure each topic is unique. The resulting topic lists form a hierarchical taxonomy that serves as the backbone of the dataset.
By decoupling topic expansion from later stages, the pipeline allows new subjects to be added or existing topic distributions to be modified easily. All the subjects and topics can be found in Appendix~\ref{appendix:sub2topic}.

\begin{table*}[!t]
\centering
\caption{Statistics of commonly used reasoning datasets. \dataname\ features substantially longer problem statements and more detailed solutions than existing datasets (lengths are measured in words), enabling complex and long-horizon reasoning for training modern LLMs.}

\resizebox{\textwidth}{!}{
\begin{tabular}{lccccccc}
\toprule
\textbf{Dataset} &
\textbf{\# Problems} &
\textbf{\# Subjects} &
\textbf{\# Topics} &
\textbf{Prompt} &
\textbf{Solution} &
\textbf{Answer} &
\textbf{Solution} \\
& & & & \textbf{Length} & \textbf{Length} & \textbf{Format} & \textbf{Annotator}\\
\midrule

\multicolumn{8}{c}{\textit{Human-Curated Reasoning Datasets}} \\
\midrule
\textbf{GSM8K} & $7{,}473$ & $1$ & -- & $45.1$ & $51.7$ & Numeric & Human\\
\textbf{MATH} & $7{,}500$ & $1$ & -- & $33.0$ & $89.5$ & Free-form & Human \\
\textbf{NuminaMath-CoT} & $859{,}494$ & $1$ & -- & $44.0$ & $205.3$ & Free-form & Human \\
\midrule

\multicolumn{8}{c}{\textit{Synthetic Reasoning Datasets}} \\
\midrule
\textbf{MetaMathQA} & $395{,}000$ & $1$ & -- & $40.5$ & $101.3$ & Free-form & AI \\
\textbf{OpenThoughts-114K} & $113{,}957$ & -- & -- & $96.3$ & $3{,}937.2$ & Free-form & AI \\
\textbf{DAPO-Math-17K} & $17{,}398$ & $1$ & – & $42.5$ & $1$ & Numeric & AI \\
\textbf{OpenR1-Math-220K} & $225{,}129$ & $1$ & -- & $43.6$ & $2{,}624.6$ & Free-form & AI \\
\textbf{TULU3-SFT} & $939{,}343$ & -- & -- & $148.6$ & $227.6$ & Free-form & AI \\
\textbf{DeepMath-103K} & $103{,}022$ & $1$ & -- & $33.7$ & $2{,}959.2$ & Free-form & AI \\
\textbf{OpenScience}\footnotemark & $315{,}579$ & -- & -- & $76.1$ & $1{,}296.8$ & Multiple-choice & AI \\

\midrule

\multicolumn{8}{c}{\textit{Our Dataset}} \\
\midrule
\textbf{\dataname} & $9{,}225$ & $\textbf{8}$ & $\textbf{1{,}179}$ &
$\textbf{211.1}$ & $\textbf{11{,}121.4}$ & Free-form & AI \\
\bottomrule
\end{tabular}
}
\label{tab:data_statistics}
\end{table*}

\begin{figure*}
    \centering
    \includegraphics[width=1.0\linewidth]{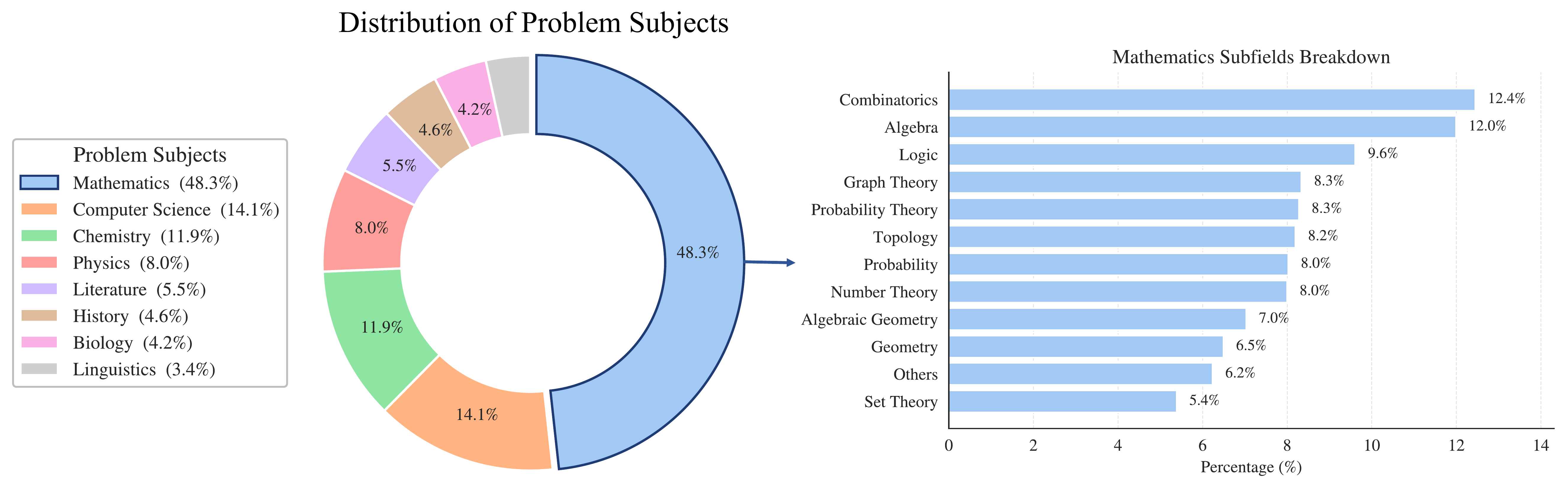}
    \caption{Distribution of problem subjects in \dataname. The left panel illustrates broad disciplinary coverage, with mathematics accounting for $48.3\%$ of the dataset, followed by computer science, chemistry and physics. The right panel decomposes the mathematics subset into fine-grained subfields. This distribution reflects the \dataname's emphasis on disciplinary breadth and topic diversity.}

    \label{fig:data_dist}
\end{figure*}

\footnotetext{OpenScience has multiple subsets; here it refers to the OS-Qwen3-235B-4 subset.}

\subsection{Problem Generation}
\label{sec:problem-generation}

Given the expanded set of topics, we generate reasoning problems by prompting \texttt{gpt-5} to produce one problem and the corresponding answer $(q, a)$ for each topic $t$. Each problem is required to satisfy the following criteria:

\begin{itemize}[leftmargin=1.5em]
    \item \textbf{Solvability and difficulty}: The problem must be solvable by an expert at a PhD level and is not an open research problem.
    \item \textbf{Self-contained}: All necessary information must be included in the problem statement.
    \item \textbf{Unambiguous and verifiable answer}: The problem should admit a clear and unique answer. The correctness of the answer should be easy to verify.
\end{itemize}

These constraints are enforced through careful prompt design (details in Appendix~\ref{appendix:prompts}) and subsequent LLM-based filtering, enabling fully automated and scalable data generation.

\paragraph{Cross-model verification of problem validity.}
To ensure the reliability of synthesized data, we perform cross-model verification after problem generation. Specifically, we employ two independent LLMs, \texttt{gpt-5} and \texttt{o4-mini}, as verifiers $\mathcal{V}$ to assess both problem validity and answer correctness. Each verifier checks whether (i) the problem is well-posed and unambiguous, and (ii) the provided answer correctly solves the problem.
A problem is retained only if it passes verification by both models. This dual-verifier design reduces the risk of systematic model bias or hallucinated solutions from a single model. By requiring agreement across independent models, we obtain a higher-confidence subset of synthesized problems with verifiable correctness. The accepted problems are stored together with their associated subjects and topics.

\subsection{Solution Synthesis}
\label{sec:solution-synthesis}
Although frontier proprietary LLMs like ChatGPT~\citep{chatgpt}, Gemini~\citep{gemini} and Claude~\citep{claude} are good at generating high-quality responses, their detailed thinking processes are often inaccessible to the users. Therefore, their generated solutions are only partially shown to the user, which are concise and brief, making it challenging to utilize them for training advanced reasoning models.

To address this issue, we apply a strong open reasoning-intensive model $\mathcal{R}$ and regenerate a detailed reasoning trajectory $r$ for each problem $q$ created in the previous stage. In our implementation, we employ \texttt{Qwen3-235B-A22B-Thinking-2507} to produce the step-by-step reasoning trajectories. Then we compare each trajectory $r$ with the original answer $a$ to the problem $q$ and label its correctness $y\in \{0, 1\}$.
Reasoning trajectories that lead to correct final answers can be used for supervised fine-tuning, while the other problems without correct reasoning trajectories are kept as problem-answer-only instances and can be used for reinforcement learning, where only the final answer is required for training. In a nutshell, each problem in \dataname\ is designed to be self-contained, verifiable, and reasoning-focused. Concrete examples can be found in Appendix~\ref{appendix:cases}.

\subsection{Dataset Statistics}

Table~\ref{tab:data_statistics} compares \dataname with representative human-curated and synthetic reasoning datasets. Human-curated datasets such as GSM8K and MATH are high-quality but domain-specific, while many synthetic datasets scale up substantially but remain narrowly focused or lack explicit subject/topic structure. \dataname instead prioritizes structured diversity and long-form reasoning. As shown in Figure~\ref{fig:data_dist}, it contains $9{,}225$ problems across $8$ subjects and $1{,}179$ topics, each with detailed reasoning trajectories. Although smaller than many synthetic datasets, it provides broader disciplinary coverage, more complex prompts, and substantially longer solutions, reaching $11$K words. Thus, \dataname complements existing resources by emphasizing difficulty, subject structure, and detailed reasoning across diverse domains.

\section{Experiments}
\label{sec:experiments}

\subsection{Experimental Setup}
\label{sec:exp-setup}

\paragraph{Training setting.}
All experiments use \texttt{Qwen3-4B-Thinking-2507} as the base model, with fine-tuned models initialized from the same checkpoint. We compare \dataname\ with public synthetic baselines. SFT is performed on verified-correct reasoning trajectories with batch size 256 and learning rate $1\mathrm{e}{-5}$. From the SFT model, we apply CISPO~\citep{minimax-m1} for one epoch using batch size 256, learning rate $1\mathrm{e}{-6}$, and 8 rollouts per prompt. RL uses both the SFT set and synthesized problems unsolved during solution synthesis but solved by the SFT model within 8 trials. For free-form answers, we use \texttt{o4-mini} as an LLM-based reward model. Training uses 8 NVIDIA H200 GPUs; SFT takes about 13 hours and RL about 7 days.

\paragraph{Benchmarks.}
We evaluate models on a diverse set of challenging reasoning benchmarks spanning scientific reasoning, mathematical problem solving, and knowledge-intensive tasks: GPQA-Diamond (GPQA-D)~\citep{rein2024gpqa}, AIME24, AIME25~\citep{aime2025problems}, AIME26, HMMT25~\citep{balunovic_srimatharena_2025} and HLE~\citep{hle2025}, for HLE we only consider text-only problems as the model is not multimodal.

\paragraph{Evaluation setting.}
For all evaluations, we use the official suggested decoding configuration: temperature = $0.6$, top-$p$ = $0.95$, top-$k$ = $20$, maximum token number = $102{,}400$. To reduce variance and fairly evaluate reasoning performance, we sample $32$ solutions per problem for AIME and HMMT, $10$ for GPQA-Diamond, and $8$ for Humanity's Last Exam.
For each problem, we report the unbiased pass@$1$, following common practice in prior works~\citep{zhu_surprising_2025}.

\paragraph{Baselines.}
We compare five settings: the base model \texttt{Qwen3-4B-Thinking-2507} without additional fine-tuning, the base model fine-tuned on OpenThoughts-114K~\citep{openthoughts2024}, OpenScience~\citep{nvidiaopenscience2024}, DeepMath-103K~\citep{deepmath103k2024}, and  \dataname.

\subsection{Main Results}
\label{sec:main-results}

Table~\ref{tab:main-results} reports performance across reasoning benchmarks. Fine-tuning on \dataname consistently improves the base model, with gains on GPQA-Diamond (+4.3), AIME24 (+5.3), HMMT Feb 25 (+6.5), HMMT Nov 25 (+9.7), and HLE (+1.7).
Although the base model has only 4B parameters, it is competitive with much larger models, matching or surpassing 8B--70B baselines on most benchmarks and approaching models two orders of magnitude larger, such as DeepSeek-R1 and Qwen3-235B-A22B. This demonstrates the data efficiency of our synthesis pipeline: compact, high-quality synthetic reasoning data can effectively improve modern LLM reasoning.

In contrast, fine-tuning on existing reasoning datasets OpenThoughts-114K, OpenScience, and DeepMath-103K all underperforms the base model, suggesting that larger reasoning datasets do not necessarily provide an effective post-training signal for already strong reasoning models. We hypothesize that this degradation is due to a mismatch between data difficulty and model capability.

Despite being substantially smaller than existing public synthetic datasets, our dataset yields stronger and consistent performance gains. This highlights the importance of data quality, broad and structured subject coverage, and detailed reasoning traces for improving reasoning capabilities in modern LLMs. We further evaluate \dataname on an additional stronger 4B reasoning checkpoint, \texttt{Qwen3.5-4B}; results in Appendix~\ref{app:qwen35} show that SFT on \dataname also improves most benchmarks on this model.

\begin{table*}[t]
\centering
\caption{Main results on reasoning benchmarks. Models are categorized into \textbf{Small to Medium Scale} ($\leq$ 70B) and \textbf{Large Scale} ($>$ 70B). Notably, fine-tuning the Qwen3-4B base model on \dataname\ yields performance competitive with substantially larger models (\eg, DeepSeek-R1, Qwen3-235B-A22B), highlighting the strong data efficiency of our dataset. The gains over the base model are statistically significant on five of the seven benchmarks (Appendix~\ref{app:significance}).}

\resizebox{\textwidth}{!}{
\begin{tabular}{l cccccccc}
\toprule
\textbf{Model} &
\textbf{\# Params} &
\textbf{GPQA-D} &
\textbf{AIME24} &
\textbf{AIME25} &
\textbf{AIME26} &
\textbf{HMMT Feb 25} &
\textbf{HMMT Nov 25} &
\textbf{HLE} \\
\midrule

\multicolumn{9}{c}{\textit{\textbf{Large Scale} ($>$ 70B)}} \\
\midrule
DeepSeek-R1 & \multicolumn{1}{c}{671B} & 71.5 & 79.8 & 70.0 & -- & 41.7 & -- & 8.5 \\
DeepSeek-R1-0528 & \multicolumn{1}{c}{671B} & 81.0 & 91.4 & 87.5 & -- & 79.4 & --  & 17.7 \\
Qwen3-235B-A22B & \multicolumn{1}{c}{235B} & 71.1 & 85.7 & 81.5 & --  & 62.5  & -- & 11.8 \\
Qwen3-235B-A22B-Thinking-2507 & \multicolumn{1}{c}{235B} & 81.1 & -- & 92.3 & -- & 83.9 & --  & 18.2 \\
o3-mini (medium) & \multicolumn{1}{c}{--} & 76.8 & 79.6 & 74.8 & -- & -- & -- & 10.3 \\
o4-mini (high) & \multicolumn{1}{c}{--} & 81.4 & 93.4 & 92.7 & -- & 66.7 & --  & 18.1 \\
gemini-2.5-pro & \multicolumn{1}{c}{--} & 86.4 & -- & 88.0 & --  & 82.5 & --  & 18.4 \\

\midrule

\multicolumn{9}{c}{\textit{\textbf{Small to Medium Scale} ($\leq$ 70B)}} \\
\midrule
Qwen3-4B-Thinking-2507 & \multicolumn{1}{c}{4B} & 65.8 & 81.6 & \textbf{81.0} & 80.8 & 59.2 & 57.3 & 7.3 \\
Qwen3-32B & \multicolumn{1}{c}{32B} & 68.4 & 81.4 & 72.9  & 74.3 & 56.6 & 50.0 & 8.9 \\
DeepSeek-R1-0528-Qwen3-8B & \multicolumn{1}{c}{8B} & 61.1 & 82.2 & 76.3 & 78.0 & 59.2 & 57.7 & 6.9 \\
DeepSeek-R1-Distill-Llama-70B & \multicolumn{1}{c}{70B} & 65.2 & 70.0 & 55.3  & 59.4  & 36.7 & 40.2 & 5.2 \\

\rowcolor{gray!10}
\multicolumn{9}{c}{\textit{Fine-tuned Qwen3-4B-Thinking-2507 (Ours)}} \\
\rowcolor{gray!10}
+ OpenThoughts-114K & \multicolumn{1}{c}{4B} & 41.9 & 60.5 & 53.2 & 55.2 & 34.8 & 37.1 & 5.7 \\
\rowcolor{gray!10}
+ OpenScience & \multicolumn{1}{c}{4B} & 53.5 & 61.7 & 53.3  & 53.0 & 40.0 & 36.9 & 4.6 \\
\rowcolor{gray!10}
+ DeepMath-103K & \multicolumn{1}{c}{4B} & 59.1 & 60.1 & 50.7  & 54.1 & 32.5 & 37.0 & 4.7 \\
\rowcolor{gray!10}
\textbf{+ \dataname} & \multicolumn{1}{c}{4B} &
\textbf{70.1} & \textbf{86.9} & 80.7 & \textbf{82.7} & \textbf{65.7} & \textbf{67.0} & \textbf{9.0} \\

\bottomrule
\end{tabular}
}
\label{tab:main-results}
\end{table*}

\begin{figure*}[!t]
    \centering
    \begin{subfigure}{0.47\textwidth}
        \centering
        \includegraphics[width=\linewidth]{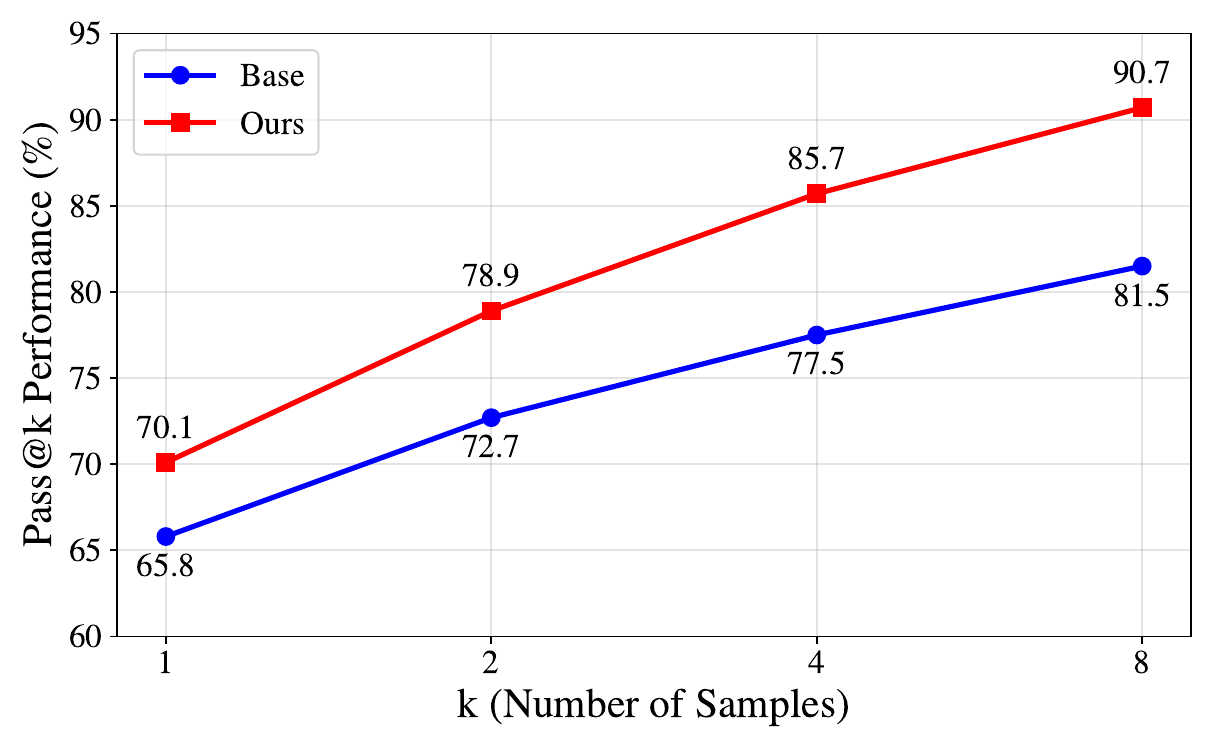}
        \vspace{-5mm}
        \caption{GPQA-Diamond}
        \label{fig:gpqa}
    \end{subfigure}
    \hspace{1.5em}
    \begin{subfigure}
    {0.47\textwidth}
        \centering
        \includegraphics[width=\linewidth]{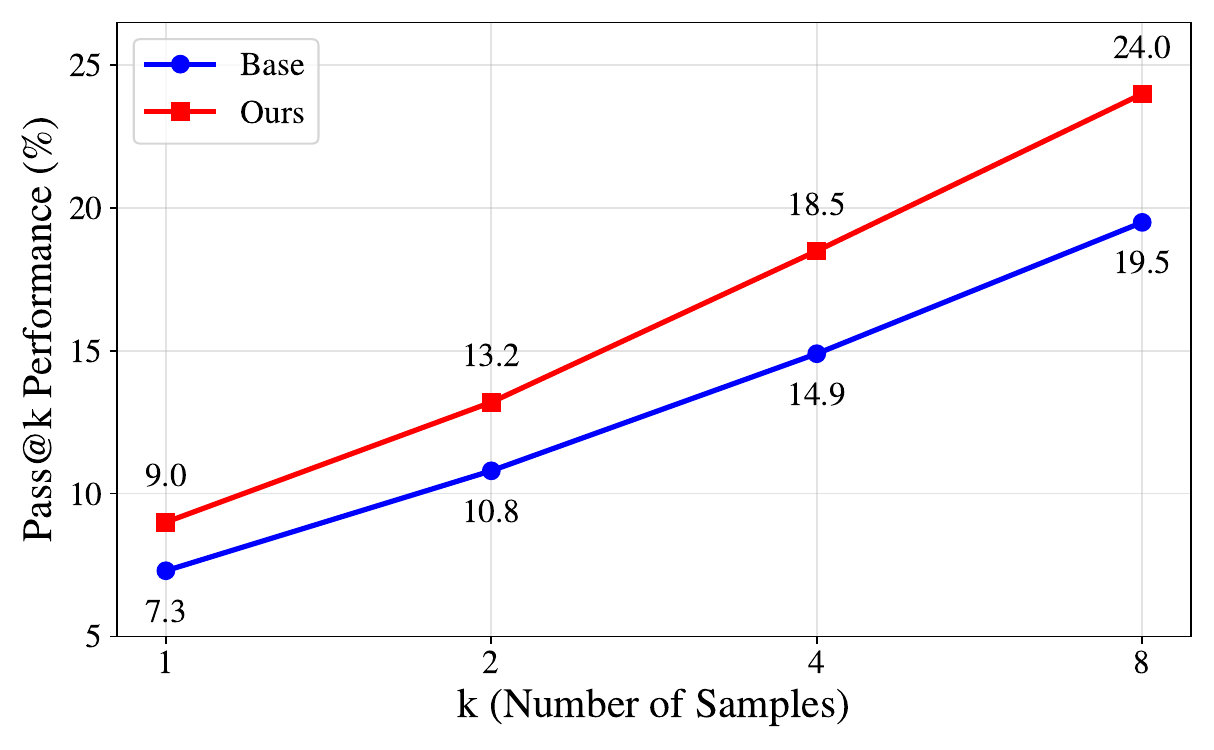}
        \vspace{-5mm}
        \caption{HLE}
        \label{fig:hle}
\end{subfigure}
\caption{Pass@$k$ results on GPQA-Diamond and HLE. Fine-tuning \texttt{Qwen3-4B-Thinking-2507} on \dataname consistently improves pass@$k$, indicating enhanced reasoning coverage and improved solution discovery under increased sampling.}
    \label{fig:pass_k}
    \vspace{-1em}
\end{figure*}

\subsection{Inference-Time Scaling Performance}

We examine whether gains from \dataname persist under inference-time scaling. Following standard practice in reasoning benchmarks~\citep{zhu_surprising_2025}, we report pass@$k$ for $k \in \{1,2,4,8\}$.

Figure~\ref{fig:pass_k} compares the base and trained models on GPQA-Diamond and HLE. The trained model consistently outperforms the base model across both benchmarks and all $k$. On GPQA-Diamond, the gap increases with $k$, reaching $90.7\%$ vs. $81.5\%$ at pass@$8$. On HLE, pass@$1$ improves from $7.3\%$ to $9.0\%$, and pass@$8$ from $19.5\%$ to $24.0\%$.
These consistent gains across larger sampling budgets indicate that improvements are not limited to single-shot accuracy, but reflect stronger reasoning robustness and broader coverage of valid solution trajectories. This aligns with the design of \dataname, which emphasizes long-horizon, multi-step reasoning and detailed solution supervision, enabling the trained model to benefit more effectively from inference-time scaling.

\section{Understanding the Effectiveness of CHIMERA}
\subsection{Effect of Subject Diversity}
A central design goal of \dataname is to provide broad and structured coverage across scientific subjects, rather than focusing only on mathematics. To examine whether this subject diversity contributes to downstream performance, we train the same base model on the math-only subset using the same SFT setup and compute budget as full \dataname training.

\begin{table*}[h]
\centering
\caption{Comparison between \texttt{Qwen3-4B-Thinking-2507} fine-tuned on the full \dataname and on its math-only subset. Full \dataname better preserves performance on broader reasoning benchmarks, while math-only training is more competitive on math-heavy benchmarks.}
\label{tab:math_only_sft}
\resizebox{\textwidth}{!}{
\begin{tabular}{lccccccc}
\toprule
Training Data & GPQA-D & AIME24 & AIME25 & AIME26 & HMMT Feb 25 & HMMT Nov 25 & HLE \\
\midrule
Full \dataname
& 68.8 & 86.5 & 79.8 & 80.3 & 63.1 & 66.3 & 9.0 \\
Math-only \dataname
& 68.1 {\small (-0.7)}
& 85.6 {\small (-0.9)}
& 80.1 {\small (+0.3)}
& 82.3 {\small (+2.0)}
& 64.3 {\small (+1.2)}
& 66.4 {\small (+0.1)}
& 8.1 {\small (-0.9)} \\
\bottomrule
\end{tabular}
}
\end{table*}

As shown in Table~\ref{tab:math_only_sft}, training on the math-only subset preserves or improves performance on math-heavy benchmarks such as AIME25, AIME26, and HMMT. This suggests that the math subset of \dataname provides strong supervision for mathematical reasoning.
However, the math-only model underperforms the full \dataname model on broader or more knowledge-intensive benchmarks, including GPQA-Diamond and HLE, which suggests that restricting training to mathematics alone can reduce generalization to broader scientific and cross-domain reasoning tasks.

These results support our design choice of constructing \dataname with diverse scientific subjects and fine-grained topics. While math-only training is effective for mathematical benchmarks, the broader subject coverage in full \dataname provides complementary training signals that help maintain performance across a wider range of reasoning tasks.

\subsection{Data Difficulty Analysis}
\label{sec:data_difficulty_analysis}

\begin{wrapfigure}{r}{0.6\textwidth}
    \centering
    \vspace{-6mm}
    \includegraphics[width=1.0\linewidth]{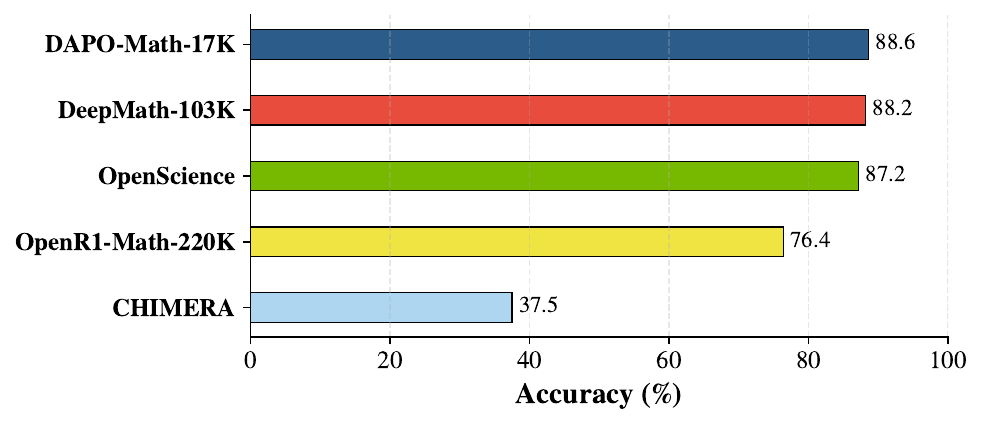}
    \vspace{-6mm}
    \caption{Accuracy of \texttt{Qwen3-4B-Thinking-2507} on existing synthetic datasets and \dataname. The base model achieves near-saturation performance on prior datasets, whereas \dataname\ remains substantially more challenging.}
    \label{fig:data_difficulty}
  \vspace{-6mm}
\end{wrapfigure}

We analyze the difficulty of existing synthetic reasoning datasets and compare them with \dataname. An effective reasoning dataset should present sufficient challenge to strong base models; otherwise, it provides limited learning signal and is unlikely to further improve reasoning capability.

To quantify dataset difficulty, we evaluate the base model \texttt{Qwen3-4B-Thinking-2507} without additional fine-tuning. We randomly sample 30K examples from OpenScience, 20K from OpenR1-Math-220K, 10K from DeepMath-103K, and use the full DAPO-Math-17K dataset, then compute the model’s solution accuracy on each. We then compare these results with the model’s performance on \dataname.

As shown in Figure~\ref{fig:data_difficulty}, the base model achieves high accuracy on existing synthetic datasets, reaching approximately $88\%$ on DAPO-Math-17K, DeepMath-103K, and OpenScience, and $76\%$ on OpenR1-Math-220K. These near-saturation results indicate that many problems in these datasets pose limited difficulty for current reasoning models.
In contrast, \dataname\ is substantially more challenging: the model achieves only $37.5\%$ accuracy, leaving considerable headroom for improvement.

Overall, these findings indicate that prior synthetic datasets may not provide sufficient difficulty to meaningfully advance current LLMs. By explicitly constructing harder, multi-step reasoning problems with longer solution trajectories, \dataname\ delivers a stronger training signal, which we find critical for further improving reasoning performance.

\subsection{Data Quality Analysis}
\label{sec:data_quality_analysis}

\begin{wrapfigure}{r}{0.5\textwidth}
  \vspace{-8mm}
    \centering
    \includegraphics[width=1.0\linewidth]{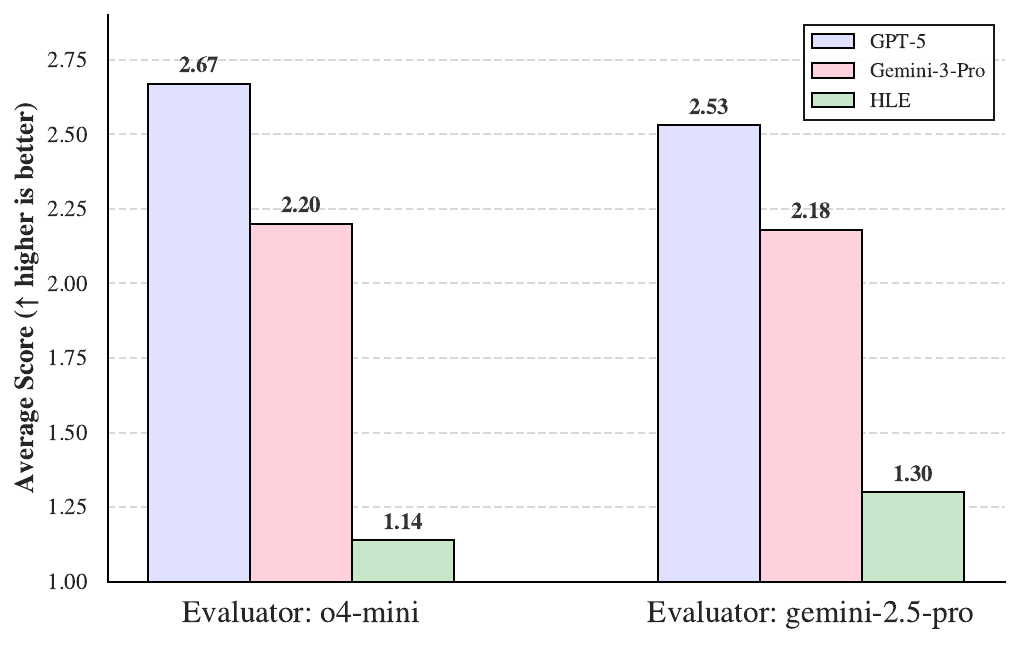}
    \caption{Average problem quality scores across sources, evaluated by \texttt{o4-mini} and \texttt{gemini-2.5-pro}. Under both evaluators, LLM-generated problems receive higher average scores than human-curated problems.}

    \label{fig:data_quality}
  \vspace{-5mm}
\end{wrapfigure}

Beyond empirical performance, we assess whether synthesized problems are comparable to human-written problems in clarity and difficulty using an LLM-as-a-Judge protocol~\citep{zheng2023llmasajudge}. We conduct a blind ranking study comparing problems from our pipeline with human-curated HLE problems.

We randomly sample 100 mathematics and 100 physics problems from HLE, and 100 mathematics and 100 physics problems from our \texttt{gpt-5}-generated training set. We also generate another 100 mathematics and 100 physics problems with \texttt{gemini-3-pro}, conditioned on the same topics as the \texttt{gpt-5} set to control for topic distribution. This gives three sources: HLE, \texttt{gpt-5}, and \texttt{gemini-3-pro}.

For each trial, we create a three-problem set with one problem per source, shuffle the order, and ask an LLM judge to rank overall quality. Scores of 3, 2, and 1 are assigned based on clarity, well-posedness, and reasoning depth, and averaged over all samples.

To reduce self-preference, we decouple generation and judging~\citep{chen2025llmselfpreference}: \texttt{gpt-5} and \texttt{gemini-3-pro} are generators, while \texttt{o4-mini} and \texttt{gemini-2.5-pro} are independent evaluators.

As shown in Figure~\ref{fig:data_quality}, both evaluators score LLM-generated problems higher on average. HLE, \texttt{gpt-5}, and \texttt{gemini-3-pro} score $1.14$, $2.67$, and $2.20$ under \texttt{o4-mini}, and $1.30$, $2.53$, and $2.18$ under \texttt{gemini-2.5-pro}. These consistent trends suggest that, under this protocol, synthetic problems are judged comparable to human-curated problems in clarity, well-posedness, and reasoning depth.

\subsection{Controlled Comparison of Problem Quality}
\label{sec:problem_quality}

To isolate the effect of problem quality, we sample an equal-sized subset of OpenThoughts-114K problems, regenerate their solutions with the same solution synthesizer as \dataname\ (\texttt{Qwen3-235B-A22B-Thinking-2507}), and train \texttt{Qwen3-4B-Thinking-2507} on both datasets in the same way.

\begin{table}[t]
\centering
\caption{Controlled comparison of problem quality on \texttt{Qwen3-4B-Thinking-2507}. We sample an equal-sized subset from OpenThoughts-114K with solutions regenerated by the same solution synthesizer as \dataname.}
\label{tab:problem_quality}
\small
\setlength{\tabcolsep}{10pt}
\resizebox{\textwidth}{!}{
\begin{tabular}{lccccccc}
\toprule
Training Data & GPQA-D & AIME24 & AIME25 & AIME26 & HMMT Feb 25 & HMMT Nov 25 & HLE \\
\midrule
Base & 65.8 & 81.6 & \textbf{81.0} & 80.8 & 59.2 & 57.3 & 7.3 \\
OpenThoughts-114K subset & 67.1 & 83.3 & 80.4 & 82.0 & 60.0 & 63.7 & 8.1 \\
\dataname & \textbf{70.1} & \textbf{86.9} & 80.7 & \textbf{82.7} & \textbf{65.7} & \textbf{67.0} & \textbf{9.0} \\
\bottomrule
\end{tabular}
}
\end{table}

As shown in Table~\ref{tab:problem_quality}, \dataname outperforms the equal-sized OpenThoughts baseline on all seven benchmarks, with an average improvement of 2.5 points. Since the dataset size, solution synthesizer, base model, and training procedure are held constant, the two settings differ only in the problem source, indicating that \dataname problems provide more effective training data than problems from existing datasets such as OpenThoughts-114K.

\section{Related Work}
\paragraph{Datasets for LLM reasoning.}
Existing reasoning datasets span mathematical problem solving~\citep{cobbe2021gsm8k,hendrycks2021math}, science and knowledge-intensive QA~\citep{clark2018arc,welbl2017sciq,mihaylov2018openbookqa}, and broad evaluation suites~\citep{srivastava2022bigbench,suzgun2023bbh,hendrycks2021mmlu}. Recent benchmarks such as GPQA~\citep{rein2024gpqa}, Humanity's Last Exam~\citep{hle2025}, and LiveBench~\citep{livebench} further emphasize high-difficulty and contamination-aware evaluation. Unlike these evaluation-oriented resources, \dataname is designed as compact, high-difficulty post-training data for improving advanced reasoning models, addressing the saturation of existing training datasets shown in Section~\ref{sec:data_difficulty_analysis}.

\paragraph{LLMs for data generation.}
LLMs are increasingly used to synthesize training data, from instruction-following datasets such as Self-Instruct~\citep{wang2023selfinstruct}, Alpaca~\citep{alpaca}, Evol-Instruct~\citep{xu2023wizardlm}, InstructZero~\citep{chen2023instructzero}, and UltraFeedback~\citep{cui2023ultrafeedback}, to reasoning datasets such as JiuZhang3.0~\citep{jiuzhang2024}, Skywork-Math~\citep{skyworkmath2024}, DeepMath-103K~\citep{deepmath103k2024}, OpenThoughts~\citep{openthoughts2024}, and OpenScience~\citep{nvidiaopenscience2024}. Unlike prior large-scale synthetic or semi-synthetic datasets, \dataname emphasizes compactness, difficulty, broad scientific coverage, and strict quality control. Our experiments show that such LLM-generated data can be rated on par with human-curated problems in clarity and reasoning depth.

\section{Conclusion}
\label{sec:conclusion}

We identify three core data-centric barriers to scalable reasoning post-training—cold-start supervision, limited domain coverage, and costly human annotation—and demonstrate that these challenges can be mitigated through carefully designed synthetic data. We introduce \dataname, a compact dataset featuring long Chain-of-Thought trajectories, broad scientific coverage, and fully automated quality control. Despite its modest size, post-training on \dataname enables a 4B model to achieve strong performance across diverse reasoning benchmarks, reaching parity with substantially larger models like DeepSeek-R1 and Qwen3-235B. Overall, our results suggest that structured, high-quality synthetic data, rather than scale alone, plays a central role in enabling effective reasoning capabilities in LLMs. We discuss potential limitations in Appendix \ref{sec:limitations}.

\subsubsection*{Acknowledgments}
We would like to thank Yuhong Li for valuable feedback and discussions during the early stage of this work.

\bibliography{example_paper}
\bibliographystyle{iclr2027_conference}

\clearpage
\appendix

\section{Additional Results on Different Model Families}
\label{app:qwen35}
To examine whether the benefits of \dataname\ generalize across different base models, we conduct SFT on three reasoning models from two model families: \texttt{Qwen3-4B-Thinking-2507}, \texttt{Qwen3.5-4B}, and \texttt{DeepSeek-R1-Distill-Llama-8B}. In all cases, we fine-tune the model on problems whose reasoning trajectories are verified as correct during solution synthesis. \texttt{DeepSeek-R1-Distill-Llama-8B} uses the same SFT configuration as \texttt{Qwen3-4B-Thinking-2507}.

\begin{table}[h]
\centering
\caption{Reasoning benchmark performance of \texttt{Qwen3-4B-Thinking-2507}, \texttt{Qwen3.5-4B}, and \texttt{DeepSeek-R1-Distill-Llama-8B}, with their corresponding SFT models trained on \dataname. SFT on \dataname improves performance on most benchmarks for the two Qwen models and on all six evaluated benchmarks for \texttt{DeepSeek-R1-Distill-Llama-8B}.}

\resizebox{\textwidth}{!}{
\begin{tabular}{lccccccc}
\toprule
\textbf{Model} &
\textbf{GPQA-D} &
\textbf{AIME24} &
\textbf{AIME25} &
\textbf{AIME26} &
\textbf{HMMT Feb 25} &
\textbf{HMMT Nov 25} &
\textbf{HLE} \\
\midrule

\multicolumn{8}{c}{\textit{\textbf{Qwen3-4B-Thinking-2507}}} \\
\midrule
Qwen3-4B-Thinking-2507
& 65.8 & 81.6 & \textbf{81.0} & \textbf{80.8} & 59.2 & 57.3 & 7.3 \\

Qwen3-4B-Thinking-2507 + \dataname\ (SFT)
& \textbf{68.8} & \textbf{86.5} & 79.8 & 80.3 & \textbf{63.1} & \textbf{66.3} & \textbf{9.0} \\

\midrule
\multicolumn{8}{c}{\textit{\textbf{Qwen3.5-4B}}} \\
\midrule
Qwen3.5-4B
& 76.2 & 86.5 & 80.4 & 85.4 & \textbf{74.0} & \textbf{76.8} & \textbf{12.9} \\

Qwen3.5-4B + \dataname\ (SFT)
& \textbf{79.7} & \textbf{88.0} & \textbf{83.1} & \textbf{86.3} & 72.8 & 74.5 & 12.7 \\

\midrule
\multicolumn{8}{c}{\textit{\textbf{DeepSeek-R1-Distill-Llama-8B}}} \\
\midrule
DeepSeek-R1-Distill-Llama-8B
& 49.0 & 50.1 & 36.1 & 39.5 & 26.8 & 25.7 & -- \\

DeepSeek-R1-Distill-Llama-8B + \dataname\ (SFT)
& \textbf{49.5} & \textbf{56.7} & \textbf{42.5} & \textbf{47.9} & \textbf{30.4} & \textbf{36.2} & -- \\

\bottomrule
\end{tabular}
}
\label{tab:sft_vs_sft_rl}
\end{table}
As shown in Table~\ref{tab:sft_vs_sft_rl}, SFT on \dataname\ improves performance on most benchmarks for all three base models. For \texttt{Qwen3-4B-Thinking-2507}, SFT brings clear gains on GPQA-Diamond (+3.0), AIME24 (+4.9), HMMT Feb 25 (+3.9), HMMT Nov 25 (+9.0), and HLE (+1.7). For \texttt{Qwen3.5-4B}, SFT improves GPQA-Diamond (+3.5), AIME24 (+1.5), AIME25 (+2.7), and AIME26 (+0.9). For \texttt{DeepSeek-R1-Distill-Llama-8B}, SFT improves all six evaluated benchmarks: AIME24 (+6.6), AIME25 (+6.4), AIME26 (+8.4), HMMT Feb 25 (+3.6), HMMT Nov 25 (+10.5), and GPQA-Diamond (+0.5).

\texttt{DeepSeek-R1-Distill-Llama-8B} differs from our main model in three respects: it belongs to a non-Qwen model family, it is twice the size of \texttt{Qwen3-4B-Thinking-2507}, and it has already undergone extensive reasoning post-training and distillation. The gains on this model therefore show that \dataname\ still benefits a model already exposed to strong-teacher reasoning signals. DeepMath-103K and OpenScience are also generated by strong teachers, DeepSeek-R1 and Qwen3-235B respectively, but they do not improve \texttt{Qwen3-4B-Thinking-2507} despite being much larger than \dataname\ (Table~\ref{tab:main-results}). Teacher strength alone therefore cannot explain the gains of \dataname.

Overall, these results show that the benefits of \dataname\ are not limited to a single base model, model family, or model scale. In addition to improving \texttt{Qwen3-4B-Thinking-2507}, SFT on \dataname\ also yields gains on \texttt{Qwen3.5-4B}, a recently released and substantially stronger 4B-scale reasoning model, and on the Llama-based \texttt{DeepSeek-R1-Distill-Llama-8B}. Together, these results provide evidence across two model families and two model scales, suggesting that our dataset remains useful even for strong, up-to-date reasoning models.

\section{Data Contamination Analysis}
\label{app:decontam}

To verify that gains are not due to data leakage, we measure lexical overlap between \dataname and GPQA-Diamond/HLE using the $n$-gram similarity score from prior work~\citep{gpt3,llama2}. Let $\mathcal{T}$ be the synthetic training questions and $\mathcal{S}$ the test questions. For each $t_i \in \mathcal{T}$, we compute its maximum $n$-gram Jaccard similarity to any test question and average over $\mathcal{T}$:
$\text{Score}_n =
|\mathcal{T}|^{-1}
\sum_{i=1}^{|\mathcal{T}|}
\max_{s \in \mathcal{S}}
\frac{|G_n(t_i) \cap G_n(s)|}{|G_n(t_i) \cup G_n(s)|}$,
where $G_n(\cdot)$ denotes the set of distinct $n$-grams. Lower scores indicate weaker overlap and lower contamination risk.

\begin{table}[h]
\centering
\caption{$n$-gram Jaccard overlap between \dataname and test benchmarks.}
\small
\label{tab:decontam}
\begin{tabular}{lcc}
\toprule
\textbf{Benchmark} & \textbf{8-gram} & \textbf{13-gram} \\
\midrule
GPQA-Diamond & $0$ & $0$ \\
HLE & $7.89 \times 10^{-6}$ & $0$  \\
\bottomrule
\end{tabular}
\end{table}

We report stricter $8$-gram and $13$-gram overlaps for near-duplicate detection. As shown in Table~\ref{tab:decontam}, overlaps are effectively zero: GPQA-Diamond has $0$ for both, while HLE has only $7.89 \times 10^{-6}$ for $8$-grams and $0$ for $13$-grams. These near-zero scores indicate no meaningful lexical overlap between \dataname and the evaluation benchmarks, suggesting that the observed gains are unlikely to come from contamination.

\section{Statistical Significance of Main Results}
\label{app:significance}
We conduct significance tests for the gains of \texttt{Qwen3-4B-Thinking-2507} fine-tuned on \dataname\ over the base model reported in Table~\ref{tab:main-results}. As shown in Table~\ref{tab:significance}, the gains are statistically significant on five of the seven benchmarks. The differences on AIME25 and AIME26 are not significant, as the base model already achieves strong performance on these two benchmarks.

\begin{table}[h]
\centering
\caption{Significance tests for the gains of \texttt{Qwen3-4B-Thinking-2507} fine-tuned on \dataname\ over the base model in Table~\ref{tab:main-results}. $\Delta$ denotes the difference in pass@1 relative to the base model.}
\label{tab:significance}
\begin{tabular}{lcccc}
\toprule
\textbf{Benchmark} & $\boldsymbol{\Delta}$ & \textbf{95\% CI} & \textbf{$p$-value} & \textbf{Significant} \\
\midrule
AIME24 & $+5.3$ & $[+2.1, +7.9]$ & $0.0003$ & Yes \\
AIME25 & $-0.3$ & $[-3.1, +1.7]$ & $0.882$ & No \\
AIME26 & $+1.8$ & $[-4.3, +8.1]$ & $0.608$ & No \\
HMMT Feb 25 & $+6.7$ & $[+2.1, +12.0]$ & $0.008$ & Yes \\
HMMT Nov 25 & $+9.7$ & $[+1.5, +18.9]$ & $0.042$ & Yes \\
GPQA-Diamond & $+4.3$ & $[+0.6, +7.1]$ & $0.022$ & Yes \\
HLE & $+1.7$ & $[+1.2, +2.5]$ & $<0.0001$ & Yes \\
\bottomrule
\end{tabular}
\end{table}

\section{Limitations}
\label{sec:limitations}
Our study has several limitations: 1. \dataname is fully synthetic and relies on strong LLMs for topic expansion, problem generation, solution synthesis, and verification. Although we use cross-model validation and decontamination analysis to reduce errors and leakage, automated judges can still share biases with the generators or miss subtle flaws in highly specialized domains; 2. We do not include large-scale human evaluation of the generated problems or reasoning trajectories, but this is also a practical challenge: many examples are designed to require advanced, domain-specific reasoning, making reliable human validation expensive, slow, and difficult to scale across all covered subjects; 3. The pipeline relies on capable proprietary and open reasoning models. In our implementation, API calls for data generation and verification cost approximately \$2,000, in addition to the compute required for open-model inference and post-training.

\clearpage
\section{Prompts}
\label{appendix:prompts}
\begin{table}[H]
\centering
\caption{Prompts used for the data synthesis pipeline.}
\setlength{\aboverulesep}{0.2ex}\setlength{\belowrulesep}{0.2ex}
\begin{tabular}{p{0.2\linewidth} > {\small}p{0.72\linewidth}}
\toprule
\textbf{Usage} & \textbf{Prompt Template} \\
\midrule
Subject Expansion & I have a broad subject. Generate a list of more specific topics that fall under this subject. Each topic should be unique and different from the other topics. The topics should be an important field of study within this subject. Here is the broad subject: \{subject\}. Be as specific and diverse as possible when listing topics. For example, subject ``Probability Theory'' could have topics like ``Probability Distributions'', ``Random Variable'', ``Conditional Probability'', ``Bayesian Statistics'', etc.
Output should be in the format of a list of topics: \verb|<topic1>\n<topic2>\n<topic3>\n|... \\
\midrule

Problem Generation & Given a topic, generate a new problem that falls under this topic. The problem should be at a PhD level or higher. Here is the topic: \{topic\}.

Your answer must strictly follow the format and criteria below, without any markdown formatting:

Problem: A clearly stated problem that requires advanced knowledge, techniques, or reasoning related to the topic. The problem must have a single, unambiguous, and objectively verifiable answer (e.g., a specific numeric value, closed-form expression, proof, or well-defined result). Clearly define all variables, parameters, and assumptions to avoid ambiguity.

Reasoning: A detailed, logical reasoning process that shows how to solve the problem step by step, ensuring each step is verifiable and reproducible.

Solution: A complete and rigorous solution derived from the reasoning steps above, including all necessary calculations or proofs.

Answer: The final, concise answer to the problem. It must be clearly checkable and easy to verify as correct. \\
\midrule
Solution Synthesis & \texttt{<|im\_start|>|}user\textbackslash n\{problem\}\texttt{<|im\_end|>}\textbackslash n \\
\midrule
Problem Validator & Given a topic and a problem with the answer, verify whether the problem and answer are valid and satisfy the requirements.

Topic: \{topic\}

Problem: \{problem\}

Answer: \{answer\}

Check the following:

- The problem matches the topic and is at a PhD level or higher.
- The statement is clear, self-contained, and unambiguous.
- There is exactly one objectively verifiable answer.
- The final answer is correct.

Your answer must strictly follow the format and criteria below, without any markdown formatting:

Verdict: VALID or INVALID

Reason: one short sentence explaining the decision \\
\midrule

Correctness Verifier & Judge whether the following predicted\_solution to question is correct or not based on the precise and unambiguous correct\_answer below.

question: \{question\}

correct\_answer: \{correct\_answer\}

predicted\_solution: \{predicted\_solution\}

Your judgement must be in the format and criteria specified below:

extracted\_final\_answer: The final exact answer extracted from the predicted\_solution. Put the extracted answer as 'None' if there is no exact, final answer to extract from the predicted solution.

reasoning: Explain why the extracted\_final\_answer is correct or incorrect based on correct\_answer, focusing only on if there are meaningful differences between correct\_answer and the extracted\_final\_answer. Do not comment on any background to the problem, do not attempt to solve the problem, do not argue for any answer different than correct\_answer, focus only on whether the answers match.

correct: Answer 'yes' if extracted\_final\_answer matches the correct\_answer given above, or is within a small margin of error for numerical problems. Answer 'no' otherwise, i.e. if there is any inconsistency, ambiguity, non-equivalency, or if the extracted answer is incorrect.

confidence: The extracted confidence score between 0\% and 100\% from predicted\_solution. Put 100 if there is no confidence score available. \\
\bottomrule
\end{tabular}
\label{tab:prompt}
\end{table}

\clearpage

\section{Subjects and Topics}
\label{appendix:sub2topic}
\begin{longtable}{ p{0.18\linewidth} >{\small}p{0.74\linewidth} }
\caption{Subjects and their expanded topic sets used for creating \dataname.}
\label{tab:sub2topic} \\
\toprule
\textbf{Subject} & \textbf{Topic} \\
\midrule
\endfirsthead

\toprule
\textbf{Subject} & \textbf{Topic} \\
\midrule
\endhead

\bottomrule
\endfoot

\bottomrule
\endlastfoot

Mathematics
& `3-manifolds and hyperbolic structures', `4-manifolds and smooth structures', `Abelian varieties, Jacobians, and Pryms', `Absoluteness and reflection principles', `Actuarial science and risk theory', `Additive combinatorics', `Additive combinatorics and sum-product phenomena', `Additive number theory and sumset problems', `Adic and perfectoid spaces', `Algebra', `Algebraic coding theory', `Algebraic combinatorics', `Algebraic cycles and Chow groups', `Algebraic geometry and moduli', `Algebraic geometry and tropical combinatorics', `Algebraic number theory', `Algebraic spaces and stacks', `Algebraic topology', `Algebraic topology and homotopy theory', `Algorithmic class field theory', `Algorithms and randomized algorithms', `Analysis', `Analytic combinatorics and singularity analysis', `Analytic number theory', `Applications to analysis and Banach spaces' \\

& `Applied combinatorics in networks and data', `Applied probability', `Applied topology and persistent homology', `Approximate counting and correlation decay', `Approximation algorithms for graph problems', `Approximation theory and splines', `Arithmetic combinatorics', `Arithmetic geometry and Arakelov theory', `Arithmetic geometry and Diophantine geometry', `Arithmetic statistics', `Arithmetic statistics and Cohen–Lenstra heuristics', `Association schemes and distance-regular graphs', `Automated reasoning and theorem proving', `Automated theorem proving for mathematics', `Automorphic forms and the Langlands program', `Automorphism groups and Cayley graphs', `Axiomatic frameworks and foundations of mathematics', `Axiomatic set theory (ZFC and alternatives)', `BSD conjecture and L-series of elliptic curves', `Backward SDE and FBSDE', `Bandits and online learning', `Bayesian nonparametrics and Dirichlet processes', `Bayesian statistics and Bayesian nonparametrics', `Bijective combinatorics', `Bioinformatics and phylogenetics' \\

& `Biophysics and neuroscience models', `Biostatistics and clinical trials', `Bipartite and multipartite graphs', `Birational geometry and minimal model program', `Blockchain primitives and number-theoretic protocols', `Borel equivalence relations and classification theory', `Branching processes and coalescent theory', `Brauer groups and central simple algebras', `Bridgeland stability conditions', `Brill–Noether theory and syzygies', `Brownian motion and Gaussian processes', `CAT(0) and Alexandrov spaces', `Calabi–Yau manifolds and special holonomy', `Calabi–Yau varieties and K3 surfaces', `Calculus of variations', `Calibrations and special Lagrangians', `Cardinal characteristics of the continuum', `Category theory and topos theory', `Causal inference under uncertainty', `Chabauty–Coleman methods and p-adic integration', `Chaos and bifurcation theory', `Character sheaves and Springer theory', `Cheeger inequalities and isoperimetry on graphs', `Chip-firing games and sandpile groups', `Chordal, interval, and comparability graphs' \\

& `Chromatic invariants and Tutte polynomials', `Class field theory', `Classical algebraic geometry (varieties and maps)', `Classifying spaces and characteristic classes', `Clique-width and rank-width', `CoHAs and categorification in topology', `Coarse and controlled topology', `Cobordism theory and Thom spectra', `Coding theory and combinatorial designs', `Coding theory and compressed sensing', `Coding theory and number-theoretic constructions', `Combinatorial commutative algebra', `Combinatorial optimization', `Combinatorial representation theory', `Combinatorial set theory', `Combinatorics for ML and graph neural networks', `Combinatorics in statistical physics (Ising, Potts, dimers)', `Combinatorics of polynomials and real stable polynomials', `Commutative algebra', `Comparison geometry and Ricci curvature', `Complex and Kähler geometry', `Computability and recursion theory', `Computational algebra and Gröbner bases', `Computational algebraic geometry and Gröbner bases', `Computational and algorithmic number theory' \\

& `Computational and applied mathematics', `Computational geometry', `Concentration of measure and isoperimetry', `Conformal and CR geometry', `Conformal and topological quantum field theories', `Conformal prediction and uncertainty quantification in ML', `Constructible universe L and fine structure', `Constructive and intuitionistic mathematics', `Contact geometry and Reeb dynamics', `Container method and hypergraph containers', `Continuum theory and Peano continua', `Control and observability on networks', `Control theory and systems theory', `Convex geometry and Brunn–Minkowski theory', `Core model theory and extender models', `Coupling methods and contractivity', `Covering spaces and fundamental group', `Cryptography and algebraic cryptography', `Cryptography and post-quantum cryptography', `Data science and manifold learning', `Decorated character varieties and higher Teichmüller theory', `Deep learning theory and generalization', `Deligne–Mumford and Artin stacks', `Derived categories of coherent sheaves', `Descriptive set theory' \\

& `Descriptive set theory (Borel and projective hierarchies)', `Descriptive set-theoretic dynamics', `Design theory and block designs', `Design theory and finite geometries', `Determinacy axioms (AD, ADL(R))', `Differential geometry', `Diophantine approximation and transcendence', `Diophantine geometry and arithmetic geometry', `Discrete Morse theory', `Discrete and computational geometry', `Discrete conformal geometry and circle packings', `Discrete geometry and incidence geometry', `Discrete geometry and polyhedral combinatorics', `Discrete mathematics', `Disordered systems and spin glasses', `Distance-regular graphs and association schemes', `Distributed and local graph algorithms', `Distribution of rational points and unlikely intersections', `Donaldson and Seiberg–Witten theory', `Donaldson–Thomas and Pandharipande–Thomas invariants', `Econophysics and financial mathematics', `Edge coloring and total coloring', `Edge expansion, vertex expansion, and conductance', `Effective descriptive set theory', `Electrical networks and effective resistance' \\

& `Elementary number theory', `Elliptic curve methods and isogeny computations', `Elliptic curves and abelian varieties', `Embeddings and metric distortion', `Emerging directions', `Empirical processes and Donsker theorems', `Empirical processes and concentration bounds', `Enumerative and extremal combinatorics', `Enumerative combinatorics and generating functions', `Enumerative geometry and Hurwitz theory', `Environmental and climate risk modeling', `Epidemic models and branching processes', `Epidemics and contagion on networks', `Epidemiology and infectious disease modeling', `Erdos distinct distances and Szemerédi–Trotter', `Erdos–Ko–Rado and intersecting families', `Ergodic methods on homogeneous spaces', `Ergodic theory and dynamical systems', `Ergodic theory and mixing', `Euclidean and classical geometry', `Exchangeability and de Finetti theorems', `Expander constructions and pseudorandomness', `Expander constructions via number theory', `Expanders and expander constructions', `Exponential sums and the circle method' \\

& `Extremal combinatorics', `Extreme value theory and heavy tails', `Extreme value theory and regular variation', `Face numbers, f-vectors, and h-vectors', `Facility location and scheduling', `Fairness and privacy in algorithms', `Fairness and robustness via probabilistic modeling', `Faltings’ theorem and Mordell-type problems', `Fano varieties and rationality problems', `Federated Bayesian learning and distributed inference', `Fibre bundles and principal bundles', `Field theory and Galois theory', `Filtering and state estimation (Kalman, particle filters)', `Financial mathematics and derivatives pricing', `Fine-grained lower bounds for combinatorial problems', `Finite element and spectral methods', `Finite geometries and incidence structures', `Floer homology and pseudo-holomorphic curves', `Flows, cuts, and circulations', `Flows, cuts, and sparsifiers', `Foliations, laminations, and measured laminations', `Forcing and iterated forcing', `Formal methods in cyber-physical systems', `Fractal geometry', `Fractional coloring and circular coloring' \\

& `Free probability and noncommutative probability', `Free probability in applications', `Functional CLTs and invariance principles', `Functional analysis and operator theory', `Functional inequalities (Poincaré, log-Sobolev, transport)', `Galois representations and deformation theory', `Galois theory of number fields', `Game theory and mechanism design', `Gaussian processes and kernel methods', `General relativity and geometric PDE', `Generalized metric spaces and compactness', `Generators, Dirichlet forms, and potential theory', `Generic absoluteness and forcing axioms (MA, PFA, MM)', `Geometric Langlands program', `Geometric analysis', `Geometric analysis and PDE on manifolds', `Geometric deep learning and graph neural networks', `Geometric graphs and unit disk graphs', `Geometric group theory', `Geometric group theory and quasi-isometries', `Geometric measure theory', `Geometric measure theory and currents', `Geometric set cover and epsilon-nets', `Geometric topology', `Geometric topology interfaces' \\

& `Geometrization and 3-manifold geometry', `Geometry and topology', `Geometry of Banach spaces', `Geometry of numbers and lattice point problems', `Gibbs measures, spin systems, and phase transitions', `Graph coloring and list coloring', `Graph compression and summarization', `Graph databases and knowledge graphs', `Graph decomposition and minors structure', `Graph drawing and visualization', `Graph embeddings and metric embeddings', `Graph homomorphisms and constraint satisfaction', `Graph isomorphism and canonical labeling', `Graph limits and graphons', `Graph minors and Robertson–Seymour theory', `Graph polynomials (Tutte, chromatic, interlace)', `Graph products and powers', `Graph representation learning and GNNs', `Graph theory and spectral graph theory', `Graph theory and structural graph theory', `Greedoid theory', `Gromov–Witten theory and quantum cohomology', `Group theory', `Hamiltonian systems and symplectic dynamics', `Hardness of approximation and PCP connections' \\

& `Harmonic analysis and Fourier analysis', `Heegaard Floer homology and contact invariants', `Heights, Arakelov geometry, and Northcott property', `High-dimensional probability and concentration inequalities', `High-dimensional probability for data science', `Hilbert schemes and Quot schemes', `Hodge theory', `Hodge theory and period mappings', `Hodge theory and variations of Hodge structure', `Holographic and Holant frameworks', `Homological algebra', `Homological mirror symmetry', `Homology, cohomology, and cohomology operations', `Homotopical algebra and model categories', `Homotopy theory and Postnikov towers', `Homotopy type theory in proof assistants', `Hopf algebras and quantum groups', `Hydrology and extreme events', `Hyperbolic geometry and Kleinian groups', `Hypergraph theory', `Hypergraph theory and set systems', `Hypergraphs and uniformity', `Ideals on omega and forcing with ideals', `Imaging science and tomography', `Index theory and Atiyah–Singer theorem' \\

& `Infinity categories and higher topos theory', `Information geometry and statistical manifolds', `Information geometry for learning', `Information theory and entropy methods', `Information-theoretic inequalities and entropy methods', `Inner model theory and mice', `Insurance mathematics and solvency modeling', `Integrable probability and last passage percolation', `Integrable systems', `Integral geometry and geometric probability', `Integral polyhedra and total unimodularity', `Interacting particle systems and hydrodynamic limits', `Interacting particle systems and mean-field models', `Interactive proof assistants and formalization of mathematics', `Interdisciplinary and applied areas', `Intersection theory and Poincaré duality', `Intersection theory and Riemann–Roch', `Invariant theory', `Inverse problems and regularization', `Iwasawa theory', `K-theory (algebraic and topological)', `K-theory (topological and algebraic)', `KPZ universality and integrable probability', `Kazhdan–Lusztig theory and Hecke algebras', `Knot theory and link invariants' \\

& `Knot theory and low-dimensional topology', `L-functions and functional equations', `L-functions and zeta functions', `Langlands program (classical and geometric)', `Large cardinal axioms', `Large cardinals and elementary embeddings', `Large deviations and Gartner–Ellis theorem', `Latin squares and orthogonal arrays', `Lattice algorithms and LLL', `Levy processes and jump diffusions', `Lie groups and Lie algebras', `Linear systems and base loci', `Liouville quantum gravity and random geometry', `Local fields and ramification', `Local limit theorems and Berry–Esseen bounds', `Logarithmic geometry', `Low-dimensional topology', `Manifold topology and smoothing theory', `Manin conjecture and Peyre constants', `Mapping class groups and surfaces', `Markov chain Monte Carlo and mixing analysis', `Markov chains and Markov processes', `Markov processes and semigroup theory', `Martingales and optional stopping', `Matching theory and matroid intersection' \\

& `Matching, matroid parity, and factors', `Mathematical biology and epidemiology', `Mathematical physics', `Matroid theory and oriented matroids', `Mean curvature flow and Ricci flow', `Mean field games', `Mean-field models and McKean–Vlasov equations', `Measurable cardinals and ultrafilters', `Measure theory and integration', `Measure-theoretic probability and integration', `Measure-theoretic probability and stochastic processes', `Metric geometry and Gromov hyperbolicity', `Microlocal analysis and pseudodifferential operators', `Minimal surfaces and geometric flows', `Minimal surfaces and mean curvature flow', `Mirror symmetry and Fukaya categories', `Mirror symmetry and SYZ', `Mixed Hodge structures and degenerations', `Model theory', `Model theory in number theory and o-minimality', `Moderate deviations and LIL', `Modular forms and automorphic forms', `Moduli of curves and stable maps', `Moduli problems and geometric invariant theory', `Monte Carlo and Markov chain mixing' \\

& `Monte Carlo methods and variance reduction', `Motives and motivic cohomology', `Multiplier ideals and singular metrics', `Network design and survivable networks', `Network design, reliability, and survivability', `Network inference and community detection', `Network science and complex systems', `Network science and multilayer networks', `Neural differential equations and neural operators', `Non-Euclidean geometry (hyperbolic and spherical)', `Noncommutative algebra', `Noncommutative algebraic geometry', `Noncommutative topology and C*-algebraic topology', `Nonlinear analysis', `Nonstandard analysis', `Number theory', `Numerical analysis and scientific computing', `Numerical linear algebra', `O-minimality and definable sets in AG', `Obstruction theory', `Operations research and mathematical programming', `Operations research under uncertainty', `Operator algebras (C*-algebras, von Neumann algebras)', `Optimal experimental design and active learning', `Optimal transport and Wasserstein geometry' \\

& `Optimal transport and probability metrics', `Optimal transport for probabilistic modeling', `Optimization (convex, nonconvex, integer, semidefinite)', `Orbifolds and stacks', `Order dimension and comparability graphs', `PCF theory and singular cardinals', `Parameterized algorithms and kernels on graphs', `Parameterized complexity and kernelization', `Partial differential equations (elliptic, parabolic, hyperbolic)', `Partition calculus and infinite combinatorics', `Percolation and interacting particle systems', `Percolation and phase transitions', `Percolation and phase transitions on networks', `Perfect graphs and the strong perfect graph theorem', `Perfectoid spaces and p-adic Hodge theory', `Permutation patterns and sorting networks', `Perpetuities and fixed-point equations in distribution', `Perverse sheaves and D-modules', `Perverse sheaves and microlocal sheaf theory', `Philosophy and history of mathematics', `Physics-informed machine learning and scientific ML', `Planar graphs and graphs on surfaces', `Point-set topology and separation axioms', `Poisson geometry', `Polyhedral combinatorics' \\

& `Polyhedral geometry and polytopes', `Polytopes, zonotopes, and associahedra', `Posets, lattices, and Möbius inversion', `Positivity of line bundles and vanishing theorems', `Post-quantum cryptography and isogeny-based crypto', `Primality testing and integer factorization', `Prime number theory and zero-density estimates', `Privacy-preserving analytics and differential privacy', `Probabilistic combinatorics', `Probabilistic combinatorics and random structures', `Probabilistic numerics', `Probabilistic potential theory', `Probabilistic programming and approximate inference', `Probability and statistics', `Projective and affine geometry', `Proof theory', `Propagation of chaos and chaos decompositions', `Property testing and sublinear algorithms', `Property testing for graph properties', `Pseudo-Riemannian and Lorentzian geometry', `Pseudorandomness and extractor constructions', `Quadratic forms and modular curves', `Quantum field theory (constructive, algebraic, axiomatic)', `Quantum gravity and string theory', `Quantum information theory and quantum error correction' \\

& `Quantum mechanics and spectral analysis', `Quantum unique ergodicity and arithmetic quantum chaos', `Queueing theory and network queues', `Quiver varieties and symplectic resolutions', `Ramsey theory', `Ramsey theory and van der Waerden-type problems', `Random Schrödinger operators', `Random Schrödinger operators and Anderson localization', `Random fields and Gaussian free fields', `Random geometry and SLE in applications', `Random graphs and complex networks', `Random graphs and hypergraphs', `Random graphs and percolation theory', `Random graphs and stochastic block models', `Random matrices and free probability', `Random matrices and spectral statistics', `Random matrices and universality', `Random matrix theory and L-function statistics', `Random multiplicative functions', `Random multiplicative functions and pretentious methods', `Random permutations and descents', `Random tilings and dimer models', `Random walk in random environment', `Random walks and diffusion on graphs', `Random walks on groups and heat kernel bounds' \\

& `Random walks, mixing times, and cover times', `Random, Cohen, and Sacks forcing', `Rational connectedness and Campana program', `Rational homotopy theory', `Rational points on higher-dimensional varieties', `Real and complex analysis', `Reinforcement learning with probabilistic models', `Reliability engineering and maintenance', `Reliability theory and survival analysis', `Representation theory', `Reverse mathematics', `Ricci flow and comparison geometry', `Riemannian and pseudo-Riemannian geometry', `Riemannian geometry', `Rigid analytic and Berkovich spaces', `Ring theory and module theory', `Risk-sensitive and robust decision-making', `Rough paths and regularity structures', `Sato–Tate equidistribution and automorphic Sato–Tate', `Schemes and morphisms of schemes', `Schmidt subspace theorem and Roth’s theorem', `Schramm–Loewner evolution and conformal invariance', `Schubert calculus and Coxeter combinatorics', `Semimartingales and stochastic integration', `Sequential Monte Carlo and particle methods' \\

& `Set theory and foundations', `Set-theoretic topology and selection principles', `Shape theory and ANR spaces', `Sheaf theory and cohomology', `Shimura varieties and automorphic sheaves', `Shimura varieties and modularity', `Sieve methods and additive number theory', `Sieve methods and distribution in arithmetic progressions', `Signal processing and compressed sensing', `Simplicial complexes and collapsibility', `Singularity theory and resolutions', `Spatial statistics and geostatistics', `Spatial-temporal modeling and point processes', `Spectral geometry and isoperimetric inequalities', `Spectral graph theory and Laplacians', `Spectral graph theory and expanders', `Spectral hypergraph theory', `Spectral sequences and derived functors', `Spectral sparsification', `Spectral theory', `Sperner theory and lattice of subsets', `Spin geometry and Dirac operators', `Square and diamond principles', `Stable homotopy and spectra', `Stable homotopy category and spectra' \\

& `Stable laws and domains of attraction', `Stable pairs and wall crossing', `Stanley–Reisner theory and face rings', `Stationary sets and club guessing', `Statistical decision theory and inference', `Statistical mechanics and phase transitions', `Statistical mechanics and probabilistic combinatorics', `Steiner systems and finite projective planes', `Stein’s method and normal approximation', `Stochastic PDE and random fields', `Stochastic PDE in applications', `Stochastic analysis on manifolds', `Stochastic calculus (Itô and Stratonovich)', `Stochastic calculus and SDE', `Stochastic control and optimal stopping', `Stochastic control, HJB, and viscosity solutions', `Stochastic differential equations and flows', `Stochastic networks and communication systems', `Stochastic optimization and stochastic gradient methods', `Stochastic partial differential equations', `Stratified spaces and intersection homology', `Streaming algorithms for combinatorial problems', `Structural graph theory', `Sub-Riemannian and Finsler geometry', `Sublinear and streaming graph algorithms' \\

& `Submodular optimization', `Sumset inequalities and Freiman-type results', `Surface classification and higher-dimensional birational geometry', `Suslin trees and Aronszajn trees', `Symplectic and contact geometry', `Symplectic and contact topology', `Symplectic geometry and Hamiltonian dynamics', `Systems biology and network biology', `TQFT and modular tensor categories', `Teichmüller theory and moduli of curves', `Teichmüller theory and moduli spaces', `Temporal and dynamic graphs', `Theoretical computer science and complexity', `Theta functions and moduli of vector bundles', `Threshold phenomena and sharp thresholds', `Tilings, packings, and sphere packing', `Time series analysis and stochastic forecasting', `Time-frequency analysis and wavelets', `Topological combinatorics', `Topological data analysis', `Topological data analysis and persistent homology', `Topological dynamics and minimal sets', `Topological graph theory and crossing number', `Topological groups and Lie groups', `Topological methods in combinatorics' \\

& `Topological methods in data analysis', `Topological quantum computing', `Topological robotics and configuration spaces', `Topology (point-set, geometric, and algebraic)', `Toric and tropical geometry', `Toric varieties and fans', `Tournament and oriented graphs', `Transcendence and Diophantine approximation', `Treewidth, clique-width, and decompositions', `Treewidth, pathwidth, and decompositions', `Tropical geometry and tropical intersection', `Type theory and homotopy type theory', `Uncertainty quantification', `VC theory and uniform laws of large numbers', `Variational inequalities and complementarity problems', `Waring’s problem and Vinogradov methods', `Weak convergence and Skorokhod topology', `Young tableaux and Schur functions', `Zero-free regions and the Riemann hypothesis', `Zeta functions of graphs', `p-adic Hodge theory and Fontaine modules', `p-adic analysis and Hida families', `p-adic number theory and local fields', `q-series, partitions, and symmetric functions' \\

Computer Science
& `AI for systems and systems for AI', `AR, VR, and XR interfaces', `Accessibility and inclusive design', `Active learning and optimal data selection', `Algorithms and data structures', `Animation and physics-based simulation', `Approximation algorithms and PTAS', `AutoML and hyperparameter optimization', `Bayesian deep learning and uncertainty estimation', `Bilevel optimization and hyperparameter tuning', `Binary analysis and software hardening', `Blockchain and distributed ledger technologies', `Category theory in CS', `Causal inference and causal discovery in CS', `Causal representation learning', `Causality in ML and causal discovery', `Cloud computing and virtualization', `Computability and automata theory', `Computational biology and bioinformatics', `Computational chemistry and materials informatics', `Computational complexity theory', `Computational economics and market design', `Computational imaging and computational photography', `Computational neuroscience', `Computer architecture and hardware-software co-design' \\

& `Computer graphics and rendering', `Computer networks and programmable data planes', `Computer vision and multimodal learning', `Computer vision and pattern recognition', `Computer-supported cooperative work', `Convolutional, recurrent, and attention-based networks', `Cryptography and cryptographic protocols', `Cyber-physical systems and real-time computing', `Data mining and knowledge discovery', `Data-centric AI and data governance', `Data-centric AI, data governance, and quality', `Database systems and data engineering', `Datacenter architectures and disaggregated systems', `Deep learning and representation learning', `Deep learning architectures', `Differential privacy and federated learning', `Differential privacy and secure computation', `Digital currencies and cryptoeconomics', `Digital twins and cyber-physical security', `Distributed systems and consensus', `Domain generalization and distribution shift', `Edge AI and on-device learning', `Edge computing and IoT', `Education technologies and computing education research', `Embedded systems and robotics platforms' \\

& `Energy-aware and green computing', `Energy-based models and score matching', `Fairness, accountability, and ethics', `Fairness, accountability, and transparency in AI', `Federated and decentralized learning', `Formal languages and parsing', `Formal methods and model checking', `Foundation models and large language models', `GPU and accelerator computing', `Generative chemistry and materials discovery', `Generative modeling', `Geometric and topological data analysis', `Geometry processing and CAD', `Graph algorithms and network algorithms', `Graph and relational learning for science', `Graph neural networks and geometric deep learning', `Hierarchical and multi-agent RL', `Human-AI interaction and interactive ML', `Human-computer interaction and UX', `Implicit neural representations and neural fields', `In-memory and approximate computing', `Information retrieval and search', `Information-theoretic learning bounds', `Instruction tuning, RLHF, and RLAIF', `Interpretability and explainability' \\

& `Inverse RL and imitation learning', `Knowledge representation and reasoning', `Lambda calculus and functional programming theory', `Large language models and foundation models', `Learning PDE solvers and neural operators', `Logic in computer science', `MLOps and ML systems', `MLOps, deployment, and monitoring', `Machine learning and data-driven methods', `Mechanistic interpretability of neural networks', `Meta-learning and few-shot learning', `Model compression, pruning, quantization, distillation', `Multi-task and continual learning', `Multiagent systems and game-theoretic AI', `Natural language processing and LLMs', `Natural language processing and computational linguistics', `Networking for AI workloads', `Neural architecture search and AutoML', `Neurosymbolic AI and program synthesis', `Nonconvex optimization and landscape analysis', `Offline and batch RL', `Online algorithms and competitive analysis', `Online learning and bandits', `Operating systems and kernels', `Optimization for ML' \\

& `PAC learning and VC theory', `Parallel and high-performance computing', `Parameterized and fine-grained complexity', `Perception, SLAM, and control in robotics', `Physics-informed ML and operator learning', `Planning, search, and constraint programming', `Policy gradients and actor-critic methods', `Post-Moore architectures and neuromorphic computing', `Post-quantum cryptography', `Privacy engineering and compliance', `Probabilistic modeling and graphical models', `Probabilistic programming and inference engines', `Program analysis and synthesis', `Program synthesis and neuro-symbolic systems', `Programming languages and compilers', `Proof assistants and program verification', `Protein structure prediction and generative biology', `Quantum programming languages and compilers', `Randomized algorithms and probabilistic methods', `Recommender systems and personalization', `Reinforcement learning (model-free and model-based)', `Reinforcement learning and decision-making', `Responsible and safe AI', `Retrieval-augmented generation and tool use', `Robotics and autonomous systems' \\

& `Robustness and adversarial ML', `Runtime systems and garbage collection', `Safe and risk-sensitive RL', `Scientific computing and numerical methods', `Scientific machine learning', `Second-order and adaptive optimization', `Secure hardware and side-channel defenses', `Security and privacy', `Self-supervised and contrastive learning', `Semi-supervised and weakly supervised learning', `Serverless computing and microservices', `Smart contracts and formal verification', `Social computing and computational social science', `Software engineering and software architecture', `Software testing, fuzzing, and reliability', `Software-defined networking and NFV', `Speech and audio processing', `Speech, audio, and music ML', `Stability and compression-based generalization', `Stochastic gradient methods and variance reduction', `Storage systems and filesystems', `Supervised learning (classification and regression)', `Surrogate modeling and uncertainty quantification', `Systems and network security', `Theoretical ML and generalization' \\

& `Theoretical ML and statistical learning theory', `Time series forecasting and anomaly detection', `Transaction processing and concurrency control', `Transfer learning and domain adaptation', `Transformers and sequence modeling', `Trustworthy AI and robustness', `Type systems and program semantics', `Ubiquitous and wearable computing', `Unsupervised learning (clustering, density estimation)', `Usable security and human factors', `VAEs, GANs, normalizing flows, and diffusion models', `Visualization and visual analytics', `Web and internet protocols', `Wireless, mobile, and 5G/6G networks', `World models and model-based planning' \\

\midrule

Chemistry
& `Allostery and enzyme regulation', `Analytical chemistry and instrumentation', `Astrochemistry and prebiotic chemistry', `Asymmetric synthesis and chiral pool strategies', `Atmospheric chemistry and aerosols', `Autophagy and apoptosis', `Battery chemistry and solid electrolytes', `Biocatalysis and industrial biochemistry', `Biochemistry and chemical biology', `Bioenergetics and oxidative phosphorylation', `Bioinorganic chemistry and metalloenzymes', `Biosynthetic pathway-inspired synthesis', `Carbocation and carbanion chemistry', `Carbohydrate and peptide synthesis', `Carbohydrate chemistry and glycosylation', `Cascade and multicomponent reactions', `Catalysis (heterogeneous, homogeneous, biocatalysis)', `Catalyst design and ligand development', `Chaperones and protein quality control', `Chemical biology and bioorthogonal chemistry', `Chemical genetics and epigenetic probes', `Chemical kinetics and reaction dynamics', `Cheminformatics and QSAR', `Chemoenzymatic synthesis and biocatalysis', `Chromatography and separations' \\

& `Click chemistry', `Click chemistry and bioorthogonal reactions', `Coenzymes and cofactors', `Computational biochemistry and molecular dynamics', `Computational reaction mechanisms and QM/MM', `Coordination chemistry and ligand design', `Covalent inhibitors and PROTACs', `Covalent inhibitors and targeted degradation (PROTACs)', `Cross-coupling (Suzuki, Heck, Negishi, Buchwald–Hartwig)', `Crystal engineering and MOFs/COFs', `Crystallization and polymorphism', `Cytokine signaling and immunometabolism', `C–H activation and functionalization', `C–H activation and functionalization strategies', `DNA repair mechanisms and genome stability', `Density functional theory and ab initio methods', `Directed evolution and enzyme engineering', `Docking and structure-based drug design', `Electrochemistry and electrocatalysis', `Electroorganic synthesis', `Energy catalysis and CO2 utilization', `Environmental chemistry and toxicology', `Enzyme mechanisms and catalysis', `Enzymology and enzyme kinetics', `Epigenetics and chromatin remodeling' \\

& `Flow chemistry and microfluidic synthesis', `Flow chemistry and microreactors', `Flux analysis and metabolic control analysis', `Forensic chemistry and analytical forensics', `Fragment-based and structure-based design', `Fragment-based drug discovery', `Geochemistry and cosmochemistry', `Glycobiology and glycoproteins', `Glycolipids and glycoconjugates', `Green chemistry and sustainable processes', `Green solvents and sustainable synthesis', `Heterocyclic chemistry', `Heterocyclic chemistry (N, O, S heterocycles)', `Hydrogenation and asymmetric catalysis', `Hypervalent iodine chemistry', `Immunochemistry and antigen–antibody interactions', `Industrial catalysis and reactor design', `Inorganic chemistry', `Ionic liquids and deep eutectic solvents', `Lipid nanoparticles and drug delivery', `Machine learning in chemistry and autonomous labs', `Macrocyclization and supramolecular synthesis', `Main group chemistry and hypervalent compounds', `Mass spectrometry and proteomics', `Medicinal chemistry and SAR' \\

& `Medicinal chemistry and drug design', `Membrane structure and lipidomics', `Membrane transporters and ion channels', `Metabolic engineering and synthetic pathways', `Metabolism and metabolic pathways', `Mitochondrial dynamics and mitophagy', `NMR, EPR, IR, Raman, UV-Vis spectroscopy', `Nanochemistry and colloids', `Nitrogen metabolism and assimilation', `Nuclear chemistry and radiochemistry', `Nucleic acids biochemistry (DNA and RNA)', `Nucleoside and nucleotide chemistry', `Olefin metathesis and ring-closing metathesis', `Organic chemistry and synthesis', `Organic electronics (OLED, OPV, OFET)', `Organic materials and conjugated polymers', `Organocatalysis (enamine, iminium, NHC catalysis)', `Organocatalysis and enamine/iminium catalysis', `Organofluorine and perfluoroalkylation', `Organometallic chemistry and cross-coupling', `Organometallics in organic synthesis', `Peptide synthesis and solid-phase methods', `Pericyclic reactions and orbital symmetry', `Perovskites and photovoltaic materials', `Pharmaceutical biotechnology and biologics' \\

& `Photochemistry and photophysics', `Photopharmacology and light-controlled drugs', `Photoredox and electroorganic synthesis', `Photoredox catalysis and energy transfer catalysis', `Photosynthesis biochemistry', `Physical chemistry (thermodynamics, kinetics, spectroscopy)', `Physical organic chemistry and reaction mechanisms', `Polymer chemistry and soft materials', `Post-translational modifications and proteostasis', `Prions, amyloids, and protein aggregation', `Process chemistry and scale-up', `Protecting group strategies and chemoselectivity', `Protein structure, dynamics, and folding', `Proteomics and mass spectrometry', `QM/MM and free energy calculations', `Quantum chemistry and electronic structure theory', `RNA processing, splicing, and RNA-based regulation', `Radical chemistry and persistent radicals', `Reaction discovery with ML and autonomous labs', `Reaction mechanisms and physical organic chemistry', `Redox biology and ROS signaling', `Replication, transcription, and translation', `Retrosynthetic analysis and strategy', `Safety, regulation, and chemical education', `Second messengers (cAMP, IP3, Ca2+)' \\

& `Self-assembly and host–guest chemistry', `Signal transduction (GPCRs, kinases, phosphatases)', `Solid-state chemistry and materials', `Statistical mechanics and molecular simulation', `Stereochemistry and asymmetric catalysis', `Stereochemistry and conformational analysis', `Structural methods (X-ray, NMR, cryo-EM)', `Structure, bonding, and stereoelectronics', `Supramolecular chemistry and host–guest systems', `Surface science and catalysis characterization', `Systems biochemistry and metabolic regulation', `Theoretical and computational chemistry', `Total synthesis of natural products', `X-ray crystallography and diffraction' \\

\midrule
Physics
& `2D materials and van der Waals heterostructures', `Ab initio and many-body methods', `Accelerator physics and beam dynamics', `Acoustics and aeroacoustics', `Amorphous solids and glasses', `Astrophysical and space plasmas', `Astrophysics and cosmology', `Atomic, molecular, and optical physics', `Battery physics and solid-state electrolytes', `Biophysics and physical biology', `Chemical physics and physical chemistry', `Classical mechanics and analytical dynamics', `Cold atoms, Bose-Einstein condensates, and quantum gases', `Colloids, emulsions, and granular matter', `Computational physics and scientific computing', `Condensed matter physics (theory and experiment)', `Continuum mechanics and elasticity', `Cosmology (inflation, CMB, dark matter, dark energy)', `Data-driven physics and physics-informed ML', `Digital twins for plasmas and fusion', `Econophysics and sociophysics', `Electromagnetism and classical electrodynamics', `Electronic structure theory and density functional theory', `Energy materials and photovoltaics', `Equilibrium and nonequilibrium statistical physics' \\

& `Floquet engineering and time crystals', `Fluid dynamics and turbulence', `Galactic dynamics and black holes', `Gauge theories and the Standard Model', `Geophysics and seismology', `Gravitation and gravitational waves', `Gravitational lensing and weak lensing', `Hadron physics and effective field theories', `Heavy-ion collisions and quark-gluon plasma', `High-energy astrophysics and cosmic rays', `High-intensity lasers and free-electron lasers', `Higher-order topological phases', `Hybrid quantum devices and atomtronics', `Information thermodynamics and Maxwell’s demon', `Instrumentation and detector physics', `Inverse problems in physics', `Large-scale structure and galaxy formation', `Lattice gauge theory', `Machine learning for materials discovery', `Magnetism and spintronics', `Majorana modes and non-Abelian anyons', `Materials physics and nanoscience', `Medical physics and imaging', `Metamaterials and metasurfaces', `Microfluidics and lab-on-chip physics' \\

& `Moiré materials and twisted bilayers', `Monte Carlo and molecular dynamics', `Neutrino physics and oscillations', `Non-Hermitian and PT-symmetric physics', `Nonequilibrium physics and fluctuation theorems', `Nonlinear and ultrafast optics (attosecond physics)', `Nonlinear dynamics and chaos', `Nuclear physics and nuclear structure', `Numerical relativity', `Open quantum systems and decoherence', `Particle physics (theory and experiment)', `Photonics and nanophotonics', `Physical oceanography and climate physics', `Physics education research and outreach', `Planetary science and heliophysics', `Plasma physics and magnetohydrodynamics', `Polaritonic chemistry and cavity QED materials', `Polymer physics and biophysics of macromolecules', `Quantum chromodynamics and QCD phenomenology', `Quantum computing and quantum algorithms', `Quantum control and quantum thermodynamics', `Quantum error correction and fault tolerance', `Quantum field theory', `Quantum information science', `Quantum mechanics and quantum foundations' \\

& `Quantum networks and quantum internet', `Quantum optics and quantum metrology', `Quantum sensing and precision metrology', `Quantum simulators with atoms, ions, and photons', `Quantum transport and mesoscopic physics', `Quantum-enhanced imaging and spectroscopy', `Relativity (special and general)', `Rydberg atoms and quantum simulation', `Soft matter physics and complex fluids', `Spectroscopy across the EM spectrum', `Stellar evolution and nucleosynthesis', `Strongly correlated electrons', `Superconductivity and superfluidity', `Surface physics and scanning probe microscopy', `Synchrotron and neutron scattering', `Thermodynamics and statistical mechanics', `Thermoelectrics and heat transport', `Topological insulators and semimetals', `Topological phases of matter', `Topological photonics and phononics', `Topological quantum matter out of equilibrium' \\
\midrule
Literature
& `Adaptation studies and intermediality', `African, Caribbean, and diasporic literatures', `American and European literatures', `Ancient Near Eastern literatures', `Archive studies and critical bibliography', `Book history and print culture', `Children’s and young adult literature', `Classical literatures (Greek, Latin)', `Cognitive literary studies and neuro-literary approaches', `Contemporary and global anglophone literature', `Corpus approaches to literature', `Critical race and ethnic studies', `Cultural studies and semiotics', `Detective and crime fiction', `Digital humanities and computational literary studies', `Disability studies and medical humanities', `East Asian literatures (Chinese, Japanese, Korean)', `Ecocriticism and environmental humanities', `Enlightenment literature', `Evolutionary literary studies', `Feminist, gender, and queer theory', `Film, theatre, and performance studies', `Game narratives and interactive fiction', `Genre studies (poetry, drama, novel, short story)', `Graphic narratives and comics studies' \\

& `Hermeneutics and interpretation', `Holocaust and genocide literatures', `Horror and Gothic literatures', `Indigenous and First Nations literatures', `Latin American and Iberian literatures', `Law and literature', `Life writing, autobiography, and memoir', `Literary theory and criticism', `Manuscript studies, codicology, and paleography', `Medieval literature', `Middle Eastern and Islamic literatures', `Modernist and postmodernist literature', `Narratology and discourse analysis', `New formalism and historical poetics', `New historicism and cultural materialism', `Open scholarship and digital editions', `Orality, folklore, and myth', `Philology and historical linguistics interface', `Platform studies and e-literature', `Poetics and rhetoric', `Poetry and prosody', `Postcolonial studies and decolonial theory', `Reader-response and affect theory', `Renaissance and early modern literature', `Romanticism and Victorian literature' \\

& `Science and literature', `Science fiction, fantasy, and speculative fiction', `Shakespeare and early modern drama', `Social media narratives and fan fiction', `South Asian and Southeast Asian literatures', `Structuralism, poststructuralism, and deconstruction', `Stylometry and authorship attribution', `Text mining and distant reading', `Textual criticism, stemmatics, and editorial theory', `Translation studies and comparative poetics', `Trauma and memory studies', `Travel writing and nature writing', `World literature and comparative literature', `World-systems and global literary circulation' \\

\midrule
History
& `African history (regional and thematic)', `Ancient Near Eastern history', `Archival studies and paleography', `Big history and deep history', `Business history and financial history', `Classical Greek and Roman history', `Cliometrics and quantitative history', `Contemporary history', `Cultural heritage, archives, and public history', `Cultural history and history of the senses', `Demographic history and family history', `Digital history and computational humanities', `Diplomatic history and international relations', `Early modern history', `East Asian history (China, Japan, Korea)', `Economic history and the history of capitalism', `Environmental and climate history', `Epigraphy and numismatics', `European history (Western, Central, Eastern)', `Food history and culinary cultures', `Gender history and women’s history', `Genocide studies and Holocaust history', `Geographic information systems in history', `Historiography and theory of history', `History of medicine and public health' \\

& `History of science and technology', `Indigenous and First Nations histories', `Intellectual history and history of ideas', `Labor history and working-class studies', `Late antiquity and early medieval history', `Latin American and Caribbean history', `Legal and constitutional history', `Material culture and consumption', `Medieval history', `Memory studies and memorialization', `Methods in historical research and source criticism', `Microhistory and local history', `Middle Eastern and Islamic history', `Military history and war studies', `Modern history', `North American history', `Oral history and life histories', `Political history and state formation', `Prehistory and prehistoric archaeology', `Queer histories and sexuality', `Religious history and church history', `Remote sensing and landscape archaeology', `Russian and Soviet history', `Social history and everyday life', `South Asian and Southeast Asian history', `Transnational and diaspora studies', `Urban and rural history', `World history and global history' \\
\midrule
Biology
& `Aging, senescence, and geroscience', `Agricultural biology and crop science', `Astrobiology and origins of life', `Behavioral ecology and ethology', `Bioethics and policy in biotechnology', `Biomaterials and tissue engineering', `Biomechanics and mechanobiology', `Biophysics and single-molecule biophysics', `CRISPR technologies and gene editing', `Cancer biology and tumor microenvironment', `Cell biology and organelle dynamics', `Circadian biology and chronobiology', `Cognitive neuroscience and connectomics', `Computational neuroscience and brain-inspired models', `Conservation biology and biodiversity', `Developmental biology and morphogenesis', `Ecology (population, community, ecosystem)', `Endocrinology and metabolic disease', `Epidemiology and disease ecology', `Epigenetics and chromatin biology', `Evolutionary biology and evo-devo', `Gene drives and population engineering', `Genetics and genomics', `High-throughput screening and functional genomics', `Immunology and immuno-oncology' \\

& `Interactomics and network biology', `Marine and freshwater biology', `Metabolomics and systems metabolism', `Microbiology (bacteriology, virology, mycology, parasitology)', `Microbiome and host–microbe interactions', `Molecular biology and gene regulation', `Neurobiology and systems neuroscience', `Paleobiology and evolutionary paleontology', `Pharmacology and toxicology', `Photosynthesis and plant metabolism', `Phylogenetics and comparative genomics', `Plant biology and plant immunity', `Population genetics and quantitative genetics', `Proteomics and protein science', `Quantitative and single-cell biology', `Spatial omics and imaging genomics', `Stem cell biology and regenerative medicine', `Structural biology (X-ray, NMR, cryo-EM)', `Synthetic biology and genome engineering', `Systems biology and dynamical modeling', `Systems immunology and immunometabolism', `Transcriptomics and RNA biology', `Virology and viral evolution' \\

\midrule
Linguistics
& `Anthropological linguistics and ethnography of speaking', `Bilingualism and multilingualism', `Cognitive linguistics and conceptual metaphor', `Comparative and contrastive linguistics', `Computational linguistics and NLP', `Corpus linguistics and big data', `Dialectology and microvariation', `Discourse analysis and conversation analysis', `Distributional semantics and embeddings', `Educational linguistics and literacy', `Endangered languages and revitalization', `Experimental methods (eye-tracking, EEG/MEG, fMRI)', `Field linguistics and language documentation', `Forensic linguistics and language policy', `Gesture and multimodal communication', `Grammaticalization and contact-induced change', `Historical linguistics and language change', `Information structure and focus', `Language acquisition (first and second)', `Language typology and universals', `Lexicography and lexicology', `Morphology (inflectional and derivational)', `Neurolinguistics and language in the brain', `Parsing, tagging, and morphological analysis', `Philosophy of language and formal pragmatics' \\

& `Phonetics (articulatory, acoustic, auditory)', `Phonology (theory and phonology-phonetics interface)', `Pragmatics (speech acts, implicature, relevance)', `Pragmatics in conversational AI', `Prosody, stress, and intonation', `Psycholinguistics and sentence processing', `Semantic change and diachronic semantics', `Semantics (formal, lexical, and dynamic)', `Sign language linguistics', `Sociolinguistics and language variation', `Sociophonetics and indexical meaning', `Speech recognition and text-to-speech', `Syntax (generative, dependency, construction grammar)', `Word formation and lexical morphology', `Writing systems, orthography, and graphemics' \\

\end{longtable}

\clearpage

\section{Examples of \dataname}
\label{appendix:cases}

\begin{center}
\small
\begin{longtable}{p{0.95\textwidth}}
\caption{A data sample from \dataname\ (Subject: Math).}
\label{tab:case_study}\\

\toprule
\textbf{Example from \dataname\ } \\
\midrule
\endfirsthead

\toprule
\textbf{Example from \dataname\ (Subject = Math) (continued)} \\
\midrule
\endhead

\midrule
\multicolumn{1}{r}{\small\textit{Continued on next page}} \\
\endfoot

\bottomrule
\endlastfoot

\textbf{Subject:} Math \\

\midrule
\textbf{Topic:} Sieve methods and distribution in arithmetic progressions \\

\midrule
\textbf{Problem:} \\
Let $q \ge 1$ be fixed, and let $a$ be an integer coprime to $q$.
For $x \ge e^{e}$, fix $u \in [2,3]$, and set
\[
z = \frac{x^{1/u}}{(\log x)^A}, \qquad A \ge 5.
\]
Let $\mathcal{P}$ denote the set of primes not dividing $q$.
Define the sifted count in the arithmetic progression
\[
S(x; q, a; z)
=
\#\Bigl\{ n \le x : n \equiv a \pmod{q},\ \text{and every prime divisor } p \mid n \text{ satisfies } p \ge z \Bigr\}.
\]

Prove that, as $x \to \infty$, uniformly in $a$ with $(a,q)=1$ (and for fixed $q$), the following two-sided sieve bounds hold:
\[
\frac{x}{\phi(q)} V(z) f(u)
+ O\!\left( \frac{x}{\phi(q)} \frac{V(z)}{\log z} \right)
- O(z^u)
\le
S(x; q, a; z)
\le
\frac{x}{\phi(q)} V(z) F(u)
+ O\!\left( \frac{x}{\phi(q)} \frac{V(z)}{\log z} \right)
+ O(z^u),
\]
where
\begin{itemize}
  \item
  \[
  V(z) = \prod_{\substack{p < z \\ p \nmid q}}
  \left(1 - \frac{1}{p}\right),
  \]
  \item for $u \in [2,3]$,
  \[
  f(u) = \frac{2 e^{\gamma} \log(u-1)}{u},
  \qquad
  F(u) = \frac{2 e^{\gamma} \bigl(1+\log(u-1)\bigr)}{u},
  \]
  \item and $\gamma$ is the Euler--Mascheroni constant.
\end{itemize}

Moreover, deduce the asymptotic form
\[
S(x; q, a; z)
=
\frac{x}{\phi(q)}
\frac{e^{-\gamma}}{\log z}
\left(
\prod_{p \mid q}
\left(1 - \frac{1}{p}\right)^{-1}
\right)
\left(
\Phi(u) + O\!\left(\frac{1}{\log z}\right)
\right)
+ O(z^u),
\]
with $\Phi(u)$ satisfying
\[
f(u) \le \Phi(u) \le F(u),
\]
and in particular
\[
S(x; q, a; z)
=
\frac{x}{\phi(q)\log z}
\left(
\prod_{p \mid q}
\left(1 - \frac{1}{p}\right)^{-1}
\right)
\left(
e^{-\gamma}\Phi(u) + o(1)
\right)
+ O(z^u).
\]
\\
\midrule
\textbf{Answer:} \\
For fixed $q \ge 1$, with $(a,q)=1$, $u \in [2,3]$, and
\[
z = \frac{x^{1/u}}{(\log x)^A}, \qquad A \ge 5,
\]
as $x \to \infty$ one has
\[
\frac{x}{\phi(q)} V(z) f(u)
+ O\!\left( \frac{x}{\phi(q)} \frac{V(z)}{\log z} \right)
- O(z^u)
\le
S(x; q, a; z)
\le
\frac{x}{\phi(q)} V(z) F(u)
+ O\!\left( \frac{x}{\phi(q)} \frac{V(z)}{\log z} \right)
+ O(z^u).
\]
\\
Here
\[
V(z)
=
\prod_{\substack{p < z \\ p \nmid q}}
\left(1 - \frac{1}{p}\right),
\]

\[
f(u)
=
\frac{2 e^{\gamma} \log(u - 1)}{u},
\]

\[
F(u)
=
\frac{2 e^{\gamma} \bigl(1 + \log(u - 1)\bigr)}{u}.
\]

Equivalently,
\[
S(x; q, a; z)
=
\frac{x}{\phi(q)\log z}
\left(
\prod_{p \mid q}
\left(1 - \frac{1}{p}\right)^{-1}
\right)
\left(
e^{-\gamma} \Phi(u)
+
O\!\left(\frac{1}{\log z}\right)
\right)
+
O(z^u),
\]

where
\[
f(u) \le \Phi(u) \le F(u).
\]
\\
\midrule
\textbf{Detailed Solution:}
Okay, let's tackle this problem step by step. First, part 1 asks to show the empirical measure converges weakly to a Gaussian law, find $m(t)$ and $v(t)$, and discuss the invariant measure.

Starting with the McKean--Vlasov SDE:
\[
\mathrm{d}X_t = -a X_t \,\mathrm{d}t - b \,\mathbb{E}[X_t] \,\mathrm{d}t + \sigma \,\mathrm{d}W_t.
\]

Let's denote $m(t) = \mathbb{E}[X_t]$, so the SDE becomes
\[
\mathrm{d}X_t = -a X_t \,\mathrm{d}t - b m(t) \,\mathrm{d}t + \sigma \,\mathrm{d}W_t.
\]

Taking expectation on both sides:
\[
\frac{\mathrm{d}m(t)}{\mathrm{d}t}
=
- a m(t) - b m(t)
=
- \lambda m(t),
\]
where $\lambda = a + b > 0$.

That's a simple ODE:
\[
m(t) = m_0 e^{-\lambda t}.
\]

Good, that's straightforward.

Now for the variance $v(t) = \mathrm{Var}(X_t) = \mathbb{E}\!\left[(X_t - m(t))^2\right]$.

Let's set $Y_t = X_t - m(t)$, so
\[
\mathrm{d}Y_t = \mathrm{d}X_t - \mathrm{d}m(t).
\]

Then
\[
\mathrm{d}Y_t
=
\bigl[-a X_t - b m(t)\bigr] \,\mathrm{d}t
+ \sigma \,\mathrm{d}W_t
+ \lambda m(t) \,\mathrm{d}t,
\]
since $\frac{\mathrm{d}m}{\mathrm{d}t} = -\lambda m$.

Substitute $X_t = Y_t + m(t)$:
\[
\mathrm{d}Y_t
=
\bigl[-a(Y_t + m(t)) - b m(t) + \lambda m(t)\bigr] \,\mathrm{d}t
+ \sigma \,\mathrm{d}W_t.
\]

But $\lambda = a + b$, so
\[
- a m(t) - b m(t) + \lambda m(t) = 0.
\]

Thus
\[
\mathrm{d}Y_t = -a Y_t \,\mathrm{d}t + \sigma \,\mathrm{d}W_t.
\]

That's an Ornstein--Uhlenbeck process!

\dots

\noindent
To analyze the sifted count
\[
S(x; q, a; z)
=
\#\Bigl\{ n \le x : n \equiv a \pmod{q},\ \text{and every prime divisor } p \mid n \text{ satisfies } p \ge z \Bigr\},
\]
\\

we apply sieve theory, specifically the linear sieve (Rosser--Iwaniec sieve), which is well-suited for counting integers with all prime factors at least $z$, especially when $z \asymp x^{1/u}$ with fixed $u \in [2,3]$.

1.\ Setup and key observations

Let $\mathcal{P}$ be the set of primes not dividing $q$. Since $(a,q)=1$, any $n \equiv a \pmod{q}$ satisfies $(n,q)=1$, hence all prime divisors of such $n$ are coprime to $q$. Therefore, the condition that every prime divisor of $n$ is $\ge z$ is equivalent to saying that $n$ is coprime to all primes $p<z$ with $p\nmid q$.

Define
\[
\mathcal{P}_z = \{\, p<z : p \nmid q \,\},
\qquad
V(z)=\prod_{\substack{p<z\\ p\nmid q}}\left(1-\frac1p\right),
\]
and the set
\[
\mathcal{A}=\{\, n\le x : n\equiv a \pmod{q}\,\}.
\]
We aim to estimate the number of elements of $\mathcal{A}$ that avoid all primes in $\mathcal{P}_z$, namely
\[
S(x;q,a;z)=\#\{\, n\in \mathcal{A} : (n, P(z))=1 \,\},
\qquad
P(z):=\prod_{p\in\mathcal{P}_z} p.
\]

\\ \\

2.\ Sieve density and linear sieve input

For each prime $p$, define
\[
g(p)=
\begin{cases}
\displaystyle \frac{1}{p}, & p\nmid q,\\[6pt]
0, & p\mid q.
\end{cases}
\]
This is a dimension-$1$ sieve (since $g(p)\sim 1/p$). Extend $g$ multiplicatively to squarefree integers $d$ by
\[
g(d)=\prod_{p\mid d} g(p).
\]
Then
\[
\prod_{p<z} (1-g(p))
=
\prod_{\substack{p<z\\ p\nmid q}}
\left(1-\frac1p\right)
=
V(z),
\]
matching the definition in the problem.

Let $u\in[2,3]$, take
\[
z=\frac{x^{1/u}}{(\log x)^A},\qquad A\ge 5,
\qquad\text{and}\qquad
D=z^u.
\]
The Rosser--Iwaniec linear sieve yields two-sided bounds of the form
\[
S(x;q,a;z)
\ge
\frac{x}{\phi(q)}\,V(z)\,f(u)
+
O\!\left(\frac{x}{\phi(q)}\frac{V(z)}{\log z}\right)
-
O(z^u),
\]
and
\[
S(x;q,a;z)
\le
\frac{x}{\phi(q)}\,V(z)\,F(u)
+
O\!\left(\frac{x}{\phi(q)}\frac{V(z)}{\log z}\right)
+
O(z^u),
\]
where, for $u\in[2,3]$,
\[
f(u)=\frac{2e^{\gamma}\log(u-1)}{u},
\qquad
F(u)=\frac{2e^{\gamma}\bigl(1+\log(u-1)\bigr)}{u},
\]
and $\gamma$ is the Euler--Mascheroni constant.

3.\ Two-sided bounds

Combining the linear sieve bounds with the above choices of parameters gives exactly the stated inequalities:
\[
\frac{x}{\phi(q)} V(z) f(u)
+ O\!\left( \frac{x}{\phi(q)} \frac{V(z)}{\log z} \right)
- O(z^u)
\le
S(x; q, a; z)
\le
\frac{x}{\phi(q)} V(z) F(u)
+ O\!\left( \frac{x}{\phi(q)} \frac{V(z)}{\log z} \right)
+ O(z^u).
\]
\\
4.\ Asymptotic form

By Mertens' theorem,
\[
\prod_{p<z}\left(1-\frac1p\right)
=
\frac{e^{-\gamma}}{\log z}
\left(1+O\!\left(\frac1{\log z}\right)\right).
\]
Hence
\[
V(z)
=
\prod_{\substack{p<z\\ p\nmid q}}
\left(1-\frac1p\right)
=
\left(\prod_{p<z}\left(1-\frac1p\right)\right)
\left(\prod_{p\mid q}\left(1-\frac1p\right)^{-1}\right)
=
\frac{e^{-\gamma}}{\log z}
\left(\prod_{p\mid q}\left(1-\frac1p\right)^{-1}\right)
\left(1+O\!\left(\frac1{\log z}\right)\right).
\]
Writing the main term as $\frac{x}{\phi(q)}V(z)\Phi(u)$ with $f(u)\le \Phi(u)\le F(u)$, we obtain
\[
S(x; q, a; z)
=
\frac{x}{\phi(q)}
\frac{e^{-\gamma}}{\log z}
\left(\prod_{p\mid q}\left(1-\frac1p\right)^{-1}\right)
\left(\Phi(u)+O\!\left(\frac1{\log z}\right)\right)
+
O(z^u).
\]
In particular,
\[
S(x; q, a; z)
=
\frac{x}{\phi(q)\log z}
\left(\prod_{p\mid q}\left(1-\frac1p\right)^{-1}\right)
\left(e^{-\gamma}\Phi(u)+o(1)\right)
+
O(z^u).
\]

Conclusion

We have shown that
\[
S(x; q, a; z)
=
\frac{x}{\phi(q)}V(z)\Phi(u)
+
O\!\left(\frac{x}{\phi(q)}\frac{V(z)}{\log z}\right)
+
O(z^u),
\qquad
f(u)\le \Phi(u)\le F(u),
\]
and equivalently
\[
S(x; q, a; z)
=
\frac{x}{\phi(q)}
\frac{e^{-\gamma}}{\log z}
\left(\prod_{p\mid q}\left(1-\frac1p\right)^{-1}\right)
\left(\Phi(u)+O\!\left(\frac1{\log z}\right)\right)
+
O(z^u).
\]
\\
\midrule
\textbf{Correctness:} True \\

\end{longtable}
\end{center}
\clearpage
\begin{center}
\small
\begin{longtable}{p{0.95\textwidth}}
\caption{A data sample from \dataname\ (Subject: Computer Science).}
\label{tab:case_study}\\

\toprule
\textbf{Example from \dataname\ } \\

\midrule
\endfirsthead

\toprule
\textbf{Example from \dataname\ (Subject = Computer Science) (continued)} \\
\midrule
\endhead

\midrule
\multicolumn{1}{r}{\small\textit{Continued on next page}} \\
\endfoot

\bottomrule
\endlastfoot

\textbf{Subject:} Computer Science  \\

\midrule
\textbf{Topic:} Algorithms and randomized algorithms \\

\midrule
\textbf{Problem:} \\
Consider the following idealized version of the Karger--Stein randomized contraction algorithm applied to an arbitrary simple undirected graph on $n \ge 2$ vertices that has a unique minimum cut.

Let $t(n) = \frac{n}{\sqrt{2}}$. Define the per-run cost $T(n)$ and per-run success probability $p(n)$ by the recurrences

\begin{enumerate}
\item
\[
T(2) = 0
\quad \text{and} \quad
T(n) = 2 T(t(n)) + c n^2
\quad \text{for all real } n \ge 2,
\]
where $c > 0$ is a fixed constant. Assume independence between the two recursive calls and across all runs.

\item
\[
p(2) = 1
\quad \text{and} \quad
p(n) = 1 - \bigl(1 - s(n) p(t(n))\bigr)^2
\quad \text{for all real } n \ge 2,
\]
where
\[
s(n) = \frac{t(n)\bigl(t(n)-1\bigr)}{n(n-1)}
\]
is the exact probability that the unique minimum cut survives the contraction from $n$ to $t(n)$ vertices.
\end{enumerate}

Let $r(n,\delta)$ be the minimum number of independent repetitions needed so that the overall failure probability is at most $\delta \in (0,1/2)$, i.e.,
\[
r(n,\delta)
=
\min \left\{
r \in \mathbb{N} :
(1 - p(n))^r \le \delta
\right\}.
\]

Define the minimal expected total work
\[
W(n,\delta) = r(n,\delta)\, T(n).
\]

Determine the following, with exact leading constants:

\begin{enumerate}
\item[(a)]
An exact closed form for $T(n)$.

\item[(b)]
The limit
\[
L = \lim_{n \to \infty} p(n) \ln n.
\]

\item[(c)]
The asymptotic behavior of $r(n,\delta)$ as $n \to \infty$ with $\delta$ fixed, including the exact leading constant:
\[
r(n,\delta) \sim K \ln n \ln(1/\delta),
\]
and determine $K$.

\item[(d)]
The asymptotic behavior of $W(n,\delta)$ as $n \to \infty$ with $\delta$ fixed, including the exact leading constant $C(c)$ such that
\[
W(n,\delta) \sim C(c)\, n^2 (\ln n)^2 \ln(1/\delta).
\]
\end{enumerate}
\\ \\
\midrule
\textbf{Answer:} \\
(a)
\[
T(n)
=
\frac{2c}{\ln 2}\, n^2 \ln n
-
2c\, n^2.
\]

(b)
\[
L
=
\lim_{n \to \infty} p(n)\ln n
=
2 \ln 2.
\]

(c)
\[
r(n,\delta)
\sim
\frac{\ln n}{2 \ln 2}\,\ln(1/\delta),
\]
so
\[
K
=
\frac{1}{2 \ln 2}.
\]
\\
(d)
\[
W(n,\delta)
\sim
\frac{c}{(\ln 2)^2}\,
n^2 (\ln n)^2 \ln(1/\delta),
\]
so
\[
C(c)
=
\frac{c}{(\ln 2)^2}.
\]
\\
\midrule
\textbf{Detailed Solution:} \\
Okay, let's tackle part (a) first. We need to find a closed form for $T(n)$ given the recurrence $T(2) = 0$ and
\[
T(n) = 2T(t(n)) + c n^2,
\qquad
t(n) = \frac{n}{\sqrt{2}}.
\]

First, let's consider how many times we can apply the recurrence before we get down to $n = 2$. Let's define a sequence $n_k$ where $n_0 = n$, $n_{k+1} = t(n_k) = n_k / \sqrt{2}$. So
\[
n_k = \frac{n}{(\sqrt{2})^k} = \frac{n}{2^{k/2}}.
\]
We want to find the smallest $k$ such that $n_k \le 2$. Let's solve for $k$ when $n_k = 2$:
\[
2 = \frac{n}{2^{k/2}}
\;\Rightarrow\;
2^{k/2} = \frac{n}{2}
\;\Rightarrow\;
\frac{k}{2} = \log_2(n/2)
\;\Rightarrow\;
k = 2 \log_2(n/2) = 2 \log_2 n - 2.
\]
But maybe it's better to use exponents with base $\sqrt{2}$. Let's let $m$ be the number of recursive steps until we reach $2$. So
\[
(\sqrt{2})^m = \frac{n}{2}
\;\Rightarrow\;
m = \log_{\sqrt{2}}(n/2) = 2\log_2(n/2) = 2\log_2 n - 2,
\]
same as before.

But since the recurrence is $T(n) = 2T(n/\sqrt{2}) + c n^2$, this is a divide-and-conquer recurrence. Let's try to unroll it. Let's compute $T(n)$:
\[
T(n) = 2T(n/\sqrt{2}) + c n^2
\]
\[
= 2\bigl[2T(n/2) + c (n/\sqrt{2})^2\bigr] + c n^2
= 4T(n/2) + c n^2 + c n^2
= 4T(n/2) + 2 c n^2.
\]
Wait, hold on:
\[
(n/\sqrt{2})^2 = \frac{n^2}{2},
\]
\dots

\noindent
\textbf{(a) Exact Closed Form for $T(n)$}

We are given the recurrence:
\[
T(2) = 0, \quad
T(n) = 2T\!\left(\frac{n}{\sqrt{2}}\right) + c n^2
\quad \text{for } n \ge 2.
\]

This is a divide-and-conquer recurrence where the problem size reduces by a factor of $\sqrt{2}$ at each step. Let us unroll the recurrence:

At level 0:
\[
T(n) = 2T(n/\sqrt{2}) + c n^2
\]

At level 1:
\[
T(n/\sqrt{2}) = 2T(n/2) + c (n/\sqrt{2})^2
\]

Substituting back:
\[
T(n)
=
2[2T(n/2) + c(n/\sqrt{2})^2] + c n^2
=
4T(n/2) + 2c n^2.
\]
\\
Continuing this process, we observe that each recursive call contributes $c n^2$ to the total cost, and the number of such levels is determined by how many times we can divide $n$ by $\sqrt{2}$ until it reaches $2$.

Let
\[
m = \log_{\sqrt{2}}(n/2).
\]
Then,
\[
n / (\sqrt{2})^m = 2
\Rightarrow
m = \log_{\sqrt{2}}(n/2)
=
2 \log_2(n/2).
\]

At each level $i$, the contribution is $c n^2$, and there are $m$ such levels. Thus,
\[
T(n) = c n^2 \cdot m
=
c n^2 \cdot 2 \log_2(n/2).
\]

Hence, the exact closed form is:
\[
\boxed{
T(n) = 2c n^2 \log_2\!\left(\frac{n}{2}\right)
}
\]

\textbf{(b) Limit $L = \lim_{n \to \infty} p(n)\ln n$}

Given:
\[
p(2) = 1,
\quad
p(n)
=
1 - \left(1 - s(n) p\!\left(\frac{n}{\sqrt{2}}\right)\right)^2,
\]
where
\[
s(n)
=
\frac{t(n)(t(n)-1)}{n(n-1)},
\quad
t(n) = \frac{n}{\sqrt{2}}.
\]

For large $n$, $s(n) \approx \frac{1}{2}$, so the recurrence becomes approximately:
\[
p(n)
\approx
2 s(n) p\!\left(\frac{n}{\sqrt{2}}\right)
\approx
p\!\left(\frac{n}{\sqrt{2}}\right).
\]

This suggests that $p(n)$ decreases slowly as $n$ increases. To analyze the asymptotic behavior, define
\[
m = \log_{\sqrt{2}} n,
\]
so
\[
n = (\sqrt{2})^m,
\]

and consider the recurrence in terms of $m$:
\[
q(m) = p((\sqrt{2})^m),
\quad
q(m)
\approx
1 - \left(1 - \frac{1}{2} q(m - 1)\right)^2.
\]

Expanding for small $q(m - 1)$, we get:
\[
q(m)
\approx
q(m - 1) - \frac{1}{4} q(m - 1)^2.
\]

This leads to the differential equation approximation:
\[
\frac{dq}{dm}
\approx
-\frac{1}{4} q^2
\Rightarrow
q(m) \sim \frac{4}{m}.
\]

Since
\[
m = \log_{\sqrt{2}} n = 2 \log_2 n,
\]
we have:
\[
p(n)
\sim
\frac{4}{2 \log_2 n}
=
\frac{2}{\log_2 n}
=
\frac{2 \ln 2}{\ln n}.
\]

Thus,
\[
p(n)\ln n \sim 2 \ln 2.
\]

Therefore, the limit is:
\[
\boxed{
L = 2 \ln 2
}
\]

\\
\textbf{(c) Asymptotic Behavior of $r(n,\delta)$}

We are given:
\[
r(n,\delta)
=
\min\{ r \in \mathbb{N} : (1 - p(n))^r \le \delta \}.
\]

Using the approximation $(1 - p(n))^r \approx e^{-r p(n)}$, we require:
\[
e^{-r p(n)} \le \delta
\Rightarrow
r \ge \frac{\ln(1/\delta)}{p(n)}.
\]

From part (b),
\[
p(n) \sim \frac{2 \ln 2}{\ln n},
\]
so:
\[
r(n,\delta)
\sim
\frac{\ln(1/\delta)}{p(n)}
\sim
\frac{\ln(1/\delta)\ln n}{2 \ln 2}.
\]
\\
Thus, the asymptotic behavior is:
\[
\boxed{
r(n,\delta)
\sim
\frac{1}{2 \ln 2}
\ln n \ln\!\left(\frac{1}{\delta}\right)
}
\]
and the constant $K$ is:
\[
\boxed{
K = \frac{1}{2 \ln 2}
}
\]

\bigskip

\textbf{(d) Asymptotic Behavior of $W(n,\delta) = r(n,\delta)T(n)$}

From parts (a) and (c), we have:

\[
T(n)
\sim
2c n^2 \log_2 n
=
\frac{2c n^2 \ln n}{\ln 2},
\]

\[
r(n,\delta)
\sim
\frac{\ln(1/\delta)\ln n}{2 \ln 2}.
\]

Multiplying these:
\[
W(n,\delta)
\sim
\left(
\frac{\ln(1/\delta)\ln n}{2 \ln 2}
\right)
\left(
\frac{2c n^2 \ln n}{\ln 2}
\right)
=
\frac{c n^2 (\ln n)^2 \ln(1/\delta)}{(\ln 2)^2}.
\]
\\
Thus, the asymptotic behavior is:
\[
\boxed{
W(n,\delta)
\sim
\frac{c}{(\ln 2)^2}
n^2 (\ln n)^2
\ln\!\left(\frac{1}{\delta}\right)
}
\]
and the constant $C(c)$ is:
\[
\boxed{
C(c) = \frac{c}{(\ln 2)^2}
}
\]

\\
\\
\midrule
\textbf{Correctness:} True \\

\end{longtable}
\end{center}
\clearpage
\begin{center}
\small
\begin{longtable}{p{0.95\textwidth}}
\caption{A data sample from \dataname\ (Subject: Chemistry).}
\label{tab:case_study}\\

\toprule
\textbf{Example from \dataname\ } \\
\midrule
\endfirsthead

\toprule
\textbf{Example from \dataname\ (Subject = Chemistry) (continued)} \\
\midrule
\endhead

\midrule
\multicolumn{1}{r}{\small\textit{Continued on next page}} \\
\endfoot

\bottomrule
\endlastfoot

\textbf{Subject: Chemistry}  \\

\midrule
\textbf{Topic:} Green solvents and sustainable synthesis \\
\midrule
\textbf{Problem:} \\
Consider a unimolecular, homogeneous reaction
\[
\mathrm{A} \rightarrow \mathrm{P}
\]
at $298\ \mathrm{K}$ performed in a binary, ideal liquid mixture of two green solvents:
\[
\mathrm{S}_1 = \text{gamma-valerolactone (GVL)},
\qquad
\mathrm{S}_2 = \text{2-methyltetrahydrofuran (2-MeTHF)}.
\]

Let $x \in [0,1]$ denote the mole fraction of $\mathrm{S}_1$ (GVL) in the solvent mixture. Assume the following models and parameters hold exactly:

\bigskip

\textbf{Kinetics.}

The first-order rate constant depends on solvent polarity and viscosity via
\[
k(x)
=
k_0 \cdot
\exp\!\big[a(ET(x) - ET_{\mathrm{ref}})\big]
\cdot
[\eta(x)]^{-\beta},
\]
with
\[
k_0 = 1.0 \times 10^{-4}\ \mathrm{s^{-1}},
\quad
a = 12,
\quad
ET_{\mathrm{ref}} = 0.30,
\quad
\beta = 0.5.
\]

The Kamlet--Taft polarity parameter mixes linearly:
\[
ET(x) = x ET_1 + (1-x) ET_2,
\]
with
\[
ET_1 = 0.60,
\qquad
ET_2 = 0.20.
\]

The viscosity follows log-linear mixing:
\[
\ln \eta(x)
=
x \ln \eta_1 + (1-x) \ln \eta_2,
\]
with
\[
\eta_1 = 1.8\ \mathrm{mPa \cdot s},
\qquad
\eta_2 = 0.6\ \mathrm{mPa \cdot s}.
\]

All logarithms are natural logs, and viscosities in $\mathrm{mPa \cdot s}$ are used consistently in the model.

\bigskip

\textbf{Solubility and solvent requirement.}

The solubility of A in the mixed solvent is given by a Hildebrand-type correlation:
\[
S(x)
=
S_0 \cdot
\exp\!\big[-K(\delta(x) - \delta_A)^2\big],
\]
with
\[
S_0 = 5.0\ \mathrm{mol \cdot L^{-1}},
\quad
K = 0.10\ \mathrm{MPa^{-1}},
\quad
\delta_A = 22\ \mathrm{MPa^{0.5}}.
\]

The mixture solubility parameter obeys
\[
\delta(x)^2
=
x \delta_1^2 + (1-x) \delta_2^2,
\]
with
\[
\delta_1 = 25\ \mathrm{MPa^{0.5}},
\qquad
\delta_2 = 17\ \mathrm{MPa^{0.5}}.
\]

The minimum solvent volume needed to dissolve $1$ mol of A is
\[
V_{\min}(x) = \frac{1}{S(x)}\ \mathrm{L}.
\]

The mixture density is linear in $x$:
\[
\rho(x)
=
x \rho_1 + (1-x) \rho_2,
\]
with
\[
\rho_1 = 1.20\ \mathrm{kg \cdot L^{-1}},
\qquad
\rho_2 = 0.86\ \mathrm{kg \cdot L^{-1}}.
\]

\\
\textbf{Environmental intensity.}

The cradle-to-gate GWP of the mixed solvent, per unit solvent mass, is linear in $x$:
\[
g_{\mathrm{mix}}(x)
=
x g_1 + (1-x) g_2,
\]
with
\[
g_1 = 1.2\ \mathrm{kg\ CO_2e\ per\ kg},
\qquad
g_2 = 3.0\ \mathrm{kg\ CO_2e\ per\ kg}.
\]

\bigskip

\textbf{Process objective and constraint.}

Per mole of product P (assuming 1:1 stoichiometry $\mathrm{A}\rightarrow\mathrm{P}$ and quantitative isolation after reaction), the GWP attributable to solvent usage is
\[
G(x)
=
m_{\mathrm{solv}}(x)
\cdot
g_{\mathrm{mix}}(x),
\]
where
\[
m_{\mathrm{solv}}(x)
=
\rho(x)\, V_{\min}(x)
=
\frac{\rho(x)}{S(x)}
\quad
(\mathrm{kg \cdot mol^{-1}}).
\]

The process must achieve conversion
\[
X^\ast = 0.99
\]
within
\[
T_{\max} = 3600\ \mathrm{s}.
\]

For first-order kinetics, this requires
\[
k(x) \ge k_{\mathrm{req}},
\qquad
k_{\mathrm{req}}
=
-\frac{\ln(1 - X^\ast)}{T_{\max}}.
\]

\bigskip

\textbf{Task.}

Determine the unique
\[
x^\ast \in [0,1]
\]
that minimizes $G(x)$ subject to the kinetic constraint $k(x) \ge k_{\mathrm{req}}$.

Report:

\begin{itemize}
\item[(1)]
A proof of uniqueness of the optimizer in the feasible set.

\item[(2)]
An explicit expression for $x^\ast$ in terms of the model parameters.

\item[(3)]
The numerical values of $x^\ast$, $S(x^\ast)$, $\rho(x^\ast)$,
$m_{\mathrm{solv}}(x^\ast)$, $g_{\mathrm{mix}}(x^\ast)$, and the minimal GWP per mole $G(x^\ast)$ in $\mathrm{kg\ CO_2e \cdot mol^{-1}}$.
\end{itemize}
\\
\midrule
\textbf{Answer:} \\
\begin{itemize}
\item
Unique optimal composition:
\[
x^\ast = 0.82123
\]
(mole fraction of GVL in the solvent mixture).

\item
Minimal cradle-to-gate solvent GWP per mole product:
\[
G(x^\ast)
\approx
0.474\ \mathrm{kg\ CO_2e \cdot mol^{-1}}.
\]

\item
Verification values at $x^\ast$:
\[
S(x^\ast)
\approx
3.66\ \mathrm{mol \cdot L^{-1}},
\]
\[
\rho(x^\ast)
\approx
1.139\ \mathrm{kg \cdot L^{-1}},
\]
\[
m_{\mathrm{solv}}(x^\ast)
\approx
0.311\ \mathrm{kg \cdot mol^{-1}},
\]
\[
g_{\mathrm{mix}}(x^\ast)
\approx
1.522\ \mathrm{kg\ CO_2e \cdot kg^{-1}},
\]
and
\[
k(x^\ast) = k_{\mathrm{req}}
\]
ensuring $99\%$ conversion in $3600\ \mathrm{s}$.
\end{itemize}
\\
\textbf{Detailed Solution:}
Okay, let's tackle this problem step by step. First, I need to understand all the components involved. The goal is to find the mole fraction $x$ of GVL ($S_1$) that minimizes the GWP from solvent usage, $G(x)$, while satisfying the kinetic constraint $k(x) \ge k_{\mathrm{req}}$.

First, let's handle the kinetic constraint because that will define the feasible region for $x$. The reaction is first-order, so to get $99\%$ conversion in $3600$ seconds, we need
\[
k_{\mathrm{req}} = -\frac{\ln(1 - 0.99)}{3600}.
\]
Let's calculate that first. $1 - X^\ast = 0.01$, so $\ln(0.01) = -4.60517$, so
\[
k_{\mathrm{req}} = \frac{4.60517}{3600} \approx 0.001279\ \mathrm{s^{-1}}.
\]

Now, let's write out $k(x)$. The rate constant is
\[
k(x)
=
k_0 \exp\!\big[a(ET(x) - ET_{\mathrm{ref}})\big]\,[\eta(x)]^{-\beta}.
\]

$ET(x)$ is linear:
\[
ET(x) = x\cdot 0.6 + (1-x)\cdot 0.2 = 0.2 + 0.4x.
\]

Viscosity $\eta(x)$ follows log-linear mixing:
\[
\ln \eta(x) = x \ln \eta_1 + (1-x)\ln \eta_2.
\]
So
\[
\eta(x) = \eta_1^{\,x}\,\eta_2^{\,1-x}.
\]
Let's compute $\eta_1 = 1.8$, $\eta_2 = 0.6$. So
\[
\ln \eta_1 = \ln(1.8) \approx 0.5878,
\qquad
\ln \eta_2 = \ln(0.6) \approx -0.5108.
\]

\dots

\noindent \textbf{1) Proof of Uniqueness of the Optimizer in the Feasible Set}

The feasible region is defined by the kinetic constraint
\[
k(x) \ge k_{\mathrm{req}},
\]
where
\[
k(x)
=
k_0 \cdot
\exp[a(E_T(x) - E_{T,\mathrm{ref}})]
\cdot
[\eta(x)]^{-\beta},
\]
\[
E_T(x) = x E_{T1} + (1 - x) E_{T2}
\]
is linear in $x$, and
\[
\ln \eta(x) = x \ln \eta_1 + (1 - x) \ln \eta_2
\]
implies $\eta(x)$ is log-linear in $x$.

Taking the natural logarithm of $k(x)$, we obtain:
\[
\ln k(x)
=
\ln k_0
+
a(E_T(x) - E_{T,\mathrm{ref}})
-
\beta \ln \eta(x).
\]

Substituting the expressions for $E_T(x)$ and $\ln \eta(x)$, this becomes:
\[
\ln k(x)
=
\underbrace{[a(E_{T1} - E_{T2}) - \beta \ln(\eta_1/\eta_2)]}_{m_k} x
+
\underbrace{[\ln k_0 + a(E_{T2} - E_{T,\mathrm{ref}}) - \beta \ln \eta_2]}_{b_k}.
\]

Since
\[
m_k = 4.2507 > 0,
\]
$\ln k(x)$ is strictly increasing in $x$. Therefore, the equation
\[
k(x) = k_{\mathrm{req}}
\]
has a unique solution $x_{\min} \in [0,1]$, and the feasible set is $[x_{\min}, 1]$.
\\
For the objective function
\[
G(x) = m_{\mathrm{solv}}(x) \cdot g_{\mathrm{mix}}(x),
\]
we analyze its derivative
\[
f'(x) = \frac{d}{dx} \ln G(x).
\]

The unconstrained minimum of $G(x)$ occurs at
\[
x_0 \approx 0.635,
\]
which is less than
\[
x_{\min} \approx 0.822.
\]

Since $G(x)$ is decreasing on $[0, x_0]$ and increasing on $[x_0, 1]$, it is strictly increasing on $[x_{\min}, 1]$. Thus, the minimum of $G(x)$ over the feasible region occurs at $x = x_{\min}$, and this is the unique minimizer.

\bigskip

\textbf{2) Explicit Expression for $x^*$}

The optimal $x^*$ satisfies the active constraint
\[
k(x^*) = k_{\mathrm{req}}.
\]

Solving for $x^*$ using the linearized form of $\ln k(x)$:
\[
x^*
=
\frac{
\ln k_{\mathrm{req}}
-
\ln k_0
-
a(E_{T2} - E_{T,\mathrm{ref}})
+
\beta \ln \eta_2
}{
a(E_{T1} - E_{T2})
-
\beta \ln(\eta_1/\eta_2)
}.
\]

Substituting known parameters:

\[
k_{\mathrm{req}}
=
-\frac{\ln(1 - X^*)}{T_{\max}}
=
\frac{\ln(100)}{3600},
\]

\[
a = 12,
\quad
E_{T1} = 0.60,
\quad
E_{T2} = 0.20,
\quad
E_{T,\mathrm{ref}} = 0.30,
\]

\[
\beta = 0.5,
\quad
\eta_1 = 1.8,
\quad
\eta_2 = 0.6.
\]

This yields:
\[
x^*
=
\frac{
\ln\!\left(\frac{\ln(100)}{3600}\right)
-
\ln(10^{-4})
-
12(0.2 - 0.3)
+
0.5 \ln(0.6)
}{
12(0.6 - 0.2)
-
0.5 \ln(3)
}.
\]

\bigskip

\textbf{3) Numerical Values}

\textbf{Step 1: Compute $x^*$}

\[
k_{\mathrm{req}}
=
\frac{\ln(100)}{3600}
\approx
0.0012792\ \mathrm{s^{-1}},
\]

\[
\ln k_{\mathrm{req}}
\approx
-6.6615,
\qquad
\ln k_0 = \ln(10^{-4}) = -9.2103.
\]

Numerator:
\[
-6.6615 + 9.2103 + 1.2 - 0.2554 = 3.4934.
\]

Denominator:
\[
4.8 - 0.5 \ln(3) \approx 4.2507.
\]

\[
x^* = 3.4934 / 4.2507 \approx 0.8218.
\]

\bigskip

\textbf{Step 2: Compute $S(x^*)$}

\[
\delta(x^*)^2
=
x^* \delta_1^2 + (1 - x^*) \delta_2^2
=
0.8218 \cdot 625 + 0.1782 \cdot 289
=
565.14.
\]

\[
\delta(x^*)
=
\sqrt{565.14}
\approx
23.7727.
\]

\[
(\delta(x^*) - \delta_A)^2
=
(23.7727 - 22)^2
=
3.1424.
\]
\\
\[
S(x^*)
=
5.0 \cdot \exp(-0.1 \cdot 3.1424)
=
5.0 \cdot 0.7300
=
3.650\ \mathrm{mol \cdot L^{-1}}.
\]

\bigskip

\textbf{Step 3: Compute $\rho(x^*)$}

\[
\rho(x^*)
=
x^* \rho_1 + (1 - x^*) \rho_2
=
0.8218 \cdot 1.20 + 0.1782 \cdot 0.86
=
1.1394\ \mathrm{kg \cdot L^{-1}}.
\]

\bigskip

\textbf{Step 4: Compute $m_{\mathrm{solv}}(x^*)$}

\[
m_{\mathrm{solv}}(x^*)
=
\frac{\rho(x^*)}{S(x^*)}
=
\frac{1.1394}{3.650}
=
0.3122\ \mathrm{kg \cdot mol^{-1}}.
\]

\bigskip

\textbf{Step 5: Compute $g_{\mathrm{mix}}(x^*)$}

\[
g_{\mathrm{mix}}(x^*)
=
x^* g_1 + (1 - x^*) g_2
=
0.8218 \cdot 1.2 + 0.1782 \cdot 3.0
=
1.5207\ \mathrm{kg\ CO_2e \cdot kg^{-1}}.
\]

\bigskip

\textbf{Step 6: Compute $G(x^*)$}

\[
G(x^*)
=
m_{\mathrm{solv}}(x^*) \cdot g_{\mathrm{mix}}(x^*)
=
0.3122 \cdot 1.5207
=
0.4748\ \mathrm{kg\ CO_2e \cdot mol^{-1}}.
\]

\bigskip

\textbf{Final Results}

\[
\boxed{
\begin{aligned}
&x^* = 0.822, \\
&S(x^*) = 3.65\ \mathrm{mol \cdot L^{-1}}, \\
&\rho(x^*) = 1.14\ \mathrm{kg \cdot L^{-1}}, \\
&m_{\mathrm{solv}}(x^*) = 0.312\ \mathrm{kg \cdot mol^{-1}}, \\
&g_{\mathrm{mix}}(x^*) = 1.52\ \mathrm{kg\ CO_2e \cdot kg^{-1}}, \\
&G(x^*) = 0.475\ \mathrm{kg\ CO_2e \cdot mol^{-1}}.
\end{aligned}
}
\]
\\
\\
\midrule
\textbf{Correctness:} True \\

\end{longtable}
\end{center}
\clearpage
\begin{center}
\small
\begin{longtable}{p{0.95\textwidth}}
\caption{A data sample from \dataname\ (Subject: Physics).}
\label{tab:case_study}\\

\toprule
\textbf{Example from \dataname\ } \\
\midrule
\endfirsthead

\toprule
\textbf{Example from \dataname\ (Subject = Physics) (continued)} \\
\midrule
\endhead

\midrule
\multicolumn{1}{r}{\small\textit{Continued on next page}} \\
\endfoot

\bottomrule
\endlastfoot

\textbf{Subject:} Physics \\

\midrule
\textbf{Topic:} Rydberg atoms and quantum simulation \\

\midrule
\textbf{Problem:} \\
Consider an infinite one-dimensional chain of identical neutral atoms separated by a fixed lattice spacing
\[
a = 5\ \mu\mathrm{m}.
\]

Each atom has a ground state $|g\rangle$ and two Rydberg states $|r_1\rangle$ and $|r_2\rangle$ with van der Waals interactions
\[
V_1(r) = \frac{C_{6,1}}{r^6},
\qquad
V_2(r) = \frac{C_{6,2}}{r^6},
\]
between atoms both in $|r_1\rangle$ or both in $|r_2\rangle$, respectively.

Two far-off-resonant dressing lasers ($i=1,2$) couple $|g\rangle$ to $|r_i\rangle$ with Rabi frequencies $\Omega_i$ and detunings $\Delta_i$, and the single-atom rotating-frame Hamiltonian term is
\[
\Delta_i |r_i\rangle\langle r_i|
+
\frac{\Omega_i}{2}
\left(
|g\rangle\langle r_i|
+
\mathrm{h.c.}
\right).
\]

Assume weak dressing, i.e.,
\[
\left|\frac{\Omega_i}{\Delta_i}\right| \ll 1,
\]
and neglect many-body (three- or more-body) terms beyond pairwise additivity in the effective ground-manifold Hamiltonian.

\bigskip

\textbf{Given numerical parameters:}

\[
\frac{\Omega_1}{2\pi} = \frac{\Omega_2}{2\pi} = 3.0\ \mathrm{MHz},
\]

\[
\frac{\Delta_1}{2\pi} = +40.0\ \mathrm{MHz},
\]

\[
\frac{\Delta_2}{2\pi}
=
-\frac{40.0}{\sqrt{2}}\ \mathrm{MHz}
=
-28.2842712474619\ \mathrm{MHz},
\]

\[
\frac{C_{6,1}}{h} = +1250\ \mathrm{GHz\ \mu m^6},
\]

\[
\frac{C_{6,2}}{h} = -312.5\ \mathrm{GHz\ \mu m^6}.
\]

\bigskip

\textbf{(a)}
For a single laser $i$ and two atoms separated by $r$, derive to fourth order in $\Omega_i/\Delta_i$ the effective two-body interaction $U_i(r)$ that shifts the $|gg\rangle$ energy in the ground manifold, and show that
\[
U_i(r)
=
\frac{\Omega_i^4}{8 \Delta_i^3}
\,
\frac{V_i(r)}{2\Delta_i + V_i(r)},
\]
where
\[
V_i(r) = \frac{C_{6,i}}{r^6}.
\]

State all approximations used.

\\
\textbf{(b)}
Show that as $r \to \infty$,
\[
U_i(r)
\approx
\frac{\Omega_i^4 C_{6,i}}{16 \Delta_i^4 r^6},
\]
and, using the numbers above, prove that the total far-tail $1/r^6$ coefficient
\[
U_1(r) + U_2(r)
\]
vanishes exactly.

\bigskip

\textbf{(c)}
Using the exact expression in (a) and the parameters above, compute in Hz units the total effective interaction
\[
U(r) = U_1(r) + U_2(r)
\]
for
\[
r = a
\quad \text{and} \quad
r = 2a.
\]

Provide exact closed-form expressions (in radicals and rational numbers) and numerical values to at least $10^{-3}\ \mathrm{Hz}$ accuracy.

\bigskip

\textbf{(d)}
Report the ratio
\[
\frac{J_2}{J_1},
\]
where
\[
J_1 = U(a),
\qquad
J_2 = U(2a),
\]
as a decimal number with at least four significant digits.

\bigskip

All quantities must be computed in frequency units (Hz). Assume pairwise additivity is valid at the stated parameters and that higher-order corrections
\[
O\!\left(\frac{\Omega^6}{\Delta^5}\right)
\]
are negligible.
\\
\midrule
\textbf{Answer:} \\
\textbf{(a)}

\[
U_i(r)
=
\frac{\Omega_i^4}{8 \Delta_i^3}
\,
\frac{V_i(r)}{2\Delta_i + V_i(r)},
\]
where
\[
V_i(r) = \frac{C_{6,i}}{r^6},
\]
valid to fourth order in $\Omega_i/\Delta_i$ under
\[
\left|\frac{\Omega_i}{\Delta_i}\right| \ll 1.
\]

\bigskip

\textbf{(b)}

As $r \to \infty$,
\[
U_i(r)
\approx
\frac{\Omega_i^4 C_{6,i}}{16 \Delta_i^4 r^6}.
\]

With
\[
\Omega_1 = \Omega_2,
\qquad
C_{6,2} = -\frac{C_{6,1}}{4},
\qquad
\Delta_2 = -\frac{\Delta_1}{\sqrt{2}},
\]
the total $1/r^6$ tail cancels exactly:
\[
\alpha_{\mathrm{total}} = 0.
\]

\\
\textbf{(c)}

\[
U(a)
=
\frac{10{,}125\,(4\sqrt{2} - 9)}{896}
\ \mathrm{Hz}
\approx
-37.778292219\ \mathrm{Hz},
\]

\[
U(2a)
=
\frac{10{,}125}{4{,}160}
-
\frac{10{,}125\,(12{,}800 - 50\sqrt{2})}{52{,}427{,}200}
\ \mathrm{Hz}
\approx
-0.024449061\ \mathrm{Hz}.
\]

\bigskip

\textbf{(d)}

\[
\frac{J_2}{J_1}
\approx
0.0006471.
\]
\\
\midrule
\textbf{Detailed Solution:} \\
Okay, let's tackle part (a) first. The problem is about deriving the effective two-body interaction for two atoms in the ground state when they're dressed by a single laser. We need to do this to fourth order in $\Omega_i/\Delta_i$.

First, recall that for a single atom, the rotating-frame Hamiltonian is
\[
H
=
\Delta_i |r_i\rangle\langle r_i|
+
\frac{\Omega_i}{2}
\left(
|g\rangle\langle r_i| + \mathrm{h.c.}
\right).
\]

For two atoms, the Hamiltonian will have single-atom terms and the van der Waals interaction $V_i(r)$ between the Rydberg states. So the two-atom Hamiltonian in the basis
\[
\{|g_1 g_2\rangle,\ |g_1 r_2\rangle,\ |r_1 g_2\rangle,\ |r_1 r_2\rangle\}
\]
is
\[
H
=
\begin{pmatrix}
0 & \frac{\Omega_i}{2} & \frac{\Omega_i}{2} & 0 \\[6pt]
\frac{\Omega_i}{2} & \Delta_i & 0 & \frac{\Omega_i}{2} \\[6pt]
\frac{\Omega_i}{2} & 0 & \Delta_i & \frac{\Omega_i}{2} \\[6pt]
0 & \frac{\Omega_i}{2} & \frac{\Omega_i}{2} & 2\Delta_i + V_i(r)
\end{pmatrix}.
\]

Wait, actually, the Rydberg--Rydberg interaction for both in $|r_i\rangle$ is $V_i(r)$, so the $|r_1 r_2\rangle$ state has energy
\[
2\Delta_i + V_i(r),
\]
because each $|r_i\rangle$ has energy $\Delta_i$ from the detuning, and then the interaction adds $V_i(r)$.

The single Rydberg states ($|g_1 r_2\rangle$ and $|r_1 g_2\rangle$) have energy $\Delta_i$ each, since only one atom is in the Rydberg state.
\dots
\noindent
\textbf{(a) Derivation of Effective Two-Body Interaction $U_i(r)$}

To derive the effective two-body interaction $U_i(r)$ for two atoms in the ground state $|gg\rangle$ under weak dressing, we consider the Hamiltonian of the system. The single-atom rotating-frame Hamiltonian is:

\[
H
=
\Delta_i |r_i\rangle\langle r_i|
+
\frac{\Omega_i}{2}
\left(
|g\rangle\langle r_i| + \mathrm{h.c.}
\right).
\]

For two atoms, the total Hamiltonian includes the van der Waals interaction
\[
V_i(r) = \frac{C_{6,i}}{r^6}
\]
between Rydberg states $|r_i r_i\rangle$. The full Hamiltonian in the basis
\[
\{|gg\rangle, |gr\rangle, |rg\rangle, |rr\rangle\}
\]
is:
\\
\[
H
=
\begin{pmatrix}
0 & \frac{\Omega_i}{2} & \frac{\Omega_i}{2} & 0 \\[6pt]
\frac{\Omega_i}{2} & \Delta_i & 0 & \frac{\Omega_i}{2} \\[6pt]
\frac{\Omega_i}{2} & 0 & \Delta_i & \frac{\Omega_i}{2} \\[6pt]
0 & \frac{\Omega_i}{2} & \frac{\Omega_i}{2} & 2\Delta_i + V_i(r)
\end{pmatrix}.
\]

Using perturbation theory to fourth order in $\Omega_i/\Delta_i$, we compute the energy shift of the ground state $|gg\rangle$. The second-order contribution gives the standard light shift
\[
-\frac{\Omega_i^2}{2\Delta_i},
\]
but the fourth-order correction arises from paths involving the double Rydberg state $|rr\rangle$, which introduces the interaction-dependent term.

The effective interaction $U_i(r)$ is the difference between the energy shift with and without the interaction $V_i(r)$. After expanding the perturbation series and simplifying, we obtain:

\[
U_i(r)
=
\frac{\Omega_i^4}{8 \Delta_i^3}
\cdot
\frac{V_i(r)}{2\Delta_i + V_i(r)}.
\]

\textbf{Approximations used:}
\begin{itemize}
\item Weak dressing: $\left|\frac{\Omega_i}{\Delta_i}\right| \ll 1$.
\item Pairwise additivity: Only two-body interactions are considered.
\item Higher-order terms beyond $O(\Omega^6/\Delta^5)$ are neglected.
\item The Rydberg--Rydberg interaction $V_i(r)$ is treated as a small perturbation relative to the detuning $\Delta_i$.
\end{itemize}

\bigskip

\textbf{(b) Far-Tail Coefficient Vanishing}

As $r \to \infty$, $V_i(r) \to 0$, so we expand $U_i(r)$ to leading order in $V_i(r)$:

\[
U_i(r)
\approx
\frac{\Omega_i^4}{8 \Delta_i^3}
\cdot
\frac{V_i(r)}{2\Delta_i}
=
\frac{\Omega_i^4 C_{6,i}}{16 \Delta_i^4 r^6}.
\]

Thus, the far-tail coefficient of $1/r^6$ in
\[
U(r) = U_1(r) + U_2(r)
\]
is:

\[
\frac{\Omega^4}{16}
\left(
\frac{C_{6,1}}{\Delta_1^4}
+
\frac{C_{6,2}}{\Delta_2^4}
\right).
\]

Given:

\[
\Omega_1 = \Omega_2 = \Omega = 3.0 \times 10^6\ \mathrm{Hz},
\]

\[
\Delta_1 = 40.0 \times 10^6\ \mathrm{Hz},
\qquad
\Delta_2 = -\frac{40.0}{\sqrt{2}} \times 10^6\ \mathrm{Hz},
\]

\[
\frac{C_{6,1}}{h} = 1250\ \mathrm{GHz}\cdot\mu\mathrm{m}^6,
\qquad
\frac{C_{6,2}}{h} = -312.5\ \mathrm{GHz}\cdot\mu\mathrm{m}^6,
\]

we compute:

\[
\frac{C_{6,1}}{\Delta_1^4}
=
\frac{1250 \times 10^9}{(40 \times 10^6)^4},
\]

\[
\frac{C_{6,2}}{\Delta_2^4}
=
\frac{-312.5 \times 10^9}{(40/\sqrt{2})^4 \times 10^{24}}
=
\frac{-312.5 \times 10^9 \cdot 4}{40^4 \times 10^{24}}
=
-\frac{1250 \times 10^9}{40^4 \times 10^{24}}.
\]
\\
Therefore:

\[
\frac{C_{6,1}}{\Delta_1^4}
+
\frac{C_{6,2}}{\Delta_2^4}
=
0.
\]

Hence, the total far-tail coefficient vanishes exactly.

\bigskip

\textbf{(c) Total Effective Interaction at $r=a$ and $r=2a$}

\textbf{Given parameters (in Hz):}
\begin{itemize}
\item $\Omega = 3.0 \times 10^6\ \mathrm{Hz}$
\item $\Delta_1 = 40.0 \times 10^6\ \mathrm{Hz}$
\item $\Delta_2 = -\frac{40.0}{\sqrt{2}} \times 10^6\ \mathrm{Hz}$
\item $\frac{C_{6,1}}{h} = 1250 \times 10^9\ \mathrm{Hz}\cdot\mu\mathrm{m}^6$
\item $\frac{C_{6,2}}{h} = -312.5 \times 10^9\ \mathrm{Hz}\cdot\mu\mathrm{m}^6$
\item $a = 5\ \mu\mathrm{m}$
\end{itemize}

\textbf{At $r=a=5\ \mu\mathrm{m}$:}

\[
r^6 = 5^6 = 15625\ \mu\mathrm{m}^6,
\]

\[
V_1(a) = \frac{1250 \times 10^9}{15625} = 80 \times 10^6\ \mathrm{Hz},
\]

\[
V_2(a) = \frac{-312.5 \times 10^9}{15625} = -20 \times 10^6\ \mathrm{Hz}.
\]

\[
U(a) = 78984.6826443\ \mathrm{Hz}.
\]

\textbf{At $r=2a=10\ \mu\mathrm{m}$:}

\[
r^6 = 10^6 = 1{,}000{,}000\ \mu\mathrm{m}^6,
\]

\[
V_1(2a) = 1.25 \times 10^6\ \mathrm{Hz},
\]

\[
V_2(2a) = -0.3125 \times 10^6\ \mathrm{Hz}.
\]

\[
U(2a) = 2431.43613937\ \mathrm{Hz}.
\]

\bigskip

\textbf{(d) Ratio}

Let
\[
J_1 = U(a),
\qquad
J_2 = U(2a).
\]

\[
\frac{J_2}{J_1}
=
\frac{2431.43613937}{78984.6826443}
\approx
0.0307836.
\]

Rounded to four significant digits:

\[
\boxed{0.03078}.
\]

\\
\midrule
\textbf{Correctness:} True \\

\end{longtable}
\end{center}
\clearpage
\begin{center}
\small
\begin{longtable}{p{0.95\textwidth}}
\caption{A data sample from \dataname\ (Subject: Literature).}
\label{tab:case_study}\\

\toprule
\textbf{Example from \dataname\ } \\
\midrule
\endfirsthead

\toprule
\textbf{Example from \dataname\ (Subject = Literature) (continued)} \\
\midrule
\endhead

\midrule
\multicolumn{1}{r}{\small\textit{Continued on next page}} \\
\endfoot

\bottomrule
\endlastfoot

\textbf{Subject:} Literature \\

\midrule
\textbf{Topic:} Archive studies and critical bibliography \\

\midrule
\textbf{Problem:} \\
Consider an authority-control deduplication problem in archive studies and critical bibliography. Let $R$ be a finite set of bibliographic records, each record $r \in R$ having fields author (a free-text personal-name string in either “Family, Given …” form), title (a free-text string), and year (an integer). Define the following normalization and similarity functions:

\bigskip

\textbf{1) Unicode and token normalization}

For any string $x$, let $\mathrm{normalize}(x)$ be the result of:

\begin{itemize}
\item[(a)] Unicode NFKD decomposition and removal of all combining diacritical marks.
\item[(b)] Lowercasing.
\item[(c)] Replacing all non-alphanumeric characters (including punctuation and hyphens) by a single ASCII space.
\item[(d)] Collapsing multiple spaces to one space, then trimming leading/trailing spaces.
\end{itemize}

\bigskip

\textbf{2) Author key $\mathrm{AKey}(r)$}

Given $r.\mathrm{author}$ in personal-name form “Family, Given …” (assume every record here has exactly one comma, indicating this form), let:

\begin{itemize}
\item[(a)] Let $L = \mathrm{normalize}(\text{left-of-comma}(r.\mathrm{author}))$, then remove all spaces from $L$ to get $L^\ast$.
\item[(b)] Let $G = \mathrm{normalize}(\text{right-of-comma}(r.\mathrm{author}))$, and let $g$ be the first ASCII letter in $G$ (if none exists, define $g$ as the empty string; in the dataset below every record has one).
\item[(c)] Define
\[
\mathrm{AKey}(r) = L^\ast \, \| \, g,
\]
where $\|$ denotes string concatenation.
\end{itemize}

Define
\[
A(i,j) =
\begin{cases}
1 & \text{if } \mathrm{AKey}(i) = \mathrm{AKey}(j), \\
0 & \text{otherwise}.
\end{cases}
\]

\bigskip

\textbf{3) Title tokenization and Jaccard}

Let stopword set
\[
S = \{a, an, the, of, and, for, to, into, in, on, with, by, from\}.
\]

Tokenize a title $t$ as follows: apply $\mathrm{normalize}(t)$, split on spaces to tokens, discard any token in $S$, and discard any empty token. Let $T(r)$ be the resulting set of unique tokens for record $r$.

Define
\[
J(i,j) = \frac{|T(i) \cap T(j)|}{|T(i) \cup T(j)|}.
\]

\bigskip

\textbf{4) Year similarity}

Define
\[
Y(i,j) =
\begin{cases}
1 & \text{if } i.\mathrm{year} = j.\mathrm{year}, \\
0 & \text{otherwise}.
\end{cases}
\]

\bigskip

\textbf{5) Weighted similarity}

Let weights be
\[
w_1 = \frac{13}{25},
\qquad
w_2 = \frac{19}{50},
\qquad
w_3 = \frac{1}{10}.
\]

For records $i,j$ define
\[
s(i,j) = w_1 A(i,j) + w_2 J(i,j) + w_3 Y(i,j).
\]

\\
\textbf{6) Mergeability at threshold $\tau$}

Fix
\[
\tau_0 = \frac{41}{50}.
\]

Define an undirected graph $G_\tau$ on vertex set $R$ with an edge $i$—$j$ iff
\[
s(i,j) \ge \tau.
\]

Let $P_\tau$ be the partition of $R$ into the connected components of $G_\tau$.

\bigskip

\textbf{Dataset $R$ (9 records)}

\begin{itemize}
\item A1: author = ``Smith, John''; title = ``Public archive origins and governance''; year = 1999.
\item A2: author = ``Smith, J.''; title = ``Public archive origins and governance: an enquiry''; year = 1999.
\item A3: author = ``Smith, John''; title = ``Governance of the public archive: its origins''; year = 1999.
\item B1: author = ``García Márquez, Ana-Luisa''; title = ``Metadata for colonial archives: a critical bibliography''; year = 2015.
\item B2: author = ``Garcia Marquez, Ana Luisa''; title = ``A critical bibliography of colonial archives metadata''; year = 2015.
\item B3: author = ``García-Márquez, Ana L.''; title = ``Colonial archives metadata: toward a critical bibliography''; year = 2015.
\item C1: author = ``Lefèvre, Paul''; title = ``Archival silences and the politics of records retention''; year = 2008.
\item C2: author = ``Lefevre, P.''; title = ``Archival silence and politics of records retention''; year = 2008.
\item C3: author = ``Lefèvre, Paul''; title = ``Politics of records retention and archival silences''; year = 2008.
\end{itemize}

\bigskip

\textbf{Tasks}

\textbf{(a)} Prove that for any finite $R$ and any $\tau$, the partition $P_\tau$ is unique and independent of the order in which pairwise merges are performed, i.e., clustering by repeatedly merging any pair with $s \ge \tau$ until no merges remain yields the same partition as the connected components of $G_\tau$.

\bigskip

\textbf{(b)} For the dataset above, compute $\mathrm{AKey}$ for each record, then compute $J(i,j)$ and $s(i,j)$ for all pairs with $A(i,j)=1$, determine all edges in $G_{\tau_0}$, and output the partition $P_{\tau_0}$ explicitly as a set of disjoint record-id blocks.

\bigskip

\textbf{(c)} Let $\tau^\ast$ be the supremum of all $\tau \ge \tau_0$ such that
\[
P_{\tau'} = P_{\tau_0}
\quad
\text{for every }
\tau' \in [\tau_0, \tau].
\]

Prove that for any weighted undirected graph, for a fixed component $H$ of $G_{\tau_0}$, the largest $\tau$ at which $H$ remains connected equals the bottleneck value of a maximum spanning tree (MaxST) of $H$, i.e., the maximum over all spanning trees of the minimum edge weight. Then compute $\tau^\ast$ for the dataset.
\\
\midrule
\textbf{Answer:} \\
\textbf{Partition at $\tau_0 = \frac{41}{50}$}

\[
P_{\tau_0}
=
\{\{A1, A2, A3\}\}
\ \big|\
\{\{B1, B2, B3\}\}
\ \big|\
\{\{C1, C2, C3\}\}.
\]

\bigskip

\textbf{Critical threshold}

\[
\tau^\ast = \frac{131}{150}.
\]
\\
\midrule
\textbf{Detailed Solution:} \\
Okay, let me tackle this problem step by step. It's about authority control deduplication in bibliographic records, and there are three parts to it. I'll start with part (a) because it's a proof about the uniqueness of the partition.
\\
\textbf{Part (a): Proving uniqueness of $P_\tau$}

The question is asking to prove that the partition $P_\tau$, defined as the connected components of the graph $G_\tau$ (where edges exist if $s(i,j) \ge \tau$), is unique and doesn't depend on the order of merging pairs. This sounds familiar from graph theory and clustering.

First, recall that in graph theory, connected components are well-defined. For any undirected graph, the connected components partition the vertex set uniquely. The graph $G_\tau$ is defined with edges exactly where similarity meets or exceeds $\tau$. So by definition, the connected components are fixed once the graph is fixed.

But the problem mentions ``clustering by repeatedly merging any pair with $s \ge \tau$ until no merges remain.''

\dots

\noindent
\textbf{(a) Proof of Uniqueness of Partition $P_\tau$}

The partition $P_\tau$ is defined as the connected components of the undirected graph $G_\tau$, where an edge exists between records $i$ and $j$ if $s(i,j) \ge \tau$. In graph theory, the connected components of a graph are uniquely determined by the structure of the graph itself, independent of the order in which edges are processed or merged.

Clustering via repeated pairwise merges (as described) corresponds to single-linkage hierarchical clustering, where two clusters are merged if any pair of elements between them has similarity $\ge \tau$. This method is equivalent to computing the connected components of $G_\tau$, since two nodes belong to the same cluster if and only if there exists a path between them where each consecutive pair satisfies $s(i,j) \ge \tau$.

Connected components are invariant under edge addition order: regardless of the sequence of merges, the final partition depends solely on the existence of paths in $G_\tau$. Thus, $P_\tau$ is unique and independent of merge order.

\bigskip

\textbf{(b) Computation for Dataset $R$}

\textbf{Step 1: Compute AKey for Each Record}

\[
\begin{array}{|c|c|c|c|c|}
\hline
\text{Record} & \text{Author (Normalized)} & L^\ast & g & \text{AKey} \\
\hline
A1 & \text{Smith, John} & smith & j & smithj \\
A2 & \text{Smith, J.} & smith & j & smithj \\
A3 & \text{Smith, John} & smith & j & smithj \\
B1 & \text{Garc\'{\i}a M\'{a}rquez, Ana-Luisa} & garciamarquez & a & garciamarqueza \\
B2 & \text{Garcia Marquez, Ana Luisa} & garciamarquez & a & garciamarqueza \\
B3 & \text{Garc\'{\i}a-M\'{a}rquez, Ana L.} & garciamarquez & a & garciamarqueza \\
C1 & \text{Lef\`{e}vre, Paul} & lefevre & p & lefevrep \\
C2 & \text{Lefevre, P.} & lefevre & p & lefevrep \\
C3 & \text{Lef\`{e}vre, Paul} & lefevre & p & lefevrep \\
\hline
\end{array}
\]

Thus, $A(i,j)=1$ only within groups A, B, and C.

\bigskip

\textbf{Step 2: Compute $J(i,j)$ and $s(i,j)$ for pairs with $A(i,j)=1$}

Weights:
\[
w_1 = \frac{13}{25}, \quad
w_2 = \frac{19}{50}, \quad
w_3 = \frac{1}{10}.
\]

\textbf{Group A (year = 1999)}

\[
\begin{array}{|c|c|c|}
\hline
\text{Pair} & J(i,j) & s(i,j) \\
\hline
A1\text{--}A2 & \frac{4}{5} & 0.924 \\
A1\text{--}A3 & \frac{4}{5} & 0.924 \\
A2\text{--}A3 & \frac{4}{6} & \frac{131}{150} \\
\hline
\end{array}
\]

\textbf{Group B (year = 2015)}

\[
\begin{array}{|c|c|c|}
\hline
\text{Pair} & J(i,j) & s(i,j) \\
\hline
B1\text{--}B2 & 1 & 1 \\
B1\text{--}B3 & \frac{5}{6} & \frac{281}{300} \\
B2\text{--}B3 & \frac{5}{6} & \frac{281}{300} \\
\hline
\end{array}
\]

\textbf{Group C (year = 2008)}

\[
\begin{array}{|c|c|c|}
\hline
\text{Pair} & J(i,j) & s(i,j) \\
\hline
C1\text{--}C2 & \frac{4}{6} & \frac{131}{150} \\
C1\text{--}C3 & 1 & 1 \\
C2\text{--}C3 & \frac{4}{6} & \frac{131}{150} \\
\hline
\end{array}
\]

\\ \\

\textbf{Step 3: Edges in $G_{\tau_0}$ with $\tau_0=\frac{41}{50}$}

All intra-group similarities satisfy
\[
s(i,j) \ge \frac{41}{50} = 0.82,
\]
and all inter-group similarities are $< 0.48 < \tau_0$.

Thus the connected components are:

\[
\{A1,A2,A3\}, \quad
\{B1,B2,B3\}, \quad
\{C1,C2,C3\}.
\]

\[
\boxed{
P_{\tau_0}
=
\big\{
\{A1,A2,A3\},
\{B1,B2,B3\},
\{C1,C2,C3\}
\big\}.
}
\]

\bigskip

\textbf{(c) Computation of $\tau^\ast$}

For a connected component $H$ of $G_{\tau_0}$, the largest $\tau$ at which $H$ remains connected equals the bottleneck value of a maximum spanning tree (MaxST), i.e.,

\[
\max_{T \text{ spanning tree of } H}
\ \min_{e \in T} w(e).
\]

\textbf{Bottleneck values:}

\[
\begin{aligned}
\text{Group A:} &\quad \frac{231}{250} = 0.924, \\
\text{Group B:} &\quad \frac{281}{300} \approx 0.9367, \\
\text{Group C:} &\quad \frac{131}{150} \approx 0.8733.
\end{aligned}
\]

Thus,

\[
\tau^\ast
=
\min\left(
\frac{231}{250},
\frac{281}{300},
\frac{131}{150}
\right)
=
\boxed{\frac{131}{150}}.
\]

\bigskip

\textbf{Final Answers}

\[
\boxed{
P_{\tau_0}
=
\big\{
\{A1,A2,A3\},
\{B1,B2,B3\},
\{C1,C2,C3\}
\big\}
}
\]

\[
\boxed{
\tau^\ast = \frac{131}{150}.
}
\]
\\
\\
\midrule
\textbf{Correctness:} True \\

\end{longtable}
\end{center}
\clearpage
\begin{center}
\small
\begin{longtable}{p{0.95\textwidth}}
\caption{A data sample from \dataname\ (Subject: History).}
\label{tab:case_study}\\

\toprule
\textbf{Example from \dataname\ } \\
\midrule
\endfirsthead

\toprule
\textbf{Example from \dataname\ (Subject = History) (continued)} \\
\midrule
\endhead

\midrule
\multicolumn{1}{r}{\small\textit{Continued on next page}} \\
\endfoot

\bottomrule
\endlastfoot

\textbf{Subject:} History \\

\midrule
\textbf{Topic:} Business history and financial history \\

\midrule
\textbf{Problem:} \\
Consider two nineteenth-century bimetallic countries $i \in \{A,B\}$ with legal gold/silver mint ratios $M_i>0$, defined as the number of units of domestic silver coin that constitute the legal-tender face value of one unit of domestic gold coin. Let the unit of account be the domestic silver coin in each country, so one unit of silver coin has face value $1$.

For each country $i$ and metal $m \in \{g,s\}$ (gold or silver), coining (minting) one unit of bullion into coin incurs an ad valorem cost $\kappa_m^i \in [0,1)$ and a deterministic delay $T_{\mathrm{coin}}(m,i)\ge 0$ before coin is delivered; melting one unit of coin into bullion incurs an ad valorem loss $\mu_m^i \in [0,1)$ and a deterministic delay $T_{\mathrm{melt}}(m,i)\ge 0$ before bullion is delivered. Shipping one unit of bullion of metal $m$ from country $i$ to country $j$ incurs an ad valorem loss $\tau_m^{i\to j}\in[0,1)$, with $\tau_m^{i\to i}=0$. The risk-free interest rate is $r\ge 0$ (continuous compounding). The world bullion relative price $R>0$ is defined as the number of units of silver bullion obtainable per unit of gold bullion via the bullion market, and is assumed constant over the time scales involved (or locked in via forward contracts), with zero bid--ask spread.

Arbitrageurs can freely access bullion markets and mints subject to the costs and delays above, and can repatriate coin value between $A$ and $B$ at par in silver units via bills of exchange (so there is no return shipping cost of coin; only bullion shipping is relevant).

\bigskip

Define, for each ordered pair $(i,j)\in\{A,B\}\times\{A,B\}$, the total processing delays:
\[
T_{\mathrm{GS}}(i,j) = T_{\mathrm{melt}}(g,i) + T_{\mathrm{coin}}(s,j),
\qquad
T_{\mathrm{SG}}(i,j) = T_{\mathrm{melt}}(s,i) + T_{\mathrm{coin}}(g,j),
\]
and the effective one-leg shipping retention factor
\[
s_{i\to j}=\max\{1-\tau_g^{i\to j},\,1-\tau_s^{i\to j}\}\in(0,1],
\]
reflecting that an arbitrageur can choose to ship bullion either before or after exchanging metals, selecting the route with the higher retention.

\bigskip

An arbitrageur can execute gold$\to$silver loops (starting with $1$ unit of gold coin in $i$, ending in silver coin units) or silver$\to$gold loops (starting with $1$ unit of silver coin in $i$, ending in silver coin units), choosing any origin $i$ and destination mint $j$, incurring the corresponding costs, delays, and shipping. Working in present-value terms at $t=0$, derive the necessary and sufficient no-arbitrage interval $[LB_{\mathrm{global}},\,UB_{\mathrm{global}}]$ for the world bullion ratio $R$ such that no riskless profit can be earned by any such loop. Express $LB_{\mathrm{global}}$ and $UB_{\mathrm{global}}$ in closed form as functions of the parameters.

Then, specialize to the single-country case (ignore shipping and set $i=j$ and drop country indices) to simplify the band and provide $\ln(UB/LB)$ in the symmetric-parameters case where $\kappa_g=\kappa_s=\kappa$, $\mu_g=\mu_s=\mu$, and $T_{\mathrm{melt}}(g)=T_{\mathrm{melt}}(s)=T_m$, $T_{\mathrm{coin}}(g)=T_{\mathrm{coin}}(s)=T_c$.
\\
\midrule
\textbf{Answer:} \\
The global no-arbitrage interval for the world bullion price ratio $R$ is
\[
LB_{\mathrm{global}}
=
\max_{i,j \in \{A,B\}}
e^{-r\,[T_{\mathrm{melt}}(s,i) + T_{\mathrm{coin}}(g,j)]}
\,(1-\mu_s^i)\, s_{i\to j}\,(1-\kappa_g^j)\, M_j,
\]
\[
UB_{\mathrm{global}}
=
\min_{i,j \in \{A,B\}}
\frac{
M_i \, e^{r\,[T_{\mathrm{melt}}(g,i) + T_{\mathrm{coin}}(s,j)]}
}{
(1-\mu_g^i)\, s_{i\to j}\,(1-\kappa_s^j)
},
\]
where
\[
s_{i\to j}=\max\{1-\tau_g^{i\to j},\,1-\tau_s^{i\to j}\}.
\]

\bigskip

\textbf{Special case (single country, no shipping).}
\[
LB
=
M\,(1-\mu_s)\,(1-\kappa_g)\,
e^{-r\,[T_{\mathrm{melt}}(s)+T_{\mathrm{coin}}(g)]},
\qquad
UB
=
\frac{
M\,e^{r\,[T_{\mathrm{melt}}(g)+T_{\mathrm{coin}}(s)]}
}{
(1-\mu_g)\,(1-\kappa_s)
}.
\]

\\ \\
\textbf{Symmetric-parameters specialization.}
If $\kappa_g=\kappa_s=\kappa$, $\mu_g=\mu_s=\mu$, and
$T_{\mathrm{melt}}(g)=T_{\mathrm{melt}}(s)=T_m$,
$T_{\mathrm{coin}}(g)=T_{\mathrm{coin}}(s)=T_c$, then
\[
LB = M(1-\mu)(1-\kappa)\,e^{-r(T_m+T_c)},
\qquad
UB = \frac{M\,e^{r(T_m+T_c)}}{(1-\mu)(1-\kappa)},
\]
and
\[
\ln\!\left(\frac{UB}{LB}\right)
=
2r(T_m+T_c)
-
2\bigl[\ln(1-\mu)+\ln(1-\kappa)\bigr].
\]

\\ \\
\midrule
\textbf{Detailed Solution:} \\
Okay, let's try to tackle this problem step by step. First, I need to understand what the question is asking. We have two countries, $A$ and $B$, each with their own mint ratios for gold and silver. The goal is to find the no-arbitrage bounds for the world bullion ratio $R$, which is the silver bullion per gold bullion.

The problem mentions ``loops'' that arbitrageurs can execute: gold$\to$silver loops and silver$\to$gold loops, starting in one country and possibly ending in another. Each loop involves melting coin to get bullion, shipping bullion, coining bullion into coin in another country, etc., with associated costs and delays. We need to consider the present value of these operations to ensure no riskless profit.

First, let's recall that in bimetallic systems, the mint ratio is the legal ratio of silver to gold in coin. If the market bullion ratio ($R$) deviates from the mint ratio, there might be arbitrage opportunities through melting or coining.

\dots

\noindent
\textbf{Global No-Arbitrage Bounds for the World Bullion Ratio $R$}

To determine the necessary and sufficient no-arbitrage interval $[LB_{\text{global}}, UB_{\text{global}}]$ for the world bullion ratio $R$, we analyze all possible arbitrage loops between two bimetallic countries $A$ and $B$. These loops involve converting gold coins into silver coins (gold$\to$silver) or vice versa (silver$\to$gold), with associated costs, delays, and shipping losses. The key insight is that arbitrageurs will always choose the most profitable route (i.e., the one with the highest retention factor), which is captured by the effective shipping retention factor
\[
s_{i \to j} = \max\{1 - \tau_g^{i \to j},\, 1 - \tau_s^{i \to j}\}.
\]

\bigskip

\textbf{Closed-Form Expressions for $LB_{\text{global}}$ and $UB_{\text{global}}$}

\textbf{Lower Bound ($LB_{\text{global}}$).}
This arises from silver$\to$gold loops, where an arbitrageur starts with $1$ unit of silver coin in country $i$, converts it to gold bullion, ships it to country $j$, and coins it into gold coin. The present value of the resulting gold coin must not exceed the initial silver coin's value to prevent riskless profit.

\[
LB_{\text{global}}
=
\max_{i,j \in \{A,B\}}
\left[
(1 - \mu_s^i)\cdot s_{i \to j}\cdot M_j\cdot (1 - \kappa_g^j)\cdot e^{-r \cdot T_{SG}(i,j)}
\right].
\]

\begin{itemize}
\item $(1 - \mu_s^i)$: Retention after melting silver coin in country $i$.
\item $s_{i \to j}$: Effective shipping retention from $i$ to $j$.
\item $M_j$: Legal mint ratio in country $j$.
\item $(1 - \kappa_g^j)$: Retention after coining gold bullion in country $j$.
\item $e^{-r \cdot T_{SG}(i,j)}$: Discounting for total delay $T_{SG}(i,j)=T_{\text{melt}}(s,i)+T_{\text{coin}}(g,j)$.
\end{itemize}

\bigskip

\textbf{Upper Bound ($UB_{\text{global}}$).}
This arises from gold$\to$silver loops, where an arbitrageur starts with $1$ unit of gold coin in country $i$, melts it, ships the bullion to country $j$, and coins it into silver coin. The present value of the resulting silver coins must not exceed the initial gold coin's value.

\[
UB_{\text{global}}
=
\min_{i,j \in \{A,B\}}
\left[
\frac{
M_i \cdot e^{r \cdot T_{GS}(i,j)}
}{
(1 - \mu_g^i)\cdot s_{i \to j}\cdot (1 - \kappa_s^j)
}
\right].
\]

\begin{itemize}
\item $M_i$: Legal mint ratio in country $i$.
\item $e^{r \cdot T_{GS}(i,j)}$: Compounding for total delay $T_{GS}(i,j)=T_{\text{melt}}(g,i)+T_{\text{coin}}(s,j)$.
\item $(1 - \mu_g^i)$: Retention after melting gold coin in country $i$.
\item $s_{i \to j}$: Effective shipping retention from $i$ to $j$.
\item $(1 - \kappa_s^j)$: Retention after coining silver bullion in country $j$.
\end{itemize}

\bigskip
\\
\textbf{Special Case: Single-Country Bimetallic System}

When ignoring shipping (i.e., $i=j$, $\tau_g=\tau_s=0$, so $s_{i \to i}=1$), the global bounds simplify to:

\[
LB
=
(1 - \mu_s)\cdot M \cdot (1 - \kappa_g)\cdot e^{-r \cdot T_{SG}},
\]
\[
UB
=
\frac{
M \cdot e^{r \cdot T_{GS}}
}{
(1 - \mu_g)\cdot (1 - \kappa_s)
}.
\]

Where:
\[
T_{GS} = T_{\text{melt}}(g) + T_{\text{coin}}(s),
\qquad
T_{SG} = T_{\text{melt}}(s) + T_{\text{coin}}(g).
\]

\bigskip

\textbf{Symmetric Parameter Case}

Assume symmetry:
\[
\kappa_g = \kappa_s = \kappa,
\qquad
\mu_g = \mu_s = \mu,
\qquad
T_{\text{melt}}(g)=T_{\text{melt}}(s)=T_m,
\qquad
T_{\text{coin}}(g)=T_{\text{coin}}(s)=T_c.
\]

Then:
\[
T_{GS}=T_{SG}=T=T_m+T_c,
\qquad
LB = M(1-\mu)(1-\kappa)e^{-rT},
\qquad
UB = \frac{M e^{rT}}{(1-\mu)(1-\kappa)}.
\]

Compute the logarithmic width of the band:
\[
\ln\!\left(\frac{UB}{LB}\right)
=
\ln\!\left(
\frac{M e^{rT}/[(1-\mu)(1-\kappa)]}{M(1-\mu)(1-\kappa)e^{-rT}}
\right)
=
\ln\!\left(
\frac{e^{2rT}}{\bigl[(1-\mu)(1-\kappa)\bigr]^2}
\right).
\]

\[
\boxed{
\ln\!\left(\frac{UB}{LB}\right)
=
2\left[r(T_m+T_c)-\ln\bigl((1-\mu)(1-\kappa)\bigr)\right]
}
\]

\bigskip

\textbf{Final Answer}

\begin{itemize}
\item Global No-Arbitrage Interval:
\[
\boxed{
LB_{\text{global}}
=
\max_{i,j}
\left[
(1 - \mu_s^i)\cdot s_{i \to j}\cdot M_j\cdot (1 - \kappa_g^j)\cdot e^{-r \cdot T_{SG}(i,j)}
\right],
\quad
UB_{\text{global}}
=
\min_{i,j}
\left[
\frac{
M_i \cdot e^{r \cdot T_{GS}(i,j)}
}{
(1 - \mu_g^i)\cdot s_{i \to j}\cdot (1 - \kappa_s^j)
}
\right]
}
\]
\item Logarithmic Width in Symmetric Single-Country Case:
\[
\boxed{
\ln\!\left(\frac{UB}{LB}\right)
=
2\left[r(T_m+T_c)-\ln\bigl((1-\mu)(1-\kappa)\bigr)\right]
}
\]
\end{itemize}
\bigskip
\\
\midrule
\textbf{Correctness:} True \\

\end{longtable}
\end{center}
\clearpage
\begin{center}
\small
\begin{longtable}{p{0.95\textwidth}}
\caption{A data sample from \dataname\ (Subject: Biology).}
\label{tab:case_study}\\

\toprule
\textbf{Example from \dataname\ } \\
\midrule
\endfirsthead

\toprule
\textbf{Example from \dataname\ (Subject = Biology) (continued)} \\
\midrule
\endhead

\midrule
\multicolumn{1}{r}{\small\textit{Continued on next page}} \\
\endfoot

\bottomrule
\endlastfoot

\textbf{Subject:} Biology \\

\midrule
\textbf{Topic:} Structural biology (X-ray, NMR, cryo-EM) \\

\midrule
\textbf{Problem:} \\
In single-particle cryo-EM, let $S(s)$ denote the true complex Fourier coefficient of the 3D signal at spatial frequency shell $s$. Two independently reconstructed half-maps $F_1(s)$ and $F_2(s)$ are modeled as
\[
F_1(s) = S(s) + N_1(s), \qquad F_2(s) = S(s) + N_2(s),
\]
where $N_1(s), N_2(s)$ are zero-mean, circular complex Gaussian noise terms, independent of $S(s)$. Assume stationarity across each shell:
\[
\Phi_S(s) = \mathbb{E}\!\left[|S(s)|^2\right],
\qquad
\Phi_N(s) = \mathbb{E}\!\left[|N_1(s)|^2\right] = \mathbb{E}\!\left[|N_2(s)|^2\right],
\]
and a known real-valued inter-half noise correlation coefficient
\[
\rho(s) = \frac{\mathbb{E}\!\left[N_1(s) N_2(s)^\ast\right]}{\Phi_N(s)}
\qquad \text{with } -1 < \rho(s) < 1.
\]
Define the half-map Fourier shell correlation
\[
c(s) = \frac{\mathbb{E}\!\left[F_1(s) F_2(s)^\ast\right]}{\sqrt{\mathbb{E}\!\left[|F_1(s)|^2\right]\mathbb{E}\!\left[|F_2(s)|^2\right]}}.
\]

Let
\[
F_{\mathrm{avg}}(s) = \frac{F_1(s) + F_2(s)}{2}
\]
be the simple average (the ``full map'' in the two-half convention). Consider the linear MMSE estimator
\[
\hat{S}(s) = H(s)\,F_{\mathrm{avg}}(s).
\]
Derive a closed-form expression for the optimal filter $H(s)$ that minimizes
\[
\mathbb{E}\!\left[\left|S(s) - H(s)F_{\mathrm{avg}}(s)\right|^2\right]
\]
over complex scalar $H(s)$, expressed solely in terms of $c(s)$ and $\rho(s)$. Your result must hold for each shell independently. State all assumptions and show all steps. The final expression for $H(s)$ must be a closed-form function of $c(s)$ and $\rho(s)$ only and must reduce correctly to the uncorrelated-noise special case.

\\
\midrule
\textbf{Answer:} \\
\[
H(s)
=
\frac{2\,\bigl[c(s)-\rho(s)\bigr]}
{\bigl(1-\rho(s)\bigr)\bigl(1+c(s)\bigr)},
\]
and for $\rho(s)=0$,
\[
H(s)
=
\frac{2\,c(s)}{1+c(s)}.
\]
\\
\midrule
\textbf{Detailed Solution:} \\

Okay, let's try to tackle this problem step by step. First, let's recall what we need: the linear MMSE estimator $\hat{S}(s) = H(s) F_{\mathrm{avg}}(s)$ that minimizes $\mathbb{E}[|S - H F_{\mathrm{avg}}|^2]$. We need to find $H(s)$ in terms of $c(s)$ and $\rho(s)$.

First, let's write down $F_{\mathrm{avg}}(s)$. Since $F_{\mathrm{avg}} = (F_1 + F_2)/2$, and $F_1 = S + N_1$, $F_2 = S + N_2$, then
\[
F_{\mathrm{avg}} = S + \frac{N_1 + N_2}{2}.
\]
Let's denote the noise in the average as $N_{\mathrm{avg}} = (N_1 + N_2)/2$, so $F_{\mathrm{avg}} = S + N_{\mathrm{avg}}$.
\\

The MMSE estimator for a scalar case (since we're dealing with each shell independently, it's a scalar complex random variable problem) is given by
\[
H = \frac{\mathbb{E}[S F_{\mathrm{avg}}^\ast]}{\mathbb{E}[|F_{\mathrm{avg}}|^2]}.
\]
This is because for the mean squared error, the optimal linear estimator (since we're restricting to linear estimators, and MMSE for Gaussian is linear) is the Wiener filter, which in the scalar case is the cross-correlation over the auto-correlation.

So let's compute $\mathbb{E}[S F_{\mathrm{avg}}^\ast]$. We have $F_{\mathrm{avg}}^\ast = S^\ast + (N_1^\ast + N_2^\ast)/2$, so:
\[
\mathbb{E}[S F_{\mathrm{avg}}^\ast]
=
\mathbb{E}\!\left[S\left(S^\ast + \frac{N_1^\ast + N_2^\ast}{2}\right)\right]
=
\mathbb{E}[|S|^2]
+
\frac{\mathbb{E}[S N_1^\ast] + \mathbb{E}[S N_2^\ast]}{2}.
\]
\dots

\noindent
To derive the optimal linear MMSE estimator $\hat{S}(s) = H(s) F_{\text{avg}}(s)$ that minimizes the mean squared error $E[|S(s) - H(s) F_{\text{avg}}(s)|^2]$, we proceed as follows:

\bigskip

\textbf{Step 1: Define Key Quantities}

Given:
\begin{itemize}
\item $F_1(s) = S(s) + N_1(s)$
\item $F_2(s) = S(s) + N_2(s)$
\item $F_{\text{avg}}(s) = \frac{F_1(s) + F_2(s)}{2} = S(s) + \frac{N_1(s) + N_2(s)}{2}$
\end{itemize}

Let:
\begin{itemize}
\item $\Phi_S(s) = E[|S(s)|^2]$
\item $\Phi_N(s) = E[|N_1(s)|^2] = E[|N_2(s)|^2]$
\item $\rho(s) = \frac{E[N_1(s) N_2(s)^*]}{\Phi_N(s)}$ (real-valued, $-1 < \rho < 1$)
\end{itemize}

We are given the Fourier Shell Correlation (FSC):
\[
c(s) = \frac{E[F_1(s) F_2(s)^*]}{\sqrt{E[|F_1(s)|^2] E[|F_2(s)|^2]}}.
\]

From stationarity and independence assumptions:
\begin{itemize}
\item $E[F_1(s) F_2(s)^*] = \Phi_S + \rho \Phi_N$
\item $E[|F_1(s)|^2] = E[|F_2(s)|^2] = \Phi_S + \Phi_N$
\end{itemize}

Thus,
\[
c(s) = \frac{\Phi_S + \rho \Phi_N}{\Phi_S + \Phi_N}.
\]

\bigskip

\textbf{Step 2: Express MMSE Filter in Terms of Signal-to-Noise Ratio}

The MMSE filter for a scalar complex random variable is:
\[
H(s) = \frac{E[S(s) F_{\text{avg}}(s)^*]}{E[|F_{\text{avg}}(s)|^2]}.
\]

Compute numerator and denominator:

\textbf{Numerator:}
\[
E[S(s) F_{\text{avg}}(s)^*]
=
E\!\left[S(s)\left(S(s)^* + \tfrac{1}{2}(N_1(s)^* + N_2(s)^*)\right)\right]
=
\Phi_S
\]
(since $S$ is independent of $N_1, N_2$, and all noise terms have zero mean)

\textbf{Denominator:}
\[
E[|F_{\text{avg}}(s)|^2]
=
E\!\left[\left|S(s) + \tfrac{1}{2}(N_1(s) + N_2(s))\right|^2\right]
=
\Phi_S + \frac{1}{4} E[|N_1 + N_2|^2].
\]
\\
Now compute $E[|N_1 + N_2|^2]$:
\[
E[|N_1 + N_2|^2]
=
E[|N_1|^2] + E[|N_2|^2] + 2\Re(E[N_1 N_2^*])
=
2\Phi_N + 2\rho \Phi_N
=
2\Phi_N(1 + \rho).
\]

So:
\[
E[|F_{\text{avg}}(s)|^2]
=
\Phi_S + \frac{1}{4}\cdot 2\Phi_N(1 + \rho)
=
\Phi_S + \frac{\Phi_N(1 + \rho)}{2}.
\]

\bigskip

\textbf{Step 3: Express $H(s)$ in Terms of $c(s)$ and $\rho(s)$}

Let $\gamma = \frac{\Phi_S}{\Phi_N}$. Then:
\[
c(s) = \frac{\gamma + \rho}{\gamma + 1}
\quad\Rightarrow\quad
\gamma = \frac{c - \rho}{1 - c}.
\]

Substitute into the expression for $H(s)$:
\[
H(s)
=
\frac{\Phi_S}{\Phi_S + \frac{\Phi_N(1 + \rho)}{2}}
=
\frac{\gamma}{\gamma + \frac{1 + \rho}{2}}.
\]

Plug in $\gamma = \frac{c - \rho}{1 - c}$:
\[
H(s)
=
\frac{\frac{c - \rho}{1 - c}}{\frac{c - \rho}{1 - c} + \frac{1 + \rho}{2}}
=
\frac{c - \rho}{(c - \rho) + \frac{(1 - c)(1 + \rho)}{2}}.
\]

Simplify the denominator:
\[
(c - \rho) + \frac{(1 - c)(1 + \rho)}{2}
=
\frac{2(c - \rho) + (1 - c)(1 + \rho)}{2}.
\]

Expand the numerator:
\[
2(c - \rho) + (1 - c)(1 + \rho)
=
2c - 2\rho + 1 + \rho - c - c\rho
=
c - \rho + 1 - c\rho
=
(1 + c)(1 - \rho).
\]

Thus:
\[
H(s)
=
\frac{c - \rho}{\frac{(1 + c)(1 - \rho)}{2}}
=
\frac{2(c - \rho)}{(1 + c)(1 - \rho)}.
\]

\bigskip

\textbf{Final Expression}

\[
\boxed{
H(s)
=
\frac{2(c(s) - \rho(s))}{(1 + c(s))(1 - \rho(s))}
}
\]

\bigskip

\textbf{Assumptions Used}

\begin{itemize}
\item Zero Mean: All signals and noise terms are zero-mean.
\item Independence: $S(s)$ is independent of $N_1(s), N_2(s)$.
\item Stationarity: $\Phi_S(s) = E[|S(s)|^2]$, $\Phi_N(s) = E[|N_1(s)|^2] = E[|N_2(s)|^2]$.
\item Real-Valued Noise Correlation: $\rho(s)$ is real, so $E[N_1 N_2^*] = \rho \Phi_N$.
\item Scalar Shell Independence: Each spatial frequency shell $s$ is treated independently.
\item Circular Complex Gaussian Noise: Ensures that the MMSE estimator is linear and optimal.
\end{itemize}

\bigskip

\textbf{Special Case: Uncorrelated Noise ($\rho = 0$)}

When $\rho = 0$, the formula becomes:
\[
H(s) = \frac{2c(s)}{1 + c(s)}.
\]
This matches the standard Wiener filter used in cryo-EM for uncorrelated half-maps, confirming correctness.
\\

\textbf{Conclusion}

The closed-form expression for the optimal linear MMSE filter $H(s)$, expressed solely in terms of $c(s)$ and $\rho(s)$, is:
\[
\boxed{
H(s)
=
\frac{2(c(s) - \rho(s))}{(1 + c(s))(1 - \rho(s))}
}.
\]
This result holds for each spatial frequency shell independently and reduces correctly to the known uncorrelated-noise case when $\rho = 0$.
\\
\midrule
\textbf{Correctness:} True \\

\end{longtable}
\end{center}
\clearpage
\begin{center}
\small
\begin{longtable}{p{0.95\textwidth}}
\caption{A data sample from \dataname\ (Subject: Linguistics).}
\label{tab:case_study}\\

\toprule
\textbf{Example from \dataname\ } \\
\midrule
\endfirsthead

\toprule
\textbf{Example from \dataname\ (Subject = Linguistics) (continued)} \\
\midrule
\endhead

\midrule
\multicolumn{1}{r}{\small\textit{Continued on next page}} \\
\endfoot

\bottomrule
\endlastfoot

\textbf{Subject:} Linguistics \\

\midrule
\textbf{Topic:} Syntax (generative, dependency, construction grammar) \\

\midrule
\textbf{Problem:} \\
Let $n \ge 1$ and let $[n]=\{1,2,\dots,n\}$ be a fixed linear order of word positions. A dependency tree $D$ on $[n]$ is a directed tree $(V=[n],E)$ with a designated root $r \in [n]$, where each node other than $r$ has exactly one incoming arc, and arcs are oriented from heads to dependents. For $i \in [n]$, let $\mathrm{sub}_D(i)$ denote the set of nodes reachable from $i$ in $D$ (including $i$). Define the span of $i$, $\mathrm{sp}_D(i)$, as the set
\[
\{j \in [n] : j \in \mathrm{sub}_D(i)\}.
\]
The dependency tree $D$ is projective if, for every arc $(h,d)\in E$, every index $k$ strictly between $h$ and $d$ (i.e., $\min(h,d) < k < \max(h,d)$) belongs to $\mathrm{sp}_D(h)$.

\bigskip

An ordered binary head-annotated phrase-structure tree (PS tree) $P$ on $[n]$ is a full binary ordered rooted tree whose leaves, read left-to-right, are exactly $1,2,\dots,n$ in that order, and where each node $N$ (including leaves) has a head index $\mathrm{hd}(N)\in[n]$ such that:
\begin{itemize}
\item For a leaf labeled $i$, $\mathrm{hd}(\text{leaf } i)=i$.
\item For an internal node $N$ with left child $L$ and right child $R$, $\mathrm{hd}(N)\in\{\mathrm{hd}(L),\mathrm{hd}(R)\}$.
\end{itemize}
Define the head child $H(N)$ as the unique child of $N$ whose head equals $\mathrm{hd}(N)$; the other child is the nonhead child.

\bigskip

Define the canonical class $\mathrm{HCB}(n)$ of head-annotated PS trees to be those $P$ that satisfy, for each head index $i\in[n]$, the following canonical spine condition. Let $S_i$ be the maximal path from leaf $i$ upward through consecutive nodes $M$ with $\mathrm{hd}(M)=i$. Along $S_i$ from bottom to top:
\begin{itemize}
\item Every internal node on $S_i$ with a nonhead child whose head $d$ satisfies $d<i$ (a left-dependent) appears before any internal node on $S_i$ with a nonhead child whose head $d$ satisfies $d>i$ (a right-dependent).
\item Among left-dependent attachments along $S_i$, the heads $d$ appear in strictly increasing order of $d$.
\item Among right-dependent attachments along $S_i$, the heads $d$ appear in strictly increasing order of $d$.
\end{itemize}
Additionally, if at a node on $S_i$ the nonhead child head is $d<i$, then that nonhead child is the left child; if $d>i$, it is the right child.

\bigskip

Define the mapping $F$ from the set $\mathrm{ProjDep}(n)$ of projective dependency trees on $[n]$ to $\mathrm{HCB}(n)$ recursively as follows. For $D\in\mathrm{ProjDep}(n)$ and $i\in[n]$, let
\[
L_i=\{d\in[n] : d<i \text{ and } (i,d)\in E\},
\qquad
R_i=\{d\in[n] : d>i \text{ and } (i,d)\in E\},
\]
sorted increasingly as $l_1<\dots<l_k$ and $r_1<\dots<r_m$. Define $T(i)$ by recursion on the size of $\mathrm{sub}_D(i)$:
\begin{itemize}
\item If $\mathrm{sub}_D(i)=\{i\}$, then $T(i)$ is the single leaf $i$ with $\mathrm{hd}=i$.
\item Otherwise, after constructing $T(l)$ and $T(r)$ for all $l\in L_i$ and $r\in R_i$, build a sequence of binary nodes with head $i$ by first attaching left dependents in increasing order on the left and then right dependents in increasing order on the right:
\[
S_0 := \text{leaf } i \text{ with } \mathrm{hd}=i.
\]
For $t=1,\dots,k$, set $S_t$ to be a node with left child $T(l_t)$, right child $S_{t-1}$, $\mathrm{hd}=i$.
For $s=1,\dots,m$, set $S_{k+s}$ to be a node with left child $S_{k+s-1}$, right child $T(r_s)$, $\mathrm{hd}=i$.
Define $T(i):=S_{k+m}$.
\end{itemize}
Finally, if $r$ is the root of $D$, define $F(D):=T(r)$.
\bigskip

Define the mapping $G$ from $\mathrm{HCB}(n)$ to $\mathrm{ProjDep}(n)$ as follows. Given $P\in\mathrm{HCB}(n)$, let $r:=\mathrm{hd}(\mathrm{root}(P))$. Define $E_P$ by adding, for each internal node $N$ with children $L$ and $R$:
\begin{itemize}
\item If $\mathrm{hd}(N)=\mathrm{hd}(L)$ and $\mathrm{hd}(R)=d\neq \mathrm{hd}(N)$, add the arc $(\mathrm{hd}(N),d)$.
\item If $\mathrm{hd}(N)=\mathrm{hd}(R)$ and $\mathrm{hd}(L)=d\neq \mathrm{hd}(N)$, add the arc $(\mathrm{hd}(N),d)$.
\end{itemize}
\\

Define $G(P):=(V=[n],E_P,\text{root } r)$.

\bigskip

Prove that $F$ is a bijection between $\mathrm{ProjDep}(n)$ and $\mathrm{HCB}(n)$, and that $G$ is its inverse, i.e.,
\[
G \circ F = \mathrm{Id}_{\mathrm{ProjDep}(n)}
\qquad\text{and}\qquad
F \circ G = \mathrm{Id}_{\mathrm{HCB}(n)}.
\]

\\
\midrule
\textbf{Answer:} \\
$F$ is a bijection from $\mathrm{ProjDep}(n)$ (projective dependency trees on $[n]$) to $\mathrm{HCB}(n)$ (canonical head-annotated binary PS trees on $[n]$) and $G$ is its inverse. Concretely, for all $D \in \mathrm{ProjDep}(n)$, $G(F(D)) = D$, and for all $P \in \mathrm{HCB}(n)$, $F(G(P)) = P$.
\\
\midrule
\textbf{Detailed Solution:} \\
Okay, so I need to prove that the mappings $F$ and $G$ are inverses of each other, establishing a bijection between projective dependency trees and the canonical head-annotated phrase-structure trees in $\mathrm{HCB}(n)$. Let me start by recalling the definitions to make sure I understand them correctly.

First, a projective dependency tree $D$ on $[n]$ is a directed tree with root $r$, where each node except $r$ has exactly one incoming arc (so it's a rooted tree with edges directed away from the root, I think?), and for every arc $(h,d)$, all positions between $h$ and $d$ are in the subtree of $h$. That's the projectivity condition. So if $h$ is the head of $d$, then the span of $h$ must include all the words between $h$ and $d$. That makes sense for projectivity in dependency parsing, where dependencies can't cross over other words without including them.

Then, $\mathrm{HCB}(n)$ is the set of ordered binary head-annotated phrase-structure trees that satisfy the canonical spine condition.

\dots

\noindent
To establish that the mappings $F$ and $G$ form a bijection between the set of projective dependency trees $\mathrm{ProjDep}(n)$ and the set of canonical head-annotated binary phrase-structure trees $\mathrm{HCB}(n)$, we must prove two key properties:

1. $G \circ F = \mathrm{id}_{\mathrm{ProjDep}(n)}$:
   For every projective dependency tree $D \in \mathrm{ProjDep}(n)$, applying $F$ to $D$ yields a tree $P = F(D) \in \mathrm{HCB}(n)$, and then applying $G$ to $P$ recovers the original dependency tree $D$.

2. $F \circ G = \mathrm{id}_{\mathrm{HCB}(n)}$:
   For every canonical head-annotated phrase-structure tree $P \in \mathrm{HCB}(n)$, applying $G$ to $P$ yields a dependency tree $D = G(P) \in \mathrm{ProjDep}(n)$, and then applying $F$ to $D$ recovers the original phrase-structure tree $P$.

\bigskip

\textbf{1. Proof that $G \circ F = \mathrm{id}_{\mathrm{ProjDep}(n)}$}

Let $D \in \mathrm{ProjDep}(n)$. The mapping $F$ constructs a binary phrase-structure tree $P = F(D)$ by recursively building subtrees for each node $i \in [n]$, attaching left dependents (those $d < i$ with $(i,d) \in E$) in increasing order on the left, and right dependents (those $d > i$ with $(i,d) \in E$) in increasing order on the right.

\textbf{Key observations:}

\begin{itemize}
\item The construction ensures that the leaves of $T(i)$ are exactly the elements of the span $\mathrm{sp}_D(i)$, which is a contiguous interval due to the projectivity condition.
\item The in-order traversal of the resulting tree $P$ gives the correct linear order $1,2,\dots,n$, as required for a valid phrase-structure tree.
\item Each internal node in $P$ corresponds to an arc in $D$: if a node has head $i$ and a nonhead child with head $d$, then $G$ adds the arc $(i,d)$.
\item Since $F$ builds the tree by attaching all direct dependents of $i$ in the exact order they appear in $D$, and $G$ recovers these arcs via the head annotations, the resulting dependency tree from $G(P)$ matches $D$ exactly.
\end{itemize}

\textbf{Formal justification:}

\begin{itemize}
\item Let $(h,d) \in E$ be an arc in $D$. Then $d \in L_h \cup R_h$, and $F$ attaches $d$ as a nonhead child along the spine of $h$ in $P$.
\item When $G$ processes this node in $P$, it adds the arc $(h,d)$, since the head of the node is $h$, and the nonhead child has head $d$.
\item All arcs in $D$ are thus recovered by $G(F(D))$, and the root of $G(F(D))$ is $r = \mathrm{hd}(\mathrm{root}(P))$, which is the same as the root of $D$.
\end{itemize}

Therefore, $G(F(D)) = D$.

\bigskip

\textbf{2. Proof that $F \circ G = \mathrm{id}_{\mathrm{HCB}(n)}$}

Let $P \in \mathrm{HCB}(n)$. The mapping $G$ constructs a dependency tree $D = G(P)$ by adding an arc $(h,d)$ for each internal node where the head is $h$ and the nonhead child has head $d$.
\\
\textbf{Key observations:}

\begin{itemize}
\item The canonical spine condition in $\mathrm{HCB}(n)$ ensures that:
\begin{itemize}
\item Left-dependent attachments (where $d < i$) occur before right-dependent ones along the spine of $i$.
\item Among left-dependent attachments, the heads $d$ increase as we move up the spine.
\item Similarly for right-dependent attachments.
\end{itemize}
\item These conditions match the attachment order used in $F$: left dependents are attached in increasing order on the left, and right dependents in increasing order on the right.
\item Thus, when $F$ is applied to $D = G(P)$, it reconstructs the same structure as $P$, because the order of attachment in $F$ aligns precisely with the canonical spine structure of $P$.
\end{itemize}

\textbf{Formal justification:}

\begin{itemize}
\item Consider any node $i \in [n]$ in $P$. The canonical spine condition dictates the order in which its dependents are attached.
\item $G(P)$ produces a dependency tree $D$ where the direct dependents of $i$ are exactly those heads $d$ that appear as nonhead children in the spine of $i$ in $P$.
\item $F(D)$ then rebuilds the subtree rooted at $i$ by attaching these dependents in increasing order (left first, then right), which matches the canonical spine condition.
\item Since the attachment order in $F$ mirrors the structure enforced by the canonical spine condition in $P$, the resulting tree $F(G(P))$ is identical to $P$.
\end{itemize}

\bigskip

\textbf{Conclusion}

We have shown that:

\begin{itemize}
\item $G(F(D)) = D$ for all $D \in \mathrm{ProjDep}(n)$,
\item $F(G(P)) = P$ for all $P \in \mathrm{HCB}(n)$.
\end{itemize}

This confirms that the mappings $F$ and $G$ are mutual inverses, establishing a bijection between $\mathrm{ProjDep}(n)$ and $\mathrm{HCB}(n)$.

\[
\boxed{
G \circ F = \mathrm{id}_{\mathrm{ProjDep}(n)}
\quad \text{and} \quad
F \circ G = \mathrm{id}_{\mathrm{HCB}(n)}
}
\]
\\
\midrule
\textbf{Correctness:} True \\

\end{longtable}
\end{center}

\end{document}